\documentclass[UTF8]{article} 
\usepackage{iclr2023_conference,times}

\usepackage{microtype}
\usepackage{amsmath}
\usepackage{amssymb}
\usepackage{booktabs}
\usepackage{colortbl}
\usepackage{CJKutf8}
\usepackage[utf8]{inputenc}
\definecolor{lightgray}{rgb}{0.9,0.9,0.9}
\usepackage{caption}
\usepackage{multicol}
\usepackage{wrapfig}
\usepackage{subcaption}
\usepackage{xcolor}
\usepackage{graphicx}
\usepackage{setspace}
\usepackage{hyperref}
\usepackage{url}
\usepackage{multirow}
\usepackage{colortbl}
\usepackage{tabularx}
\usepackage{blindtext}
\usepackage{pgfplots}
\pgfplotsset{compat=1.18} 
\usepackage{tikz}
\usetikzlibrary{er,positioning,bayesnet}
\usepackage[inline]{enumitem}
\usepackage{makecell}
\usepackage{tipa}
\usepackage{siunitx}
\usepackage{tocloft}
\usepackage{listings}
\usepackage[raster,skins]{tcolorbox} 
\usepackage{xltabular}
\usepackage{hyperref}
\usepackage[framemethod=tikz]{mdframed}
\surroundwithmdframed[
  hidealllines=true,
  innerleftmargin=0pt,
  innertopmargin=0pt,
  innerbottommargin=0pt]{lstlisting}
\lstnewenvironment{response}[1][] 
 {\lstset{
 columns=fullflexible,
  breakautoindent=false, breakindent=0pt, breaklines, linewidth=8cm, #1}}
 {}
\tcbuselibrary{breakable}

\newcommand{\method}{\texttt{Wan}\xspace}

\newcommand{\ie}{\textit{i}.\textit{e}., }
\newcommand{\eg}{\textit{e}.\textit{g}. }

\newlength\savewidth\newcommand\shline{\noalign{\global\savewidth\arrayrulewidth
  \global\arrayrulewidth 1pt}\hline\noalign{\global\arrayrulewidth\savewidth}}

\newcommand{\warptablestyle}[2]{\setlength{\tabcolsep}{#1}\renewcommand{\arraystretch}{#2}\centering\small}

\definecolor{url_color}{RGB}{113, 187, 179}

\hypersetup{
    colorlinks=true,
    linkcolor=black,
    urlcolor=url_color,
}

\title{\method: Open and Advanced Large-Scale Video Generative Models}
\author{
\textbf{Wan Team, Alibaba Group}
}

\iclrfinalcopy
\begin{document}

\maketitle

\begin{abstract}
This report presents \method, a comprehensive and open suite of video foundation models designed to push the boundaries of video generation.
Built upon the mainstream diffusion transformer paradigm, \method achieves significant advancements in generative capabilities through a series of innovations, including our novel spatio-temporal variational autoencoder (VAE), scalable pre-training strategies, large-scale data curation, and automated evaluation metrics.
These contributions collectively enhance the model's performance and versatility.
Specifically, \method is characterized by four key features:
\emph{Leading Performance}: The 14B model of \method, trained on a vast dataset comprising billions of images and videos, demonstrates the scaling laws of video generation with respect to both data and model size.
It consistently outperforms the existing open-source models as well as state-of-the-art commercial solutions across multiple internal and external benchmarks, demonstrating a clear and significant performance superiority.
\emph{Comprehensiveness}: \method offers two capable models, \emph{i.e.}, 1.3B and 14B parameters, for efficiency and effectiveness respectively. 
It also covers multiple downstream applications, including image-to-video, instruction-guided video editing, and personal video generation, encompassing up to eight tasks.
Meanwhile, \method is the first model that can generate visual text in both Chinese and English, significantly enhancing its practical value.
\emph{Consumer-Grade Efficiency}: The 1.3B model demonstrates exceptional resource efficiency, requiring only 8.19 GB VRAM, making it compatible with a wide range of consumer-grade GPUs. It also exhibits superior performance compared to larger open-source models, showcasing remarkable efficiency for text-to-video.
\emph{Openness}: We open-source the entire series of \method, including source code and all models, with the goal of fostering the growth of the video generation community. 
This openness seeks to significantly expand the creative possibilities of video production in the industry and provide academia with high-quality video foundation models.
In addition, we conduct extensive experimental analyses covering various aspects of the proposed \method, presenting detailed results and insights.
We believe these findings and conclusions will significantly advance video generation technology. 
All the code and models are available at \textbf{\url{https://github.com/Wan-Video/Wan2.1}}.
\end{abstract}

\vspace{-7mm}

\begin{figure}[b]  
    \centering  
    \includegraphics[width=1.0\textwidth]{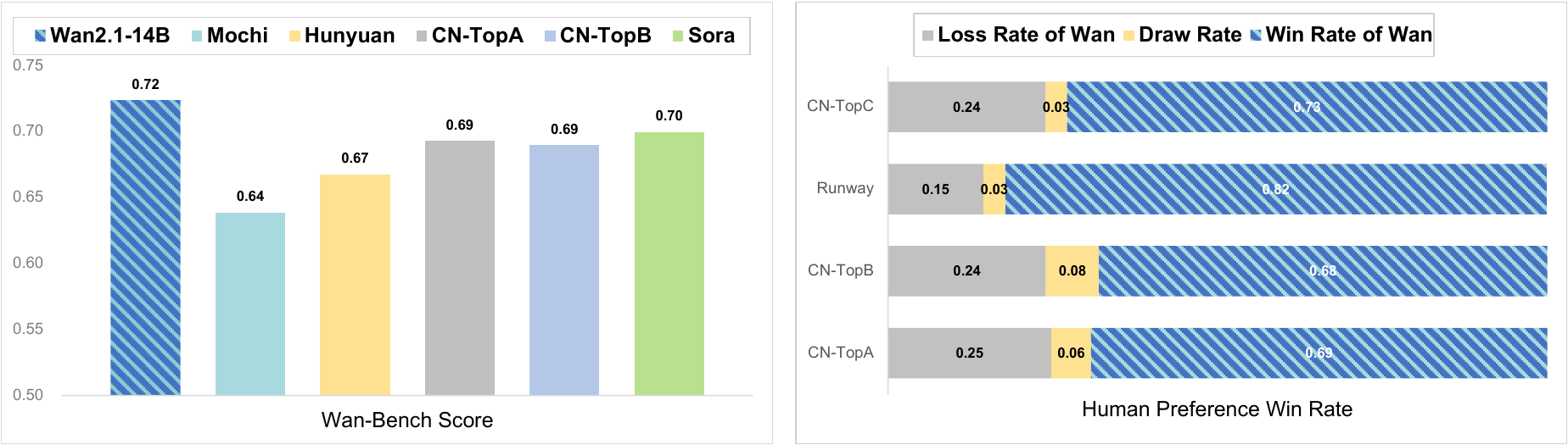}  
    \caption{Comparison of \method with state-of-the-art open-source and closed-source models. Following both benchmark and human evaluations, \method consistently demonstrated superior results. Note that HunyuanVideo~\cite{kong2024hunyuanvideo} is tested using the open-source model.} 
    \label{fig:overview_network} %
\end{figure}

\clearpage

\newpage

\newcommand{\customsize}{\fontsize{7.7}{9}\selectfont}

\begingroup
\footnotesize
\begin{spacing}{0.88}
\tableofcontents
\end{spacing}
\endgroup

\hypersetup{linkcolor=url_color}

\newpage


\section{Introduction}
\label{sec:intro}

Since the introduction of Sora~\citep{openaisora2024} by OpenAI, video generation technology has attracted substantial attention from both industry and academia, leading to rapid advancements in the field.
The emergence of models capable of generating videos that are on par with professionally crafted content has significantly improved the efficiency of content creation while simultaneously reducing the costs associated with video production.
These rapid advancements in video generation technology have also been greatly attributed to the development of the open-source community.
Notable projects like HunyuanVideo~\citep{kong2024hunyuanvideo}, Mochi~\citep{genmo2024mochi}, and CogVideoX~\citep{cogvideox} have made their video foundation model codes and weights publicly available, gradually narrowing the gap between open-source models and commercial counterparts.
However, it is essential to acknowledge the persistent gap between these outstanding open-source models and the latest closed-source models. 
This gap is primarily evident in three aspects:
\emph{Suboptimal Performance}: A notable performance gap remains, as the pace of development in commercial models far exceeds that of current open-source models, resulting in significantly superior capabilities.
\emph{Limited Capabilities}: Most foundational models are limited to general text-to-video (T2V) tasks, whereas the demands of video creation are multifaceted. Consequently, basic T2V models are insufficient to address these diverse requirements.
\emph{Insufficient Efficiency}: Despite their impressive performance and scale, these models often prove impractical for creative teams with limited computational resources, hindering their accessibility and usability.
These challenges collectively impose constraints on the continued growth and innovation within the open-source community.

To address the aforementioned challenges, this report introduces and publicly releases a novel series of high-performance foundational video generation models, referred to as \method, which sets a new benchmark in the field.
The core design of \method is inspired by the success of Diffusion Transformers (DiT)~\citep{dit} combined with Flow Matching~\citep{flow-matching}, a framework that has demonstrated the significant performance gains achievable through scaling in both text-to-image (T2I)~\citep{esser2024sd3} and text-to-video (T2V)~\citep{kong2024hunyuanvideo} tasks.
Within this architectural paradigm, cross-attention is employed to embed text conditions, while the model's design is meticulously optimized to ensure computational efficiency and precise text controllability. 
To further enhance the model's ability to capture complex dynamics, a full spatio-temporal attention mechanism is incorporated.
Through extensive experimentation, the model is validated at scale, reaching 14 billion parameters.
Subsequently, \method has seen large-scale data comprising billions of images and videos, amounting to $\mathcal{O}(1)$ trillions of tokens in total.
This extensive training facilitates the emergence of the model's capabilities, allowing it to demonstrate robust performance across multiple dimensions, such as motion amplitude and quality, visual text generation, camera control, instruction adherence, and stylistic diversity.
Building upon the powerful foundational model, we have expanded its capabilities to numerous downstream tasks, including image-to-video generation (I2V), instruction-guided video editing (V2V), zero-shot personalized customization, real-time video generation, and audio generation, among other critical applications.
To minimize inference costs, we also introduce a 1.3B model alongside a 14B model for T2V and I2V, both of which support 480p resolution and greatly enhance inference efficiency. 
Remarkably, the 1.3B model requires only 8.19G of VRAM, allowing it to run on many consumer-grade GPUs, while its performance exceeds that of many larger open-source models.

Moreover, we will also publicly present the entire training process, including the large-scale data construction pipeline, video variational autoencoder (VAE), training strategies, acceleration techniques, and automated evaluation algorithms, to empower the community in developing specialized foundational video models.
In addition, we will provide comprehensive design details and experimental results, offering insights into phenomena observed during the resource-intensive training of large generative models alongside our key findings and conclusions.
We are confident that these contributions will play a pivotal role in accelerating the advancement of video generation technology.

\begin{figure}[t]  
    \centering  
    \includegraphics[width=1.0\textwidth]{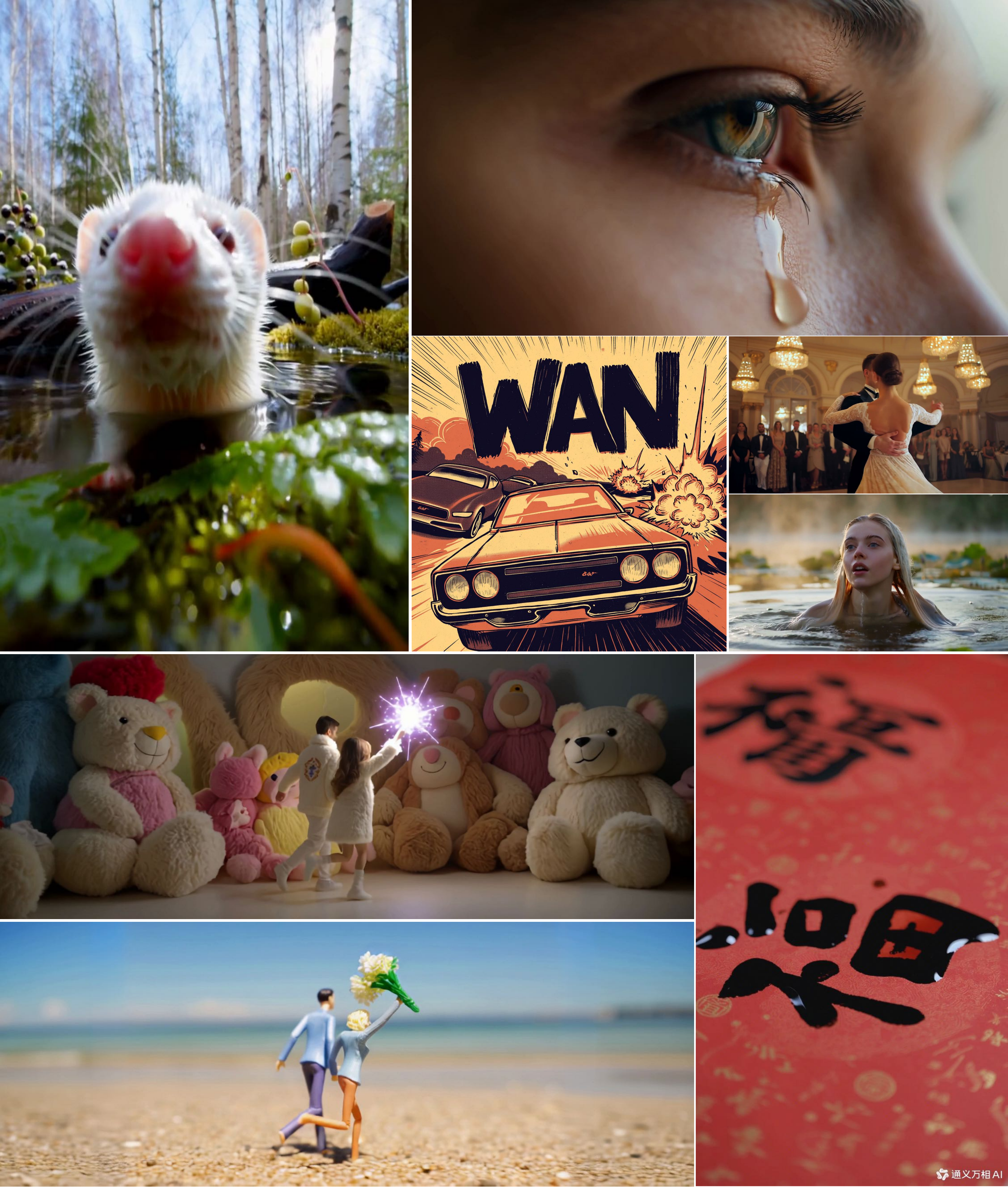}  
    \caption{Samples generated by \method. \method can produce videos with large motions, high fidelity, and realistic details. It is also the first model to generate both Chinese and English text within videos. Additionally, \method offers a range of features, including text-to-video, image-to-video, and video editing capabilities.} 
    \vspace{-2mm}
    \label{fig:wan_examples_01} %
\end{figure}

\section{Related Work}
\label{sec:related-work}

Driven by advances in generative modeling, the landscape of large-scale video models has evolved significantly, particularly in diffusion-based frameworks. Our review focuses on two broad categories: closed-source models and contributions from the open-source community.

\textbf{Closed-source models.}
Closed-source models are mainly those developed by major technology companies, which aims at high-quality, professional video generation due to the extensive resources invested. We organize the notable models released over the past year in chronological order.
In February 2024, OpenAI introduced Sora~\citep{openaisora2024} and marked a significant leap in AI-generated content. In June 2024, Kling~\citep{kuaishou2024kling} by Kuaishou and Luma 1.0~\citep{luma2024dm} by LUMA AI were opened up for public testing, offering powerful video generation capabilities. Meanwhile, Runway unveiled Gen-3~\citep{runway2024gen3}, building upon Gen-2~\citep{runway2024gen2}, to further elevate the standards in video creation. In July 2024, Shengshu AI released Vidu~\citep{bao2024vidu} equipped with self-designed U-ViT~\citep{bao2022all} architecture.
In September 2024, both Kling and Luma undergone updates to version 1.5.
During the same period, MiniMax introduced Hailuo Video~\citep{minimax2024hailuo}, delivering impressive visual results to the public. In October 2024, Pika Labs launched Pika 1.5~\citep{pika2024pika}, enabling users to customize visual and physical attributes in videos. In addition, Meta introduced Movie Gen~\citep{polyak2024moviegencastmedia}, a series of video foundation models that detailed their training processes and applications. In December 2024, Kling advanced to version 1.6 and, concurrently, Google released Veo 2~\citep{veo22025} with an improved understanding of real-world physics and human movement nuances.

These developments underscore the intense global competition in the video generation sector. In this context, our open-source \method has shown competitive or even superior performance over these commercial models across both internal and external benchmarks, leading in multiple attributes.

\textbf{Contributions from open-source community.}
On the other hand, the open-source community has made substantial contributions not only to the holistic video generative models but also to the exploration of essential model components. Diffusion-based video generation models typically build on Stable Diffusion~\citep{rombach2022sd} architecture. It mainly involves three critical modules: an autoencoder to map the original video into a compact latent space, a text encoder to extract text embeddings, and a neural network optimized via the diffusion model to learn the distribution of these video latents.
For network structure, U-Net~\citep{ronneberger2015u}, initially used in image generation~\citep{ho2020denoising,song2020denoising}, is adapted for video generation by incorporating temporal dimensions.
VDM~\citep{ho2022video} extends the 2D U-Net to a 3D version, while another paradigm~\citep{zhou2022magicvideo,wang2023modelscope,guo2023animatediff} introduces 1D temporal attention combined with 2D spatial attention block to reduce computation costs. Notably, Diffusion Transformers (DiT~\citep{dit}), which rely solely on transformer blocks, are superior to U-Net in visual generation tasks~\citep{chen2023pixartalpha}. This structure has also been transferred to video models~\citep{ma2024latte}, giving rise to two common variants: the original DiT~\citep{dit,HaCohen2024LTXVideo} that uses cross-attention for text embeddings and MM-DiT~\citep{genmo2024mochi,kong2024hunyuanvideo}, where text embeddings are concatenated to visual embeddings for full-attention processing.
Regarding autoencoders,
while the early approach~\citep{rombach2022sd} adopts standard VAE~\citep{kingma2013auto}, recent autoencoders like VQ-VAE~\citep{van2017neural} and VQGAN~\citep{esser2020taming} improve model design for better reconstruction and compression. LTX-Video~\citep{HaCohen2024LTXVideo} modifies the VAE decoder to perform the final denoising step and convert latents into pixels, where the missing high-frequency details are generated during decoding. The text encoder is also critical for text-based video generation. Current powerful video generation models primarily utilize the T5 series~\citep{t5} as the main text encoder, often in conjunction with CLIP~\citep{radford2021learning}.
In the case of HunyuanVideo~\citep{kong2024hunyuanvideo}, T5 is replaced with a Multimodal Large Language Model~\citep{llava,li2024mgm} to achieve more robust alignment between text embeddings and visual features.

By integrating these key modules with effective diffusion-based optimization techniques~\citep{ho2020denoising,song2019generative,flow-matching}, multiple promising open-source video generation models have emerged~\citep{genmo2024mochi,kong2024hunyuanvideo,HaCohen2024LTXVideo,opensora,lin2024open,jin2024pyramidal,cogvideox}.
In \method, we have carefully designed or selected each critical component to ensure high-quality video synthesis.
We provide detailed design information and ablation study, which can facilitate the design of future video generation models.

In addition, numerous studies have explored downstream tasks in video generation. These tasks include repainting~\citep{sdinp,propainter}, editing~\citep{sdedit,ip2p,magicbrush,wang2023videocomposer,dreamVideo2}, controllable generation~\citep{controlnet,scedit,wang2024unianimate}, and frame reference generation~\citep{cogvideox,i2vadapter}, often by leveraging adapter-based and ControlNet-like architectures~\citep{controlnet,chu2024diffcomplete} to incorporate user-specified conditions. We also develop various downstream applications based on \method, which have demonstrated excellent performance.

\section{Data Processing Pipeline}
\label{sec:data_processing}

High-quality data is essential for training large generative models, and an automated data construction pipeline significantly enhances the efficiency of the training process.
In developing our dataset, we prioritized three core principles: \emph{high quality}, \emph{high diversity}, and \emph{substantial scale}.
Following these principles, we curated a dataset comprising billions of videos and images. 
This section offers a detailed introduction to the data construction pipeline employed for \method.

\subsection{Pre-training Data}

We curated and deduplicated a candidate dataset sourced from both internal copyrighted sources and publicly accessible data. 
In the pre-training stage, our goal is to select high-quality and diverse data from this expansive yet noisy dataset to facilitate effective training.
Throughout the data mining process, we designed a four-step data cleaning process, focusing on fundamental dimensions, visual quality, and motion quality. 
Subsequently, we will also highlight our data processing workflow for constructing visual text data.

\textbf{Fundamental dimensions.} 
The fundamental dimensions of our data filtering framework focus on the intrinsic attributes of the source video and image data, enabling efficient preliminary filtering out of all the unsuitable data.
Specifically, our multidimensional filtering approach encompasses the following critical aspects:
\emph{Text detection}. A lightweight OCR detector is implemented to quantify text coverage ratios, effectively excluding videos and images with excessive textual elements to maintain visual clarity.
\emph{Aesthetic evaluation}: We use the widely adopted LAION-5B~\citep{schuhmann2021laion} aesthetic classifier to perform an initial quality assessment of our images, quickly filtering out low-quality data.
\emph{NSFW Score}. Through our internal safety assessment model, we systematically evaluate and filter inappropriate content based on computed NSFW scores in all training data.
\emph{Watermark and logo detection}. We detect whether the video or images contain watermarks and logos, and crop these elements during training.
\emph{Black border detection}. Utilizing heuristic-based detection methods, we automatically crop extraneous black borders to maintain focus on content-rich regions.
\emph{Overexposure detection}. Our trained expert classifier evaluates and filters out data with abnormal tonal distributions, ensuring optimal visual quality in the training dataset.
\emph{Synthetic image detection}. Empirical evidence indicates that even minimal contamination ($<10\%$) by generated images can significantly degrade the performance of the model. Therefore, we train an expert classifier to filter out these ``contaminating" images.
\emph{Blur detection}. An internally developed model assigns quantitative blur scores to training materials, enabling the systematic removal of visually indistinct content.
\emph{Duration and resolution}. We also enforce constraints where video duration must exceed 4 seconds, and resolution thresholds are applied at different training stages to filter out low-quality data.
Through the implementation of these efficient preprocessing strategies, we successfully eliminated approximately 50\% of the initial dataset. 
The retained high-quality data subsequently progresses to a more superior semantic-driven selection stage for further refinement.

\textbf{Visual quality.}  
This primarily involves selecting data from the candidate dataset that is of relatively high quality and meets pre-training standards.
This subset of data must have good visual quality and its overall distribution should align with that of natural data.
In this process, we divide the task into two parts: \emph{clustering} and \emph{scoring}. 
\emph{Clustering} is used to split all the data into smaller subsets, allowing us to process each subset individually. 
The benefit of this approach is that it prevents the loss of small yet important data segments that might occur due to a long-tail distribution, thereby maintaining the original data distribution. 
Specifically, we divide the data into 100 clusters and then select a certain amount of data from each cluster for the next stage of data processing.
\emph{Scoring} is used to assign a quality score to each video or image, facilitating the selection and processing of data at different stages. 
Specifically, we select samples from each cluster for manual scoring (ranging from 1 to 5, with 1 being the worst and 5 being the best). 
We then train an expert assessment model using all the annotated data to score the entire dataset. 

\textbf{Motion quality.}  
The goal of motion quality assessment is to select videos that are natural, complete, and with significant motion, while avoiding static or jittery movements. 
We classify video data into six distinct motion quality tiers.
\emph{Optimal motion}: This level represents videos with optimal attributes, such as significant motion layout, perspective, and amplitude, along with clean, smooth movement. 
\emph{Medium-quality motion}: Videos in this category exhibit noticeable movements but may have minor issues like multiple subjects or partial occlusion. This data ensures motion diversity and can be used during pre-training to help the model better understand spatio-temporal relationships. 
\emph{Static Videos}: This category primarily focuses on the videos, which consist largely of chat and interview-style videos. While these videos contain minimal motion information, they are of high quality. Therefore, we need to identify them separately and reduce their sampling ratio.
\emph{Camera-driven Motion}: Footage dominated by camera movement (\emph{e.g.}, aerial shots) with minimal subject motion. Given their similarity to static images, these receive substantially lower sampling priority.
\emph{Low-quality Motion}: Videos exhibiting excessive subjects, severe occlusions, or unclear main subjects (\emph{e.g.}, crowded street scenes). Such content is excluded due to its negative impact on training efficiency and motion generation quality.
\emph{Shaky camera footage}: Amateur recordings with pronounced camera shake, often causing motion blur and ambiguous foreground-background differentiation. This category is systematically excluded from training consideration.

\textbf{Visual text data.} 
In addition, we also introduce a novel data processing approach to enhance the visual text generation of \method. 
This approach comprises two distinct processing branches designed to improve both the accuracy and the harmony of text rendering.
On one hand, we synthesize hundreds of millions of text-containing images by rendering Chinese characters on a pure white background. 
On the other hand, we collect large amounts of text-containing images from vast real-world data. 
We employ multiple OCR models to accurately recognize both English and Chinese text from images and videos, and subsequently input these extracted text contents into the multimodal language model Qwen2-VL. 
This process generates natural descriptions of the images, ensuring that the descriptions incorporate as much precise text content as possible. 
We collect a mass of real-world image-text pairs by leveraging this processing pipeline.
By integrating both synthetic and real data during the pre-training stage, our approach effectively generalizes to generate rare words in videos with accurate glyphs and high realism, demonstrating significant superiority in visual text generation.

\begin{figure}[t]  
    \centering  
    \includegraphics[width=1.0\textwidth]{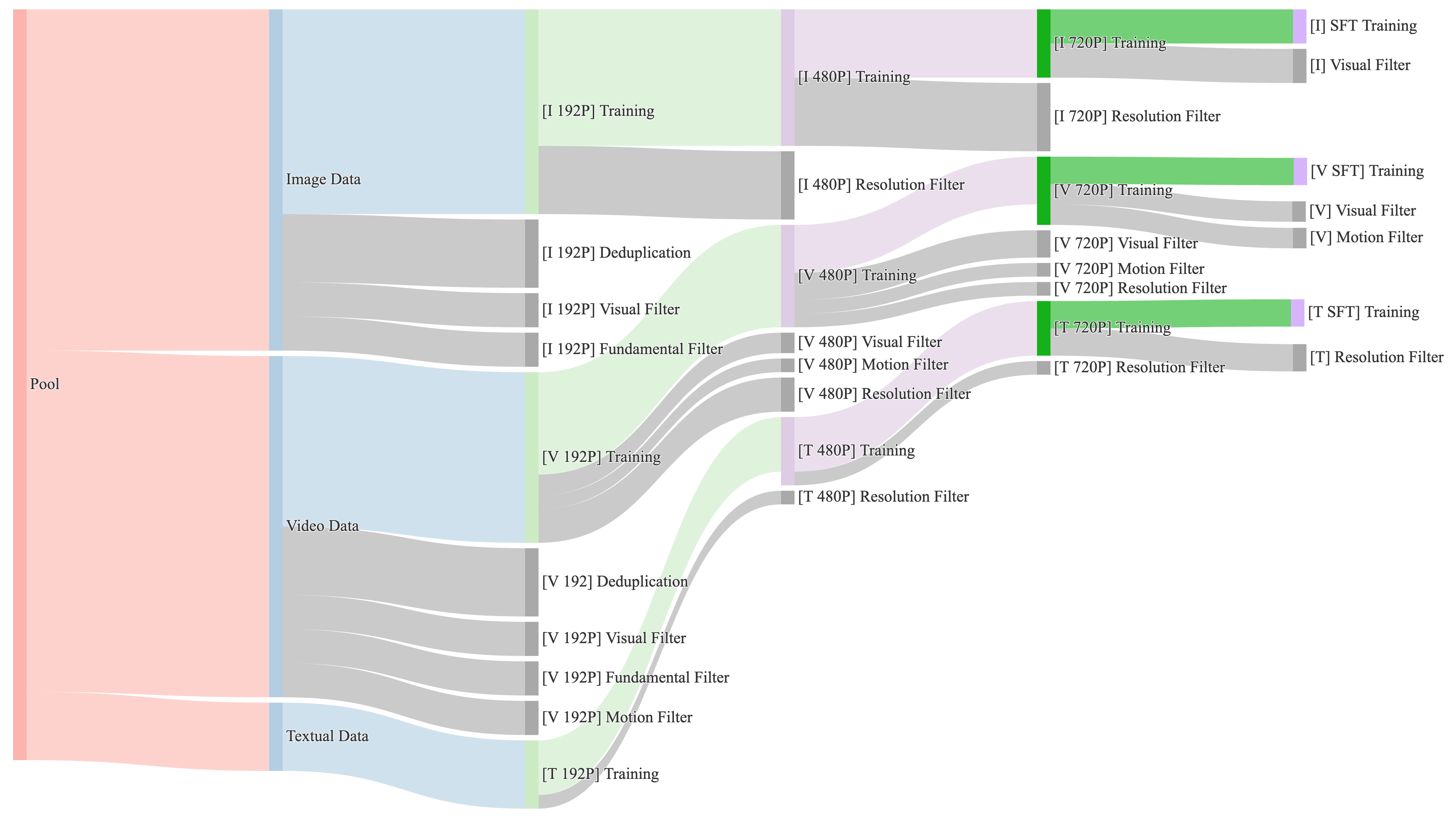}  
    \caption{Data provisioning across different training phases. For each stage, we dynamically adjust the proportions of data related to motion, quality, and category based on data throughput.} 
    \label{fig:data_stage} 
\end{figure}

\subsection{Post-training Data}
\label{sec:post-training data}
The core objective of post-training is to improve both the visual fidelity and motion dynamics of generated videos through high-quality data.
Our data processing pipeline in this stage employs distinct strategies for static and dynamic data: image data undergoes optimization for improved visual quality, while video data is specifically processed to refine motion quality.

\textbf{Image processing.}
From the pool of high-scoring image data, we perform an additional refinement process to identify top-quality samples that the images excel in aspects such as quality, composition and details.
Specifically, we construct the curated dataset using two methods: \emph{expert-based collection} and \emph{manual collection}.
The former method involves selecting the top 20\% of images based on scores predicted by an expert model.
For this subset, we also consider factors such as style and category to ensure diversity in data distribution.
The latter approach involves manually collecting top-quality images from various categories and data sources while also filling in any missing concepts in our dataset to enhance the model's generalization capabilities. 
Through this process, we collected a total of millions of curated images.

\textbf{Video processing.}
In this phase, we employ a strategy similar to the one used for images to collect top-quality videos. 
First, we filter out some top-ranked videos from the candidate datasets using the visual quality classifier. 
Then, based on the motion quality classifier, we select millions of videos featuring simple movements and millions of videos with complex movements. 
All video selections follow a strategy that emphasizes category balance and high diversity.
Meanwhile, we select data from 12 major categories, including technology, animals, arts, humans, vehicles, and \emph{etc}, in order to enhance the model's generative capabilities for commonly used categories.
\subsection{Dense Video Caption}
While our dataset includes text descriptions from original webpages for many images and videos, they are often too simplistic to convey detailed visual content.
DALL-E 3~\citep{BetkerImprovingIG} has demonstrated that the ability of visual generation models to follow prompts can be significantly enhanced by training on highly descriptive generated visual captions.
Therefore, we develop an internal caption model to generate dense captions for each image and video of our dataset.
To train this model, we incorporate both open-source vision-language datasets and additional data collected in-house.

\subsubsection{Open Source Dataset}
\label{sec:open_src_cap_data}
We collect widely used vision-language datasets for both images and videos.
Our collection includes not only caption datasets but also visual Q\&A datasets focused on visual content, such as actions, counting, and OCR.
In some scenarios, we require the caption model to generate captions in a specific style or content based on user instructions.
Therefore, we also collected pure text instruction data to enhance our model's ability to follow instructions.

\subsubsection{In-house Dataset}
\label{sec:in_house_cap_data}
We collect an in-house dataset for various tasks to enhance the model's capabilities in specific areas.

\emph{Celebrities, landmarks, and movie characters.}
To train our model to recognize celebrities, landmarks, and characters, we collect dataset comprising thousands of identities.
First, we use large language models (LLMs) to collect names, then retrieve corresponding images from our image-text database using a CLIP-style model.
We test various CLIP-style models and find that TEAM~\citep{TEAM2022MM} excels in recognizing individuals, particularly Chinese celebrities.
To further reduce noise in our dataset, we implemented keyword matching for the descriptions of the image-text pairs.
An image is retained only if it is retrieved by the TEAM model and matches the specified keywords.

\emph{Object counting.}
To enhance our model's visual counting ability, we compiled a counting dataset by retrieving images with text descriptions that include words like ``one," ``two," ``three", etc.
These numbers provide coarse annotations for the images.
We then use a LLM to extract (\emph{category}, \emph{count}) pairs from the corresponding text.
Finally, we use Grounding DINO~\citep{groundingdino} to count objects in those categories.
An image is retained only if its text description matches the results from Grounding DINO~\citep{groundingdino}.

\emph{OCR.}
To improve our model's capability in generating text descriptions, we collect an OCR-augmented image caption dataset.
Initially, we used an off-the-shelf OCR detector to extract text from images.
Then, we prompt our caption model to generate descriptions based on these OCR results.
This approach ensures accurate text descriptions by using the extracted OCR text as prior knowledge.
Currently, we include only English and Chinese texts.

\emph{Camera angle and motion.}
We observe that current multi-modal language models (MLLMs) struggle with predicting camera angles and motion, including state-of-the-art commercial models like GPT-4o and Google Gemini Pro.
To enhance these predictions, we first annotate a set of videos for camera angle and motion.
This dataset can be utilized in two ways:
directly training our caption model used in the final training stage,
or by training an expert model to annotate more video data, aiding our model during the early training phase.
This dataset enhances the accuracy of our motion and angle descriptions, thereby improving the camera motion controllability of our video generation model.

\emph{Fine-grained categories.}
To enhance our model's ability to recognize fine-grained categories such as animals, plants, and vehicles, we created a dataset containing several million images across these categories.

\emph{Relational understanding.}
We enhance our model’s relational understanding by gathering dataset that focus on spatial relationships, such as left, right, up, and down, through the incorporation of existing object detection datasets.

\emph{Re-caption.}
An important capability of a caption model is to leverage existing tags or brief captions to generate dense captions.
To facilitate this, we curate a re-caption dataset that extends brief text tags or captions to detailed descriptions based on image content.

\emph{Editing instruction caption.}
We collect a dataset that describes the differences or transformations between two images.
This type of caption is valuable for image editing tasks.

\emph{Group image description.}
We collect a dataset that provides descriptions for groups of images.
Each caption begins with a description of the common features shared by the images, followed by individual descriptions of each image, using similar phrasing.

\emph{Human-annotated image and video captions.}
We gather human-annotated dense caption for both images and videos.
These represent our highest-quality caption data and are used in the final stage of model training.

\begin{figure}[htbp]
    \centering  
    \includegraphics[width=0.9\textwidth]{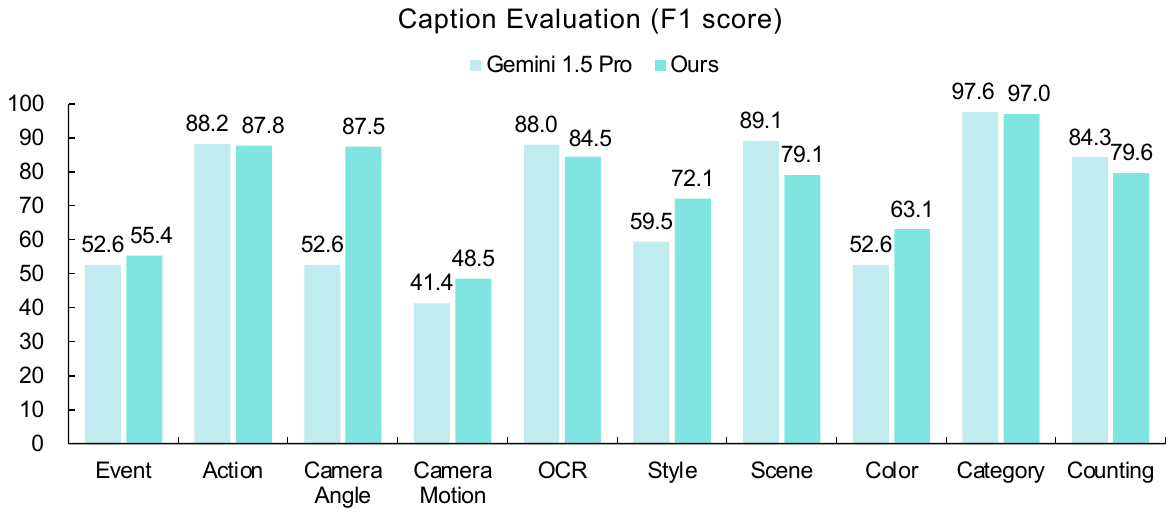}  
    \caption{Our caption model achieves overall performance comparable to Google Gemini 1.5 Pro. In particular, it excels in areas such as events, camera angle, camera motion, style, and color.} 
    \label{fig:cap_eval}
\end{figure}

\subsubsection{Model Design}
\textbf{Architecture.} Our caption model adopts a LLaVA-style architecture~\citep{llava}.
We use a ViT encoder to extract visual embeddings of images and video frames.
These embeddings are projected through a two-layer perception and then fed into the Qwen LLM~\citep{qwen2.5}.

For the image input, we adopt dynamic high resolution as in LLaVA.
An image can be divided into a maximum of seven patches.
For each patch, we adaptively pool the vision embeddings to a 12$\times$12 grid representation to reduce computation.

For the video, we sample 3 frames per second, with an upper limit of 129 frames.
To further reduce computation, we employ a slow-fast encoding strategy: maintaining the original resolution for every fourth frame and applying global average pooling to the embeddings of the remaining frames.
Experiments on long video benchmarks indicate that the slow-fast mechanism enhances understanding with a limited number of visual tokens (\textit{e.g.}, improving performance from 67.6\% to 69.1\% on VideoMME~\citep{videomme}) without subtitles.

\textbf{Training.} Following recent state-of-the-art methods, our training process is divided into three stages.
In the first stage, we freeze the ViT and LLM, and train only the multi-layer perceptron to align the visual embeddings with the LLM input space, using a learning rate of 1e-3.
In the second stage, all parameters are made trainable.
In the final stage, we conduct end-to-end training on a small set of high-quality data.
For the last two stages, the learning rates are set to 1e-5 for the LLM and MLP, and 1e-6 for the ViT.

\subsubsection{Evaluation}
Automatic evaluation is crucial for the development of our caption model, allowing us to identify the strengths and weaknesses of each model version.
Following CAPability~\citep{liu2025good}, we develop an automatic caption evaluation pipeline.
We focus on ten key visual dimensions in video generation: action, camera angle, camera motion, object category, object color, object counting, OCR, scene, style, and event.
For each dimension, we randomly sampled 1,000 videos and their corresponding captions generated by our model.
We also used Google Gemini 1.5 Pro to generate dense video captions for this dataset.
We evaluate our captions and Gemini-generated captions by calculating the F1 metrics for each dimension following CAPability~\citep{liu2025good}.
The results are illustrated in Fig.~\ref{fig:cap_eval}.
Our caption performs better in video event, camera angle, camera motion, style, and object color, while Gemini excels in action, OCR, scene, object category, and object counting.

\section{Model Design and Acceleration}
\label{sec:coding}

\subsection{Spatio-temporal Variational Autoencoder}
\label{sec:vae_arch}

\begin{figure}[th]  
    \centering  
    \includegraphics[width=0.99\textwidth]{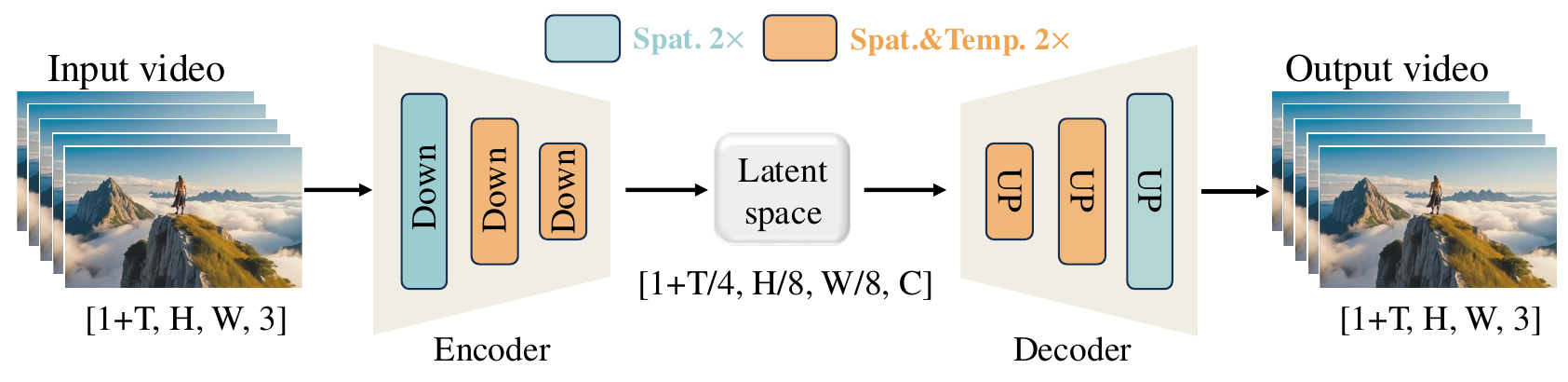}  
    \caption{Our \method-VAE Framework. \method-VAE can compress the spatio-temporal dimension of a video by $4\times8\times8$ times. The orange rectangles represent $2\times$ spatio-temporal compression, and the green rectangles represent $2\times$ spatial compression.} 
    \label{fig:vae_network} %
\end{figure}

Variational Autoencoders (VAEs)~\citep{wu2024improved, blattmann2023svd, openaisora2024}  play a crucial role in learning compact latent representations from high-dimensional visual data (especially videos), facilitating scalable and efficient training of generative models, \emph{e.g.}, diffusion models.
However, designing effective VAEs for video generation tasks faces several challenges. 
First, videos inherently possess both spatial and temporal dimensions, requiring the VAE to capture complex spatio-temporal dependencies. 
Second, the inherent high-dimensional nature of video (\emph{e.g.}, multiple frames with high-resolution pixels) increases memory consumption and computational costs, making it difficult to scale VAEs to long video sequences.
Third, ensuring temporal causality (\emph{i.e.}, future frames do not influence past frames) is critical for generating realistic and coherent video content, yet this constraint introduces additional architectural complexities.

To address these challenges, we propose a novel 3D causal VAE architecture (\method-VAE) specifically designed for video generation. 
We combine multiple strategies~\citep{wu2024improved} to improve spatio-temporal compression, reduce memory usage, and ensure temporal causality.
These enhancements make \method-VAE more efficient and scalable and better suited for integration with diffusion-based generative models such as DiT.
Next, we introduce the network design, training details, efficient inference, and experimental comparison of our \method-VAE.

\subsubsection{Model Design}

To achieve a bidirectional mapping between the high-dimensional pixel space and the low-dimensional latent space, we design a 3D causal VAE as illustrated in Fig.~\ref{fig:vae_network}. 
Given an input video $V\in \mathbb{R}^{(1+T) \times H \times W \times 3}$, \method-VAE compresses its spatio-temporal dimensions to $[1+T/4, H/8, W/8]$ while expanding the number of channels $C$ to 16.
Specifically, the first frame is only spatially compressed to better handle the image data, following MagViT-v2~\citep{yu2023language}.
In terms of architecture, we replace all GroupNorm layers~\citep{wu2018group} with RMSNorm layers~\citep{rms} to preserve temporal causality.
This modification enables the use of the feature cache mechanism (refer to Sec.~\ref{sec:cache_infer}), significantly improving the inference efficiency.
Furthermore, we halve the input feature channel in the spatial upsampling layer, resulting in a 33\% reduction in memory consumption during inference.

By carefully tuning the number of base channels further, \method-VAE achieves a compact model size of only 127M parameters. 
This optimization reduces the encoding time and memory usage, thus benefiting the training of subsequent diffusion transformer models.

\subsubsection{Training}

We adopt a three-stage approach to train \method-VAE. 
First, we construct a 2D image VAE with the same structure and train it on image data. Then, we inflate the well-trained 2D image VAE into a 3D causal \method-VAE to provide an initial spatial compression prior, which drastically improves the training speed compared to training a video VAE from scratch. 
At this stage, \method-VAE is trained on the low-resolution (128$\times$128) and small frame number (5-frame) videos to accelerate convergence speed. 
The training losses include $L1$ reconstruction loss, KL loss, and LPIPS perceptual loss, which are weighted by coefficients of 3, 3e-6, and 3, respectively.
Finally, we fine-tune the model on high-quality videos with different resolutions and frame numbers, integrating the GAN loss~\citep{esser2020taming} from a 3D discriminator.

\begin{figure}[t]  
    \centering  
    \includegraphics[width=0.99\textwidth]{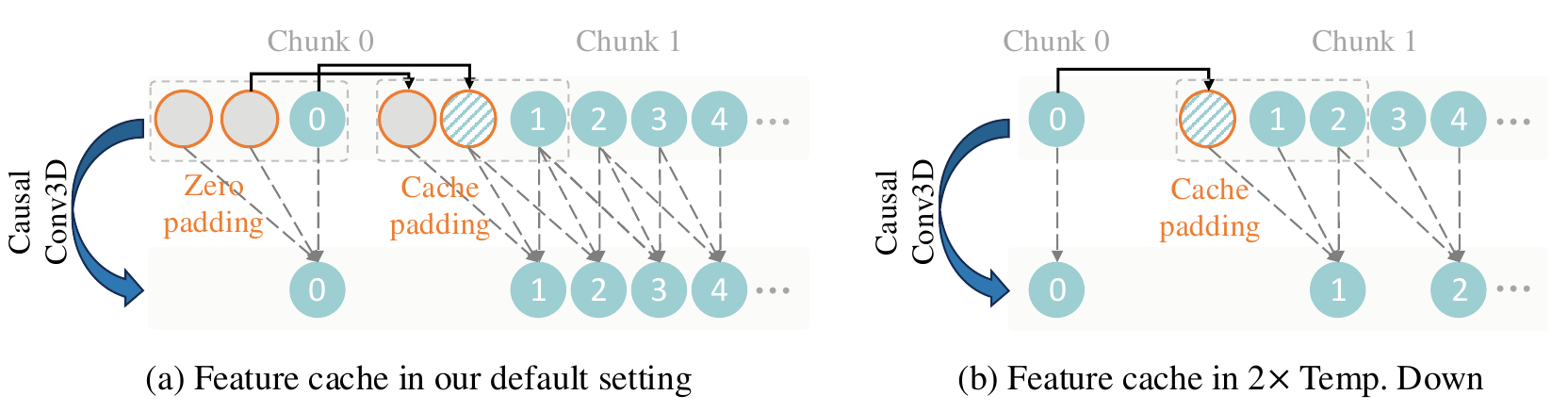}  
    \caption{Our feature cache mechanism. (a) and (b) show how we use this mechanism in regular causal convolution and temporal downsampling, respectively.} 
    \label{fig:vae_cache} %
\end{figure}

\subsubsection{Efficient Inference} \label{sec:cache_infer}

To efficiently support the encoding and decoding of arbitrarily long videos~\citep{CDT,yao2021wenet}, we implement a feature cache mechanism within the causal convolution module of \method-VAE. 

Specifically, the number of video sequence frames follows the $1+T$ input format, so we divide the video into $1+T/4$ chunks, which is consistent with the number of latent features.
When processing input video sequences, the model employs a chunk-wise strategy where each encoding and decoding operation handles only the video chunk corresponding to a single latent representation.
Based on the temporal compression ratio, the number of frames in each processing chunk is limited to at most 4, effectively preventing memory overflow.

To ensure temporal continuity between context chunks, our model strategically maintains frame-level feature caches from preceding chunks.
These cached features are systematically integrated into the causal convolution computations of subsequent chunks.
The specific process is illustrated in Fig.~\ref{fig:vae_cache}, which demonstrates two typical scenarios in our feature cache mechanism. 
In Fig.~\ref{fig:vae_cache} (a), in our default setting, the causal convolution does not change the number of frames, and we need to maintain two cached features (convolution kernel size is 3) from historical frames. 
For the initial chunk, we apply zero padding with two dummy frames to initialize the cache buffer. 
Subsequent chunks systematically reuse the last two frames from the preceding chunk as cached features while discarding obsolete historical data.
Fig.~\ref{fig:vae_cache} (b) presents the scenario involving 2$\times$ temporal downsampling (stride is equal to 2). 
Here, the scenario necessitates different cache management: we implement single-frame cache padding exclusively for non-initial chunks to ensure dimensional consistency. This configuration guarantees that the output sequence length follows the exact downsampling ratio while maintaining causal relationships across the boundaries of the chunk.

This feature cache mechanism not only optimizes memory utilization but also preserves feature coherence across chunk boundaries, thereby supporting stable inference for infinite-length videos.

\subsubsection{Evaluation}

\begin{figure}[tbp]  
    \centering  
    \includegraphics[width=0.82\textwidth]{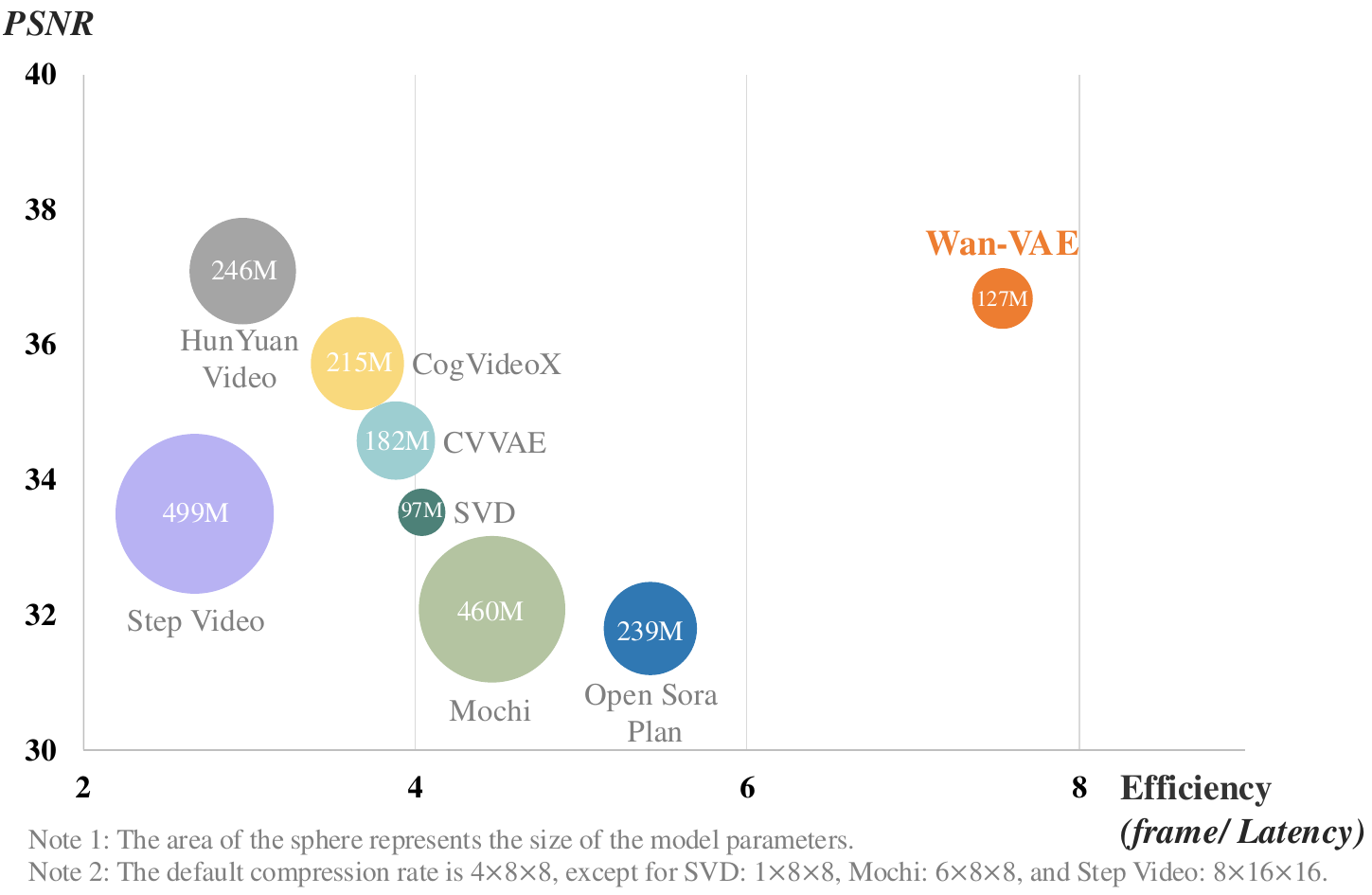}  
   \caption{Comparison of video reconstruction performance at $720\times720$ resolution and 25 frames.} 
    \label{fig:vae_psnr}  
\end{figure}

\textbf{Quantitative results.}
This study conducts a comprehensive evaluation of the performance of various state-of-the-art (SOTA)  video VAE models~\citep{kong2024hunyuanvideo,cogvideox} by analyzing their Peak Signal-to-Noise Ratio (PSNR) and processing efficiency (frame per unit latency (second)). 
Notably, for fair comparison, most models~\citep{kong2024hunyuanvideo,cogvideox} adopt the same compression rate and latent feature dimension as our approach, \textit{i.e.,} a compression rate of $4\times8\times8$ with a latent dimension of 16.
Open Sora Plan~\citep{yuan_opensora}  has the same compression ratio as ours, but the latent dimension is 4.
SVD~\citep{blattmann2023svd} employs a compression rate of $1\times8\times8$ with a latent dimension of 4,
Step Video~\citep{step_video} employs a compression rate of $8\times16\times16$ with a latent dimension of 64, 
and Mochi~\citep{genmo2024mochi} uses a compression rate of $6\times8\times8$ with a latent dimension of 12.
The evaluation is performed on 200 videos.
These videos consist of 25 frames, and the resolution is set to $720 \times 720$.
In Fig.~\ref{fig:vae_psnr}, the size of the circles is positively correlated with the number of model parameters.
 
Experimental results indicate that \method-VAE exhibits highly competitive performance across both metrics, showcasing a compelling dual advantage of superior video quality and high processing efficiency. 
It is worth noting that under the same hardware environment, our VAE's reconstruction speed is 2.5 times faster than the existing SOTA method (\textit{i.e.}, HunYuan Video). 
This speed advantage will be further demonstrated at higher resolutions due to the small size design of our VAE model and the feature cache mechanism.

This advancement offers an efficient solution for video reconstruction tasks and video generation training. 
These findings not only validate the effectiveness of the proposed model design but also provide valuable insights for the future development of VAE technologies.

\begin{figure}[tbp]  
    \centering  
    \includegraphics[width=0.93\textwidth]{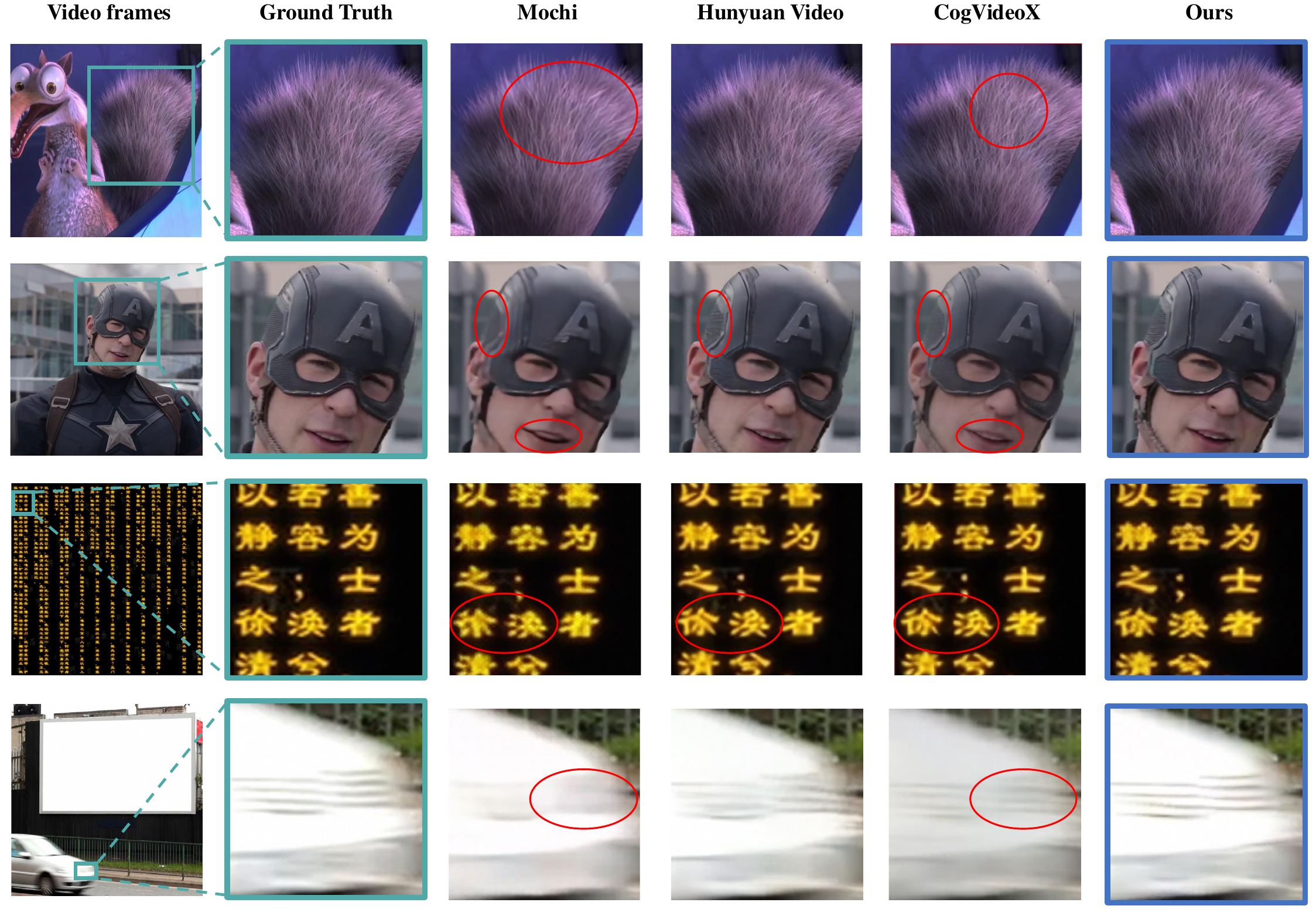}  
    \caption{Visualization results of video reconstruction across different scenarios, including texture (first row),  face (second row), text (third row), and high-motion (fourth row). The above videos have rich details but are not used in training.} 
    \label{fig:vae_visual}  
\end{figure}

\textbf{Qualitative results.}
 To validate the video reconstruction performance of our method across diverse scenarios, Fig.~\ref{fig:vae_visual}  presents visual results for texture, face, text, and high-motion scenes. 
 Compared to existing VAE models, \method-VAE demonstrates superior performance in these contexts. 
 In the texture scene shown in the first row, our method is capable of capturing details more accurately, such as the texture and direction of hair. 
 In the facial scene depicted in the second row, \method-VAE preserves facial features while reducing blurring and distortion around the lips. 
 In the text scene illustrated in the third row, our method successfully restores textual content with clarity, mitigating issues of character distortion and loss. 
 In the high-motion scene presented in the fourth row, \method-VAE maintains the motion sharpness of video frames. 
 These results indicate that our method offers significant advantages in handling a variety of complex scenarios.

\begin{figure}[!t]  
    \centering  
    \includegraphics[width=1.0\textwidth]{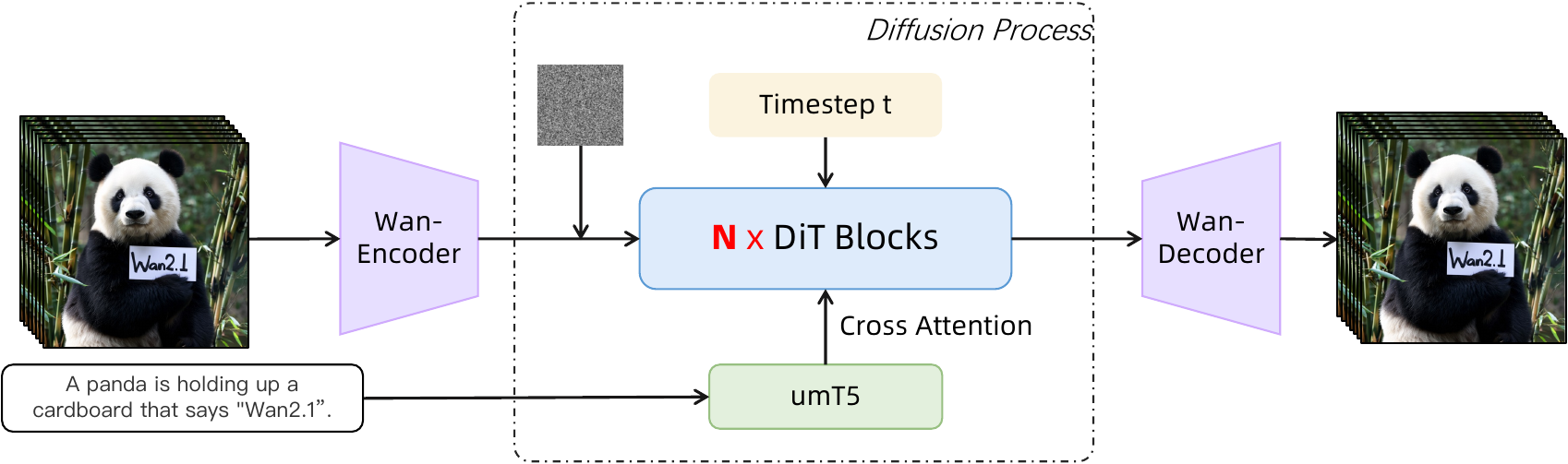}  
    \caption{Architecture of the \method.} 
    \label{fig:overview_network} %
\end{figure}

\subsection{Model Training}
\label{sec:model_training}

In this section, we will provide a detailed introduction to the architecture design of the foundational video model, specifically for the text-to-video task, along with the details of the pre-training and post-training phases.

In Fig.~\ref{fig:overview_network}, we present the architecture of \method, which is designed based on the current mainstream DiT~\citep{dit} architecture, which typically consists of three primary components: \method-VAE, diffusion transformer, and text encoder.
For a given video $V\in \mathbb{R}^{(1+T) \times H \times W \times 3}$, the encoder of \method-VAE encodes it from pixel space into the latent space $x \in \mathbb{R}^{1+T/4 \times H/8 \times W/8}$, which is then fed into the subsequent diffusion transformer structure.

\subsubsection{Video Diffusion Transformer}
\label{sec:dit}

\begin{wrapfigure}{r}{0.4\linewidth}
    \vspace{-0.2em}
    \includegraphics[width=0.24\textheight]{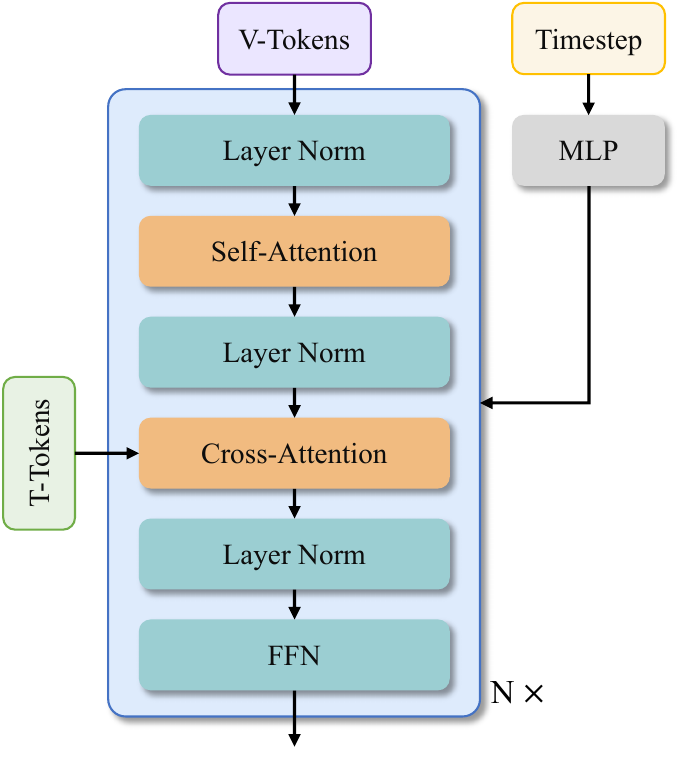}
    \caption{Transformer block of \method.}
    \label{fig:dit_block}
\end{wrapfigure}
\textbf{Diffusion transformer}. The diffusion transformer mainly consists of three components: a patchifying module, transformer blocks, and an unpatchifying module.
Within each block, we focus on effectively modeling spatio-temporal contextual relationships and embedding text conditions alongside time steps.
In the patchifying module, we use a 3D convolution with a kernel size of $(1, 2, 2)$ and apply a flattening operation to convert $x$ into a sequence of features with the shape of $(B, L, D)$, where $B$ denotes the batch size, $L=(1+T/4) \times H/16 \times W/16$ represents the sequence length, and $D$ indicates the latent dimension.
Specifically, as illustrated in Fig.~\ref{fig:dit_block}, we employ the cross-attention mechanism to embed the input text conditions, which can ensure the model's ability to follow instructions even under long-context modeling.
Additionally, we employ an MLP with a Linear layer and a SiLU~\citep{SiLU} layer to process the input time embeddings and predict six modulation parameters individually. 
This MLP is shared across all transformer blocks, with each block learning a distinct set of biases.
Through extensive experiments, we have demonstrated that this design can reduce the parameter count by approximately $25\%$ and reveal a significant performance improvement with this approach at the same parameter scale.

\textbf{Text encoder}. \method's architecture uses the umT5~\citep{chung2023unimax} to encode input text.
Through extensive experiments, we find that umT5 has several advantages in our framework: 
1) It possesses strong multilingual encoding capabilities, allowing it to understand both Chinese and English effectively, as well as input visual text; 
2) Under the same conditions, we find that umT5 outperforms other unidirectional attention mechanism LLMs in composition;
3) It exhibits superior convergence, with umT5 achieving faster convergence at the same parameter scale. 
Based on these findings, we ultimately chose umT5 as our text embedder.

\subsubsection{Pre-training}
We leverage the flow matching framework~\citep{flow-matching, esser2024sd3}  to model a unified denoising diffusion process across both image and video domains. 
We first conduct pre-training on low-resolution images, followed by multi-stage joint optimization of images and videos. 
The image-video joint training evolves using progressively upscaled data resolutions and extended temporal durations throughout the stages.

\noindent\textbf{Training objective.}
Flow matching provides a theoretically grounded framework for learning continuous-time generative processes in diffusion models. 
It circumvents iterative velocity prediction, enabling stable training via ordinary differential equations (ODEs) while maintaining equivalence to maximum likelihood objectives.
During training, given an image or video latent $x_{1}$, a random noise $x_{0} \sim \mathcal{N}(0, I)$, and a timestep $t \in[0,1]$ sampled from a logit-normal distribution, an intermediate latent $x_{t}$ is obtained as the training input.
Following Rectified Flows (RFs)~\citep{esser2024sd3}, $x_{t}$ is defined as a linear interpolation between $x_{0}$ and $x_{1}$, \ie
\begin{equation}\label{x_t}
    x_{t}=t x_{1}+(1-t) x_{0}.
\end{equation}
The groud truth velocity $v_{t}$ is
\begin{equation}\label{v_t}
v_{t}=\frac{d x_{t}}{d t}=x_{1}-x_{0}.
\end{equation}
The model is trained to predict the velocity, thus, the loss function can be formulated as the mean squared error (MSE) between the model output and $v_{t}$,

\begin{equation}\label{rf_loss}
\mathcal{L} = \mathbb{E}_{x_0, x_1, c_{txt}, t}||u(x_t, c_{txt}, t; \theta)-v_t||^2,
\end{equation}
where $c_{txt}$ is the umT5 text embedding sequence of 512 tokens long, $\theta$ is the model weights, and $u(x_t, c_{txt}, t; \theta)$ denotes the output velocity predicted by the model.

\noindent\textbf{Image pre-training.}
Our experimental analysis reveals two critical challenges in direct joint training with high-resolution images and long-duration video sequences: 
(1)~Extended sequence lengths (typically $81$ frames for $1280\times720$ video) substantially reduce training throughput. Under fixed GPU-hour budgets, this results in insufficient data throughput, thereby impeding model convergence rates. 
(2)~Excessive GPU memory consumption forces suboptimal batch sizes, inducing training instability caused by spikes in gradient variance.
To mitigate these issues, we initialize the 14B model training through low-resolution ($256~px$) text-to-image pre-training, enforcing cross-modal semantic-textual alignment and geometric structure fidelity before progressively introducing high-res video modalities.

\noindent\textbf{Image-video joint training.}
Following large-scale $256~px$ text-to-image pre-training, we implement staged joint-training with image and video data through a resolution-progressive curriculum. 
The training protocol comprises three distinct stages differentiated by spatial resolution areas:
(1)~In the first stage, we conduct joint training with $256~px$ resolution images and 5-second video clips ($192~px$ resolution, $16$ fps).
(2)~In the second stage, we initiate spatial resolution scaling by upgrading both image and video resolutions to $480px$ while maintaining a fixed 5-second video duration.
(3)~In the final phase, we escalate spatial resolution to $720 px$ for both images and 5-second video clips.

\noindent\textbf{Pre-training configurations.}
We employ efficient training at bf16-mixed precision combined with the AdamW~\citep{adamw, adam} optimizer with a weight decay coefficient of $1e^{-3}$.
The initial learning rate is set to $1e^{-4}$, and dynamically reduced triggered by plateaus of the FID and CLIP Score metrics.

\subsubsection{Post-training}

In the post-training stage, we maintain the same model architecture and optimizer configuration from the pre-training stage, initializing the network with the pre-trained checkpoint. We conduct joint training at resolutions of $480 px$ and $720 px$ using the post-training video dataset detailed in Sec.~\ref{sec:post-training data}.
\subsection{Model Scaling and Training Efficiency}
\label{sec:train}

\subsubsection{Workload Analysis}

In \method, the computational costs associated with the text encoder and VAE encoder are lower than those of the DiT model, which accounts for more than 85\% of the overall computation during training. For the DiT model, the main computational cost is given by the expression $L(\alpha bsh^2+\beta bs^2h)$, where $L$ denotes the number of DiT layers, $b$ is the micro batch size, $s$ indicates the sequence length, and $h$ denotes the hidden dimension size. Here, $\alpha$ corresponds to the cost of the linear layers and $\beta$ to that of the attention layer, for the non-causal attention used in \method, $\beta$ is 4 for the forward pass and 8 for the backward pass. Unlike general LLM models, the sequence length $s$ processed by \method often reaches hundreds of thousands or more. Since the computational cost of attention increases quadratically with the number of tokens, while the cost associated with the linear layers increases only linearly, the attention mechanism gradually becomes the training bottleneck. In scenarios where the sequence length reaches 1 million, the computation time for attention can account for up to 95\% of the end-to-end training time.

During the training process, only the DiT model is optimized while the text encoder and the VAE encoder remain frozen. Therefore, GPU memory usage is concentrated on training the DiT. The GPU memory usage for the DiT can be expressed as $\gamma Lbsh$, where $\gamma$ depends on the implementation of the DiT layers, L denotes the number of DiT layers. Because video tokens, input prompts, and timesteps are taken as model input, the value of $\gamma$ is generally larger than in ordinary LLM models. For example, while standard LLMs might have a $\gamma$ of 34~\citep{korthikanti2022megatronsp}, the DiT model's $\gamma$ can exceed 60.
Consequently, when the sequence length reaches 1 million tokens and the micro-batch size is 1, the total GPU memory usage for activations in a 14B DiT model can exceed 8 TB.

In summary, the primary computational cost in \method arises from the attention mechanism.
While the computational cost increases quadratically with the sequence length $s$, the GPU memory usage scales only linearly with $s$. This provides a valuable reference for guiding subsequent optimization efforts.

\subsubsection{Parallelism Strategy}
\label{4d_parallel}
The \method model mainly consists of three modules: VAE, text encoder, and DiT. Following the cache mechanism optimization outlined in Sec.~\ref{sec:cache_infer}, the VAE module exhibits minimal GPU memory usage and can seamlessly adopt Data Parallelism (DP). In contrast, the text encoder requires over 20 GB of GPU memory, necessitating the use of model weight sharding to conserve memory for the subsequent DiT module.
To address this, we employ a Data Parallel (DP) approach combined with Fully Sharded Data Parallel (FSDP) ~\citep{zhao2023pytorchfsdp} strategy.
The model parameters, gradients, and optimizer states of the DiT will exceed the capacity of a single GPU, with activation storage reaching approximately 8 TB in a scenario involving 1M tokens with a batch size of 1. Moreover, the DiT portion accounts for a significant proportion of the overall computational workload. Thus, we focus on developing efficient distributed strategies specifically for the DiT module.

\noindent\textbf{DiT parallel strategy.}
In light of the DiT model’s modest size, and considering the optimization of overlapping computation and communication as well as maximizing the size of data parallelism (DP), we have selected FSDP as our parameter sharding strategy.

Regarding activations, the input shape of a DiT block is $[b, s, h]$, where the $b$ dimension corresponds to data parallelism, $s$ represents the sequence length, and $h$ denotes the hidden dimension. Therefore, we need to examine sharding strategies for the $s$ and $h$ dimensions. Sharding along the s dimension is achieved through Context Parallelism (CP), with common methods including Ulysses~\citep{jacobs2023deepspeedulysses} and Ring Attention~\citep{liu2023ringattention}. Sharding along the $h$ dimension primarily involves Megatron’s tensor parallelism (TP)~\citep{megatron} combined with sequence parallelism (SP)~\citep{korthikanti2022megatronsp}, which shards the hidden dimension of the activations by splitting the weights.
Since the communication overhead of CP is smaller than that of (TP $+$ SP), we propose using CP to accelerate the batch-level computations while reducing the activation memory footprint on each GPU.
We have designed a two-dimensional (2D) CP that combines the characteristics of Ulysses and Ring Attention, similar to USP~\citep{fang2024usp}. In this design, the outer layer employs Ring Attention, and the inner layer leverages Ulysses. This combination mitigates the slow cross-machine communication inherent in Ulysses and addresses the need for large block sizes after sharding in Ring Attention, thereby maximizing the overlap between outer-layer communication and inner-layer computation. In scenarios with a sequence length of 256K and 16 GPUs across 2 machines, the communication overhead of 2D Context Parallelism is reduced from over $10\%$ with Ulysses to below $1\%$.

\begin{figure}[t]
    \scriptsize
    \centering    \includegraphics[width=1\linewidth]{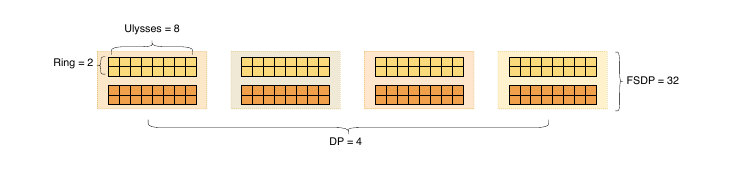}
        \vspace{-8mm}
    \caption{Illustration of DiT parallelism. Assuming a total of 128 GPUs, the innermost layer consists of a combination of Ulysses=8 and Ring=2. The outer layer employs FSDP=32, while the outermost layer utilizes DP=4. The global batch size is 8 times that of the micro-batch size.}
    \label{fig:dit_parallel}
    \vspace{-4mm}

\end{figure}

The FSDP group and CP group intersect within the FSDP group, the DP size equals the FSDP size divided by the CP size.
After satisfying memory and single-batch latency requirements, we use DP for the scaling. 
Fig.~\ref{fig:dit_parallel} illustrates the distributed scheme for the DiT module. For example, in a configuration with 128 GPUs, the CP size is 16, with Ulysses set to 8 and Ring Attention to 2; FSDP is set to 32, corresponding to a batch size of 2b; DP is set to 4, resulting in a global batch size of 8b.

\noindent\textbf{Distributed strategy switching for different modules.}
During training, different modules utilize the same resources. Specifically, we apply DP to the VAE and Text Encoder, while the DiT module employs a combination of DP and CP. Therefore, when the outputs of the VAE and Text Encoder are fed into the DiT module during training, it is necessary to switch distributed strategies to avoid resource wastage.
Specifically, CP requires that devices within the CP group read the same batch of data. Yet, if devices in the VAE and Text Encoder sections also read the same data, this leads to redundant computations. To eliminate this redundancy, devices within the CP group initially read different data. Then, before CP, we perform a loop traversal of size equal to the CP size, sequentially broadcasting the data read by different devices within the CP group to ensure that the inputs within CP are identical. In this way, the time proportion of VAE and Text Encoder within a single model iteration is reduced to $1 / CP$, thereby enhancing overall performance.

\subsubsection{Memory Optimization}

As introduced, the computational cost in \method~grows quadratically with sequence length $s$, whereas GPU memory usage scales linearly with $s$. This means that in long sequence scenarios, computation time can eventually exceed the PCIe transfer time required for offloading activations. Specifically, the PCIe transfer time for offloading the activation of one DiT layer can be overlapped with the computation of just 1 to 3 DiT layers. Compared to gradient checkpointing (GC) strategies~\citep{chen2016training}, activation offloading~\citep{rhu2016vdnn} that allows computation overlap provides an effective way to reduce GPU memory usage without sacrificing end-to-end performance. Therefore, we prioritize activation offloading to reduce GPU memory.
Since CPU memory is also prone to exhaustion in long sequence scenarios, we combine offloading with GC strategies, prioritizing gradient checkpointing for layers with high GPU memory-to-computation ratios.

\subsubsection{Cluster Reliability}

By leveraging Alibaba Cloud's advanced intelligent scheduling, slow machine detection, and robust self-healing capabilities within the training cluster, high stability is ensured. Hardware issues are identified during the job startup phase, ensuring that only healthy nodes are allocated to training tasks.
Throughout training, any faulty nodes are promptly isolated and repaired, with tasks automatically restarted and training seamlessly resumed.
This efficient orchestration effectively combines reliability with high performance, ensuring overall system stability.
\subsection{Inference}
\label{sec:inference_1}

The primary objective of inference optimization is to minimize the latency of video generation. Unlike training, the inference process involves multiple sampling steps, typically around 50. Accordingly, we employ techniques such as quantization and distributed computing to reduce the time required for each individual step. Additionally, we leverage the attention similarities between steps to decrease the overall computational load. Furthermore, for classifier-free guidance (CFG)~\citep{ho2022classifierfreediffusionguidance}, we exploit its inherent similarities to further minimize computational requirements.

\subsubsection{Parallel Strategy}

\begin{wrapfigure}{r}{0.5\linewidth}
    \vspace{-1.2em}
    \includegraphics[width=0.3\textheight]{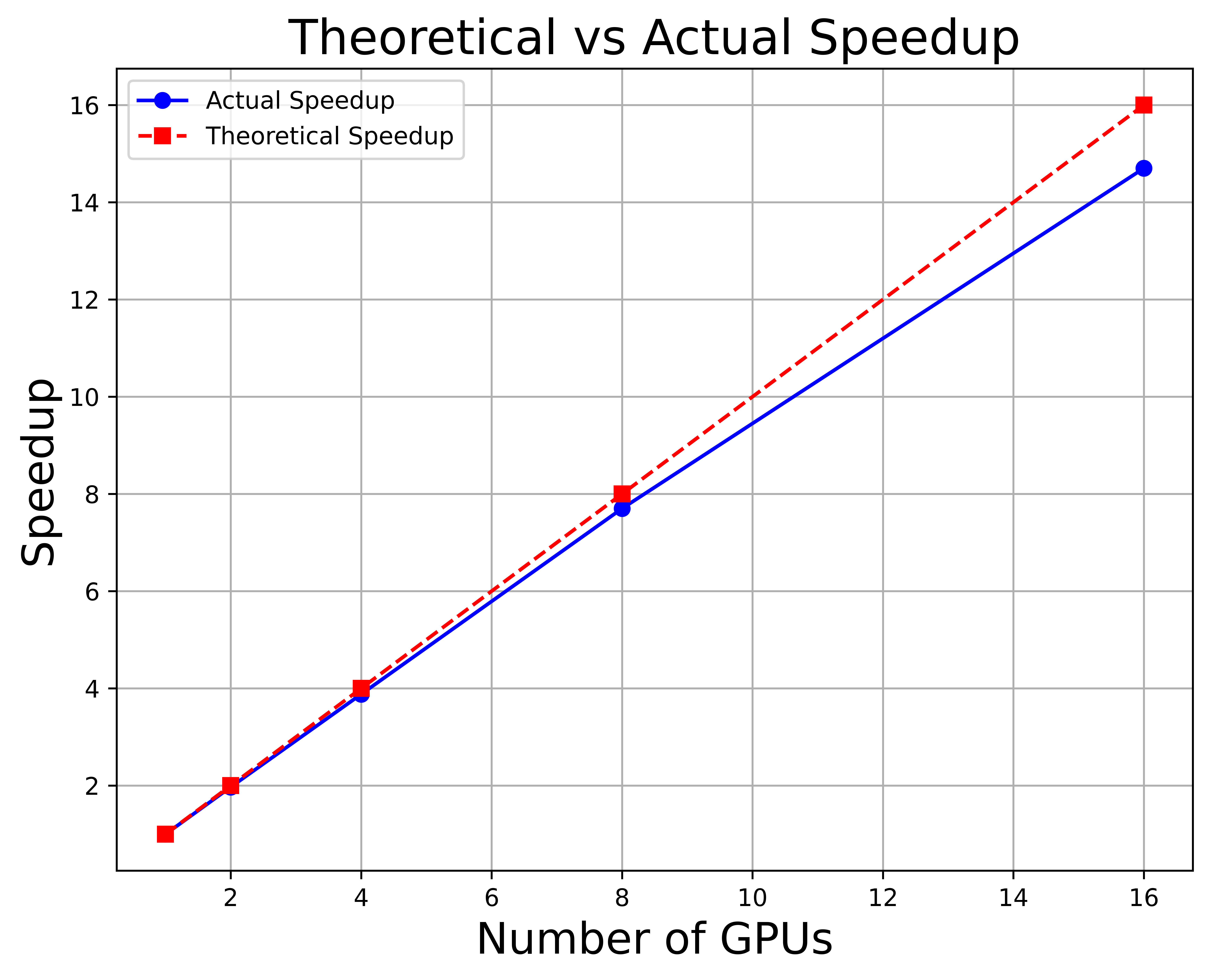}
    \vspace{-0.4em}
    \caption{Scaling inference via multiple GPUs.}
    \label{fig:scaling}
    \vspace{-0.4em}
\end{wrapfigure}
As discussed in Sec.~\ref{4d_parallel}, we utilize Context Parallel to reduce the latency of generating a single video when scaling to multiple GPUs. Additionally, for large models like \method 14B, we adopt model sharding to mitigate GPU memory constraints.
\emph{Model Sharding Strategy}:  Given the long sequence lengths during inference, FSDP incurs less communication overhead compared to TP and allows for computation overlap. Therefore, we adopt the FSDP method for model sharding, consistent with our training approach.
\emph{Context Parallel Strategy}: We employ the same 2D Context Parallelism as that in the training stage, with RingAttention serving as the outer loop and Ulysses operating as the inner loop. Equipped with 2D Context Parallel and FSDP parallel strategy, DiT achieves nearly linear speedup on the \method 14B model, as shown in the Fig.~\ref{fig:scaling}.

\subsubsection{Diffusion Cache}
We conduct a thorough analysis of the \method model and identify the following characteristics in its inference process:
\begin{itemize}
    \item Attention similarity: Within the same DiT block, attention outputs across different sampling steps exhibit significant similarity.
    \item CFG similarity: In the later stages of sampling, there is a notable similarity between conditional and unconditional DiT outputs.
\end{itemize}

These findings are also highlighted in recent works, such as DiTFastAttn~\citep{yuan2024ditfastattn} and FasterCache~\citep{lv2024fastercache}.
We leverage these characteristics to implement diffusion caching to reduce the computational workload.
In practice, we tailor the diffusion cache to the \method model to ensure lossless performance. Specifically, we use the validation set to select certain steps for caching:
\begin{itemize}
    \item Attention cache: For the attention component, we perform an attention forward pass every few steps and cache the results, reusing the cached results for the other steps.
    \item CFG cache: For the unconditional part, we perform a DiT forward pass every few steps and reuse the conditional results for the other steps. Additionally, we apply residual compensation methods similar to FasterCache to prevent the degradation of fine details.
\end{itemize}
After applying diffusion caching to the \method 14B text-to-video model, we have improved the inference performance by 1.62×.

\subsubsection{Quantization}

\noindent\textbf{FP8 GEMM.} We found that using FP8 precision for the GEMM operations, combined with per-tensor quantization on the weights and per-token quantization on the activations in the sampling steps, incurs minimal performance loss. Thus we apply FP8 quantization to all GEMM operations in the DiT block.
The FP8 GEMM delivers twice the performance of BF16 GEMM and achieves a 1.13× speedup in the DiT module.

\noindent\textbf{8-Bit FlashAttention.} Although FlashAttention3~\citep{shah2025flashattention} achieves high performance, its native FP8 implementation suffers from significant quality degradation in video generation, as observed in our experiments.
Conversely, SageAttention~\citep{zhang2024sageattention} employs a mixed precision of int8 and fp16 to reduce precision loss, but it does not provide specific adaptations for Hopper GPUs. As of this report's publication, subsequent optimizations to SageAttention have demonstrated improved Hopper compatibility.
Future work will explore the integration of these performance enhancements.

We present our optimizations applied to FA3-FP8 implementation, focusing on both numerical stability and computational efficiency.

\textbf{Accuracy.} To mitigate the numerical errors in attention observed in \method under FA3-FP8 quantization, we apply two primary techniques:

\begin{itemize}
    \item Mixed 8-Bit optimization. The native implementation of FA3 uses FP8 (E4M3) for all input tensors (\(Q\), \(K\), \(V\)). Based on both our analysis of the data distribution and the empirical evidence from the SageAttention, we adopt a hybrid quantization strategy: Int8 for \(S=QK^T\), FP8 for \(O=PV\).
    \item FP32 accumulation for cross-block reduction. The native FP8 WGMMA implementation employs $~$14-bit accumulators, making it prone to overflow during long-sequence processing. To address this limitation, our approach leverages FP8 WGMMA for intra-block \(PV\) reduction while implementing cross-block accumulation through CUDA cores with Float32 registers, a strategy inspired by DeepSeek-V3’s~\citep{liu2024deepseek} FP8 GEMM methodology.
\end{itemize}

\textbf{Performance.} To maintain kernel performance under mixed INT8-FP8 quantization while enhancing the accumulation precision of \(O=PV\), we employ two techniques:
\begin{itemize}
    \item Fuse Float32 accumulation with intra-warpgroup pipelining. FlashAttention3 employs an intra-warpgroup pipeline that overlaps WGMMA with softmax and scaling. To mitigate the performance penalty introduced by Float32 \(PV\) accumulation across blocks, we fuse the Float32 accumulation with \(rescale\_o\) operation.
    \item Block size tuning. Due to the additional register pressure introduced by Float32 \(PV\) accumulation register across blocks, we adjust the block size to reduce register spilling. After optimization, our 8-bit FlashAttention achieves 95\% MFU on NVIDIA H20 GPUs.
\end{itemize}

Consequently, with 8-bit FlashAttention, we boost the inference efficiency by more than 1.27×.

\begin{table*}[!t]
    \centering
    \begin{tabular}{p{0.3\linewidth} p{0.64\linewidth}}
        \toprule
        User prompts & Rewritten results \\
        \midrule
        A Viking warrior wields a great axe with both hands, battling a mammoth at dusk, amidst a snowy landscape with snowflakes swirling in the air. & An epic battle scene, unfolds as a tall and muscular Viking warrior wields a heavy great axe with both hands, facing off against a massive mammoth. The warrior is clad in leather armor and a horned helmet, with prominent muscles and a fierce, determined expression. The mammoth is covered in long hair, with sharp tusks, and roars angrily. It is dusk, and the snowy landscape is filled with swirling snowflakes, creating an intense and dramatic atmosphere. The backdrop features a barren ice field with the faint outlines of distant mountains. The use of cool-toned lighting emphasizes strength and bravery. The scene is captured in a dynamic close-up shot from a high-angle perspective. \\
        \midrule
        The camera follows a motorboat chasing dolphins in the sea. & In a documentary photography style, the camera follows a motorboat chasing a pod of dolphins leaping out of the vast ocean. On the motorboat, there is a driver wearing a life jacket and a safety helmet, with a focused and excited expression. The dolphins are sleek and smooth-skinned, occasionally leaping out of the water and diving back in with agility. The sky is bright blue, the sun is shining, and the sea is shimmering with light. A few white clouds dot the distant sky. The scene is dynamic and full of energy, captured from a medium shot in a tracking perspective. \\
        \midrule
        The tiny Potato King, wearing a majestic crown, sits on the throne as potato subjects pay homage to it. & In a surrealist style, the tiny Potato King wears a magnificent gold crown and sits on a towering throne. Its skin has a natural earthy yellow tone with subtle bumpy textures. The potato subjects are lined up on either side, bowing their heads in homage to the king. The background features the grand interior of a palace, with gold and red decorations that appear luxurious and solemn. A beam of light shines down from above, creating a sacred atmosphere. The scene is captured in a close-up shot from a high-angle perspective. \\
        \bottomrule
    \end{tabular}
    \caption{Examples of user prompts and their rewritten results. These examples are in Chinese originally and have been translated into English for the purpose of presentation.}
    \label{tab:rewritten_prompts}
\end{table*}

\subsection{Prompt Alignment}
\label{sec:alignment}
Prompt alignment is designed to boost the effectiveness of model inference by aligning the user's input prompt with the format and style of captions used during training.
We employ two primary strategies in this process.
First, we augment each video with a variety of captions, thereby increasing diversity.
Second, we rewrite user prompts to align them with the distribution of video captions from the training stage. In the following, we detail these strategies.

To accommodate diverse user inputs, each image or video is paired with multiple captions reflecting various styles and lengths. For example, we use captions of differing lengths (e.g., long, medium, and short) and styles (e.g., formal, informal, and poetic) to create a varied set of sample-caption pairs for our training data. These captions establish different mapping relations between text and video, effectively covering the range of prompts that users might provide.

Despite the diversity of captions, a distribution mismatch often exists between the dataset captions and user input prompts. 
Typically, users prefer concise prompts consisting of a few simple words, which are significantly shorter than the training captions. This discrepancy can adversely affect the quality of the generated videos. 
To address this issue, we rewrite prompts by utilizing a large language model (LLM) to refine user prompts. 
Our primary objective is to align the distribution of these refined prompts with the distribution of the training captions, thereby achieving improved inference performance.
The LLM is guided by the following principles when rewriting user prompts:
\begin{enumerate}
    \item We instruct the LLM to add details to prompts without altering their original meanings, enhancing the completeness and visual appeal of the generated scenes.
    \item The rewritten prompts should incorporate natural motion attributes, where we add appropriate actions for the subject based on its category to ensure smoother and more fluent motion in the generated videos.
    \item We recommend structuring the rewritten prompts similarly to the post-training captions, beginning with the video style, followed by an abstract of the content, and concluding with a detailed description. This method helps align prompts with the distribution of high-quality video captions.
\end{enumerate}

We evaluate the effectiveness of LLM-assisted prompt rewriting on our benchmark, and several results are presented in Tab.~\ref{tab:rewritten_prompts}.
Our findings suggest that LLMs with strong instruction-following capabilities can generate suitable prompts that enhance video generation. To balance speed and performance, we use Qwen2.5-Plus as our rewriting model.

\begin{figure}[!t]
    \centering
    \begin{minipage}{0.47\linewidth}
        \centering
        \includegraphics[width=\linewidth]{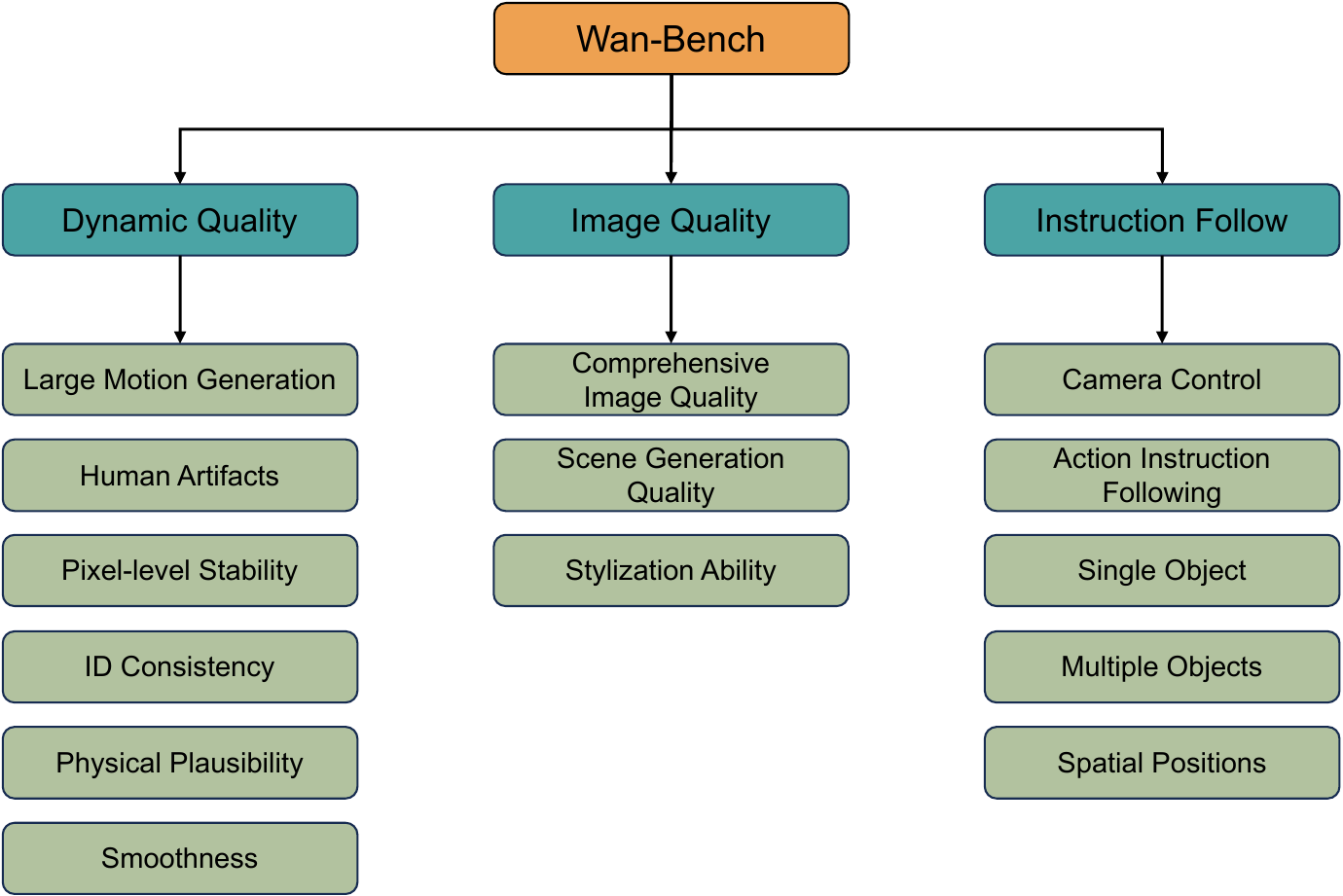}
        \caption{The dimensions covered in \method-Bench.}
        \label{vanbench-struct}
    \end{minipage}\hfill
    \begin{minipage}{0.47\linewidth}
        \centering
        \includegraphics[width=\linewidth]{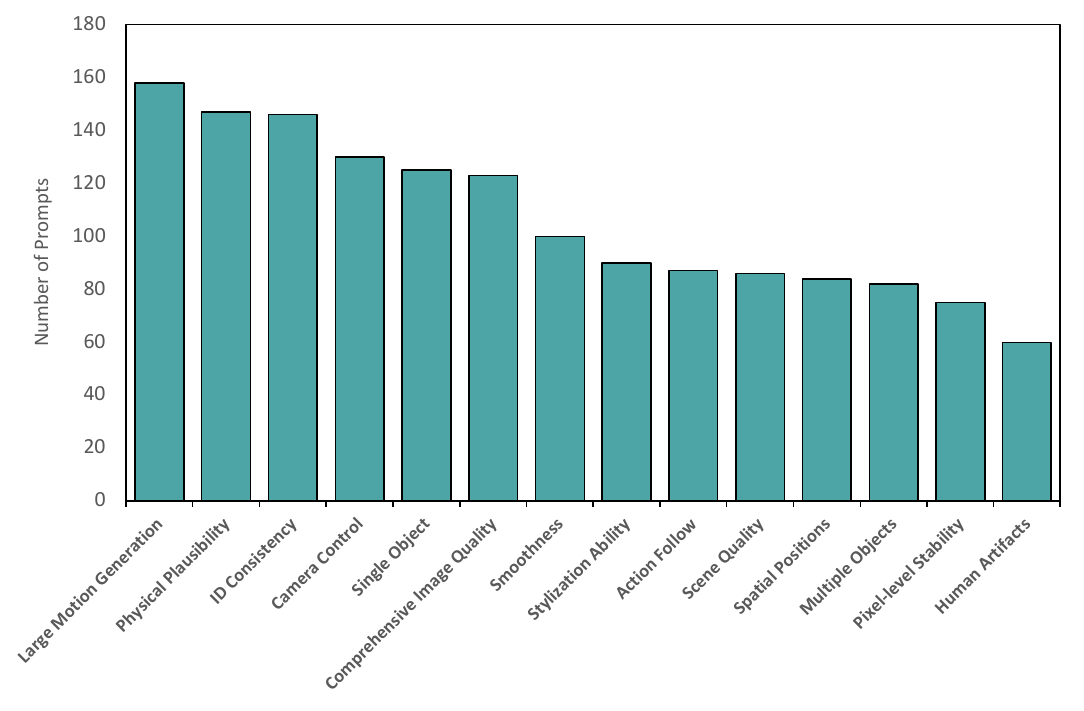}
        \caption{Distribution of the number of prompts across various dimensions of \method-Bench.}
        \label{prompt_dist}
    \end{minipage}
\end{figure}

\subsection{Benchmarks}

Existing video generation evaluation methods, such as Fréchet Video Distance (FVD)~\citep{2018Towards} and Fréchet Inception Distance (FID)~\citep{heusel2017gans}, lack alignment with human perception. We propose an automated, comprehensive, and human-aligned \method-Bench for evaluating video generation models.

\method-Bench is composed of three core dimensions: dynamic quality, image quality, and instruction following with 14 fine-grained metrics. We design specific scoring algorithm for each dimension, using traditional detectors for the simple tasks (\eg, object detection) and multi-modal Large Language Models (MLLMs) for the complex tasks.

\textbf{Dynamic quality}. Dynamic quality aims to evaluate the quality and stability of the model in non-static scenarios.

\textit{Large motion generation.} To measure the upper limit of motion generation, prompts are designed to encourage significant motions. 
We calculate the optical flow of the generated video using RAFT~\citep{2021RAFT} and assess the motion scores by the magnitude of the normalized optical flow.

\textit{Human artifacts.}
Existing evaluation methods lack the ability to detect AI-generated artifacts.
Thus, we train a YOLOv3-based model~\cite{2018YOLOv3} on 20,000 AI-generated images annotated by humans to identify the locations of artifacts.
The artifact scores take into account the predicted likelihood, the bounding boxes, and the corresponding duration.

\textit{Physical plausibility \& smoothness}. To evaluate physical plausibility, we design prompts to generate physics-related movements (\eg, ball bouncing, object interactions, water flow) and employ Qwen2-VL~\citep{qwenvl}, leveraging its extensive real-world physics knowledge. 
Through a video question-answer model, we detect violations of physical laws, such as object clipping, unrealistic collisions, or gravity-defying movements.
To evaluate smoothness, prompts are crafted to include complex motions, with Qwen2-VL identifying artifacts to assess motion fluidity.

\textit{Pixel-level stability}.
Pixel-level stability is used to determine if noise frequently appears in the video. We employ optical flow to identify static regions and calculate frame-wise differences within those areas. 
For stable videos, these differences within static regions are relatively minimal.

\textit{ID consistency.} Three kinds of sub-dimensions are considered: human consistency, animal consistency, and object consistency covering different collections of prompts. Frame-level DINO \citep{2021Emerging} features are extracted to compare the frame-wise similarity as the consistency score.

\textbf{Image quality.} The purpose of the image quality dimension is to evaluate the visual quality and aesthetic perception.

\textit{Comprehensive image quality.} Measuring the visual quality requires an evaluation of the fidelity and aesthetics. 
For fidelity, we employ MANIQA \citep{2022MANIQA}, which effectively detects blur and artifacts, ensuring precise clarity assessment. 
For aesthetics, we utilize two established models: the LAION-based aesthetic predictor~\citep{LAION-AI_aesthetic-predictor_2022} and MUSIQ~\citep{2021MUSIQ}. We compute the final image quality score by averaging these three evaluators.

\textit{Scene generation quality.}
We evaluate two key aspects: frame-wise scene consistency and scene-text alignment. For consistency, we measure CLIP similarity \citep{radford2021learning} across consecutive frames to ensure temporal coherence. 
For alignment, we compute CLIP similarity between frames and their corresponding text to verify semantic accuracy. 
We compute the quality score by weighted averaging these metrics.

\textit{Stylization.} We employ Qwen2-VL to evaluate the artistic video generation capability through frame-level question answering.

\textbf{Instruction following.} Instruction following assesses the model's ability to comprehend and execute textual instructions.

\textit{Single object \& multiple objects \& spatial positions.} 
We employ Qwen2-VL to predict object categories, counts, and spatial relationships within videos. 
The evaluation score represents the average number of frames that accurately correspond to the given prompts.

\textit{Camera control.}
We evaluate five camera movement styles: panning, lifting, zooming in/out, aerial shots, and tracking shots by 130 tailored prompts. For panning, lifting, and zooming, we analyze optical flow via RAFT to assess movement style. 
For complex aerial and tracking shots, we employ Qwen2-VL to infer camera movement through video question-answering.

\textit{Action instruction following.} We categorize motions into human (\eg, running), animal (\eg, crawling), and object movements (\eg, flying). 
As evaluating these motions requires prior knowledge of their execution, we employ Qwen2-VL by providing it with keyframes. 
We query the model for alignment of the action, completion of the action, and presence of artifacts to comprehensively evaluate alignment between text and video.

\textbf{Human feedback guided weighting strategy.}\label{hf-align} We utilize human feedback to assign a comprehensive video quality score, considering that each dimension has a different impact on user preferences rather than simply averaging them.
We collect over 5,000 pairwise comparisons of videos generated by different models including \method. Users evaluate pairs based on a given text, indicating their preference and assigning approximate scores. 
The dimension preference weighting is derived from the Pearson correlation between model-generated scores and human-rated scores, serving as a weighting factor in the final score calculation. 
\begin{table}[!t]
\centering
\renewcommand{\arraystretch}{1.3} 
\small
\resizebox{\textwidth}{!}{
\begin{tabular}{l|ccccccc}
\textbf{ \method-Bench Dimension} & \textbf{CNTopB} &  \textbf{Hunyuan} & \textbf{Mochi} & \textbf{CNTopA} & \textbf{Sora}& \textbf{\method 1.3B} & \textbf{\method 14B} \\
\shline
Large Motion Generation & 0.405  & 0.413	 & 0.420 & 0.284 & \textbf{0.482} & \underline{0.468} & 0.415 \\
Human Artifacts & 0.712 & 0.734	 & 0.622 & \textbf{0.833} & \underline{0.786} & 0.707  & 0.691 \\
Pixel-level Stability & 0.977  & \textbf{0.983} & \underline{0.981} & 0.974 & 0.952 & 0.976 & 0.972 \\
ID Consistency & \underline{0.940} & 0.935 & 0.930 & 0.936 & 0.925 & 0.938  & \textbf{0.946} \\
Physical Plausibility & 0.836  & 0.898 & 0.728 & 0.759 & \underline{0.933}& 0.912 & \textbf{0.939} \\
Smoothness & 0.765  & 0.890 & 0.530 & 0.880 & \textbf{0.930}& 0.790 & \underline{0.910} \\
Comprehensive Image Quality & 0.621 & 0.605 & 0.530 & \textbf{0.668} & \underline{0.665} & 0.596 & 0.640 \\
Scene Generation Quality & 0.369  & 0.373 & 0.368 & \underline{0.386} & \textbf{0.388} & 0.385  & \underline{0.386} \\
Stylization Ability & \textbf{0.623}  & 0.386 & 0.403 & 0.346 & \underline{0.606} & 0.430 & 0.328 \\
Single Object Accuracy & \textbf{0.987} & 0.912 & 0.949 & 0.942 & 0.932& 0.930  & \underline{0.952} \\
Multiple Object Accuracy & 0.840  & 0.850 & 0.693 & \underline{0.880} & \textbf{0.882} & 0.859 & 0.860 \\
Spatial Position Accuracy & \underline{0.518} & 0.464 & 0.512 & 0.434 & 0.458 & 0.476 & \textbf{0.590} \\
Camera Control & 0.465  & 0.406 & \textbf{0.605} & \underline{0.529} & 0.380 & 0.483 & 0.527 \\
Action Instruction Following & \textbf{0.917} & 0.735 & \underline{0.907} & 0.783 & 0.721 & 0.844 & 0.860 \\
\textbf{Weighted Score} & 0.690 & 0.673 & 0.639 & 0.693 & \underline{0.700}  & 0.689 & \textbf{0.724} \\
\end{tabular}
}
\caption{Performance comparison of commercial and open-source models using \method-Bench.}
\label{tab:model_performance}
\end{table}

\subsection{Evaluation}
\label{sec:t2v-results}

\subsubsection{Metrics and Results}

\textbf{Baselines and metrics.} As of the time of writing, there are numerous comparative competitors in the market, including commercial models such as Kling \citep{kuaishou2024kling}, Hailuo \citep{minimax2024hailuo}, Sora \citep{openaisora2024}, Runway \cite{runway2024gen3} and Vidu \citep{shengshu2024vidu}, as well as open-source models like Mochi \citep{genmo2024mochi}, CogVideoX \citep{cogvideox} and Hunyuan \citep{kong2024hunyuanvideo}. In this section, we compare our model with the commercial and open-source models that have been evaluated through benchmarks and received relatively positive user feedback.

\textbf{Quantitative results.} We collected 1,035 samples for each candidate model using \method-Bench prompts and conducted fair evaluations by scoring the models with \method-Bench. The evaluation focused on three key metrics: dynamic quality, image quality, and instruction-following accuracy. We rank the performance of the different generation models by calculating total scores weighted according to human preferences (Sec.~\ref{hf-align}). The results in Table \ref{tab:model_performance} indicate that \method outperforms existing commercial and open-source models.

\begin{figure}[htp]  
    \centering  
    \includegraphics[width=1.0\textwidth]{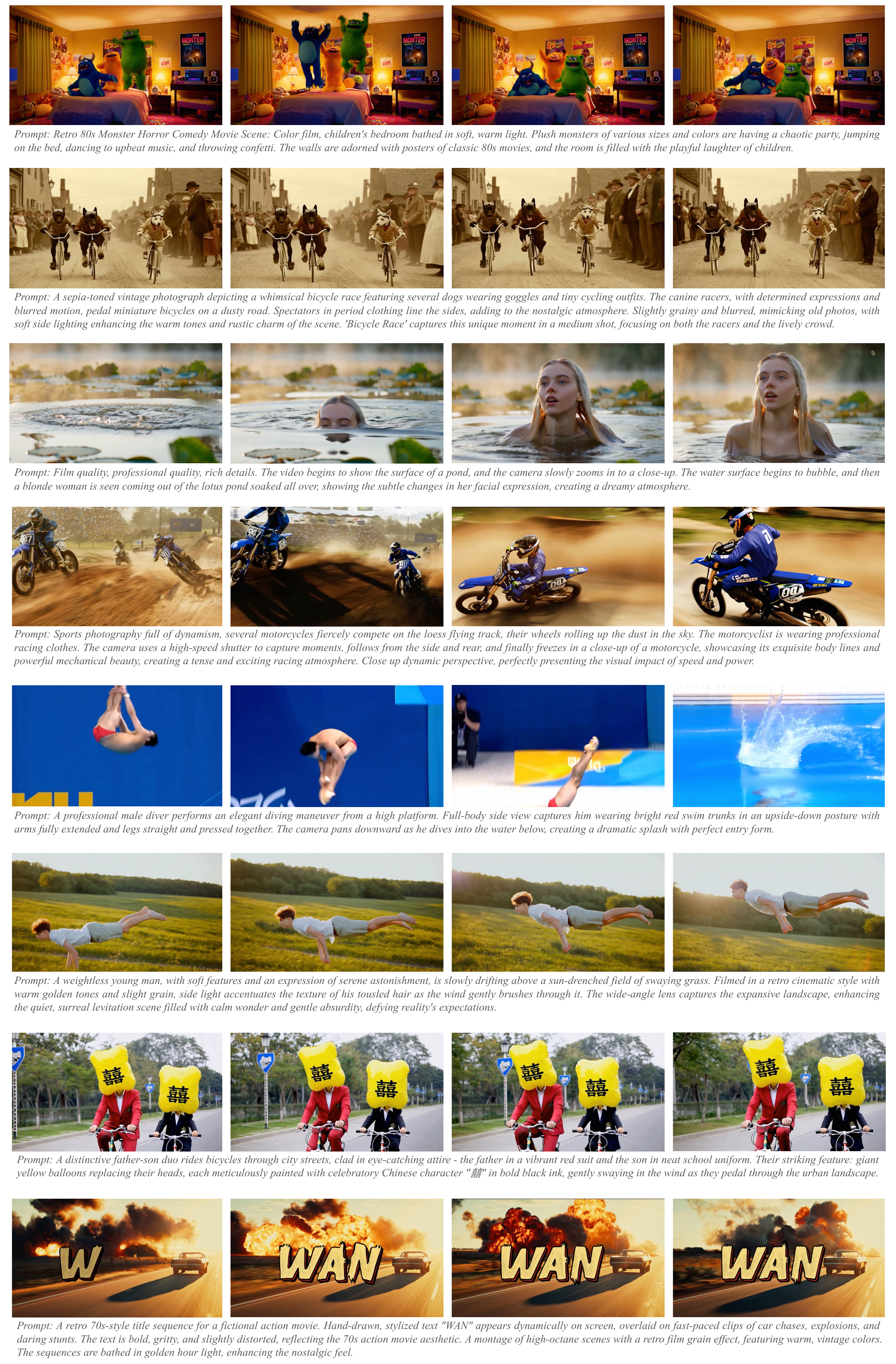}  
    \caption{Results of the \method-T2V. Our model excels at generating complex motions, creative transitions, cinematic-quality videos, and accurately produces both Chinese and English text.} 
    \label{fig:t2v_results} %
\end{figure}

\textbf{Qualitative results.} As illustrated in the Fig.~\ref{fig:t2v_results}, \method~effectively generates diverse, high-quality videos directly from textual descriptions. \method~excels at synthesizing dynamic scenes involving large-scale, complex movements, as well as scenarios reflecting physical interactions. Moreover, it adeptly handles various artistic styles and consistently produces cinematic-quality visuals. Additionally, \method~demonstrates robust multilingual text-generation capabilities, integrating Chinese and English texts into animations and giving rise to movie-level textual effects.

\textbf{Human evaluation.} We design over 700 evaluation tasks, which are annotated by more than 20 individuals and assessed by four key dimensions: alignment, image quality, dynamic quality, and overall quality.
The results in Table~\ref{tab:t2v_win_rate_gap} demonstrate that in the T2V task, the \method 14B model consistently excels across all visual quality dimensions, indicating its superior performance.

\begin{table}[t]
    \centering
    \begin{tabular}{l c c c c c}
        \toprule
        & CN-TopA & CN-TopB & CN-TopC & Runway & \textbf{All Rounds} \\ \midrule
        
        Visual Quality & \begin{tikzpicture}
            \draw[fill=teal!60, draw=none] (0,0) rectangle (0.4,0.306); 
            \end{tikzpicture} 
            30.6\% & 
            \begin{tikzpicture}
            \draw[fill=teal!60, draw=none] (0,0) rectangle (0.4,0.159); 
            \end{tikzpicture} 
            15.9\% & 
            \begin{tikzpicture}
            \draw[fill=teal!60, draw=none] (0,0) rectangle (0.4,0.278); 
            \end{tikzpicture} 
            27.8\% & 
            \begin{tikzpicture}
            \draw[fill=teal!60, draw=none] (0,0) rectangle (0.4,0.481); 
            \end{tikzpicture} 
            48.1\% & 
            5710 \\

        Motion Quality & \begin{tikzpicture}
            \draw[fill=teal!60, draw=none] (0,0) rectangle (0.4,0.161); 
            \end{tikzpicture} 
            16.1\% & 
            \begin{tikzpicture}
            \draw[fill=teal!60, draw=none] (0,0) rectangle (0.4,0.097); 
            \end{tikzpicture} 
            \space 9.7\% & 
            \begin{tikzpicture}
            \draw[fill=teal!60, draw=none] (0,0) rectangle (0.4,0.149); 
            \end{tikzpicture} 
            14.9\% & 
            \begin{tikzpicture}
            \draw[fill=teal!60, draw=none] (0,0) rectangle (0.4,0.403); 
            \end{tikzpicture} 
            40.3\% & 
            5785 \\

        Matching & \begin{tikzpicture}
            \draw[fill=teal!60, draw=none] (0,0) rectangle (0.4,0.460); 
            \end{tikzpicture} 
            46.0\% & 
            \begin{tikzpicture}
            \draw[fill=teal!60, draw=none] (0,0) rectangle (0.4,0.579); 
            \end{tikzpicture}
            57.9\% & 
            \begin{tikzpicture}
            \draw[fill=teal!60, draw=none] (0,0) rectangle (0.4,0.567); 
            \end{tikzpicture} 
            56.7\% & 
            \begin{tikzpicture}
            \draw[fill=teal!60, draw=none] (0,0) rectangle (0.4,0.691); 
            \end{tikzpicture} 
            69.1\% & 
            5578 \\

        Overall Ranking & \begin{tikzpicture}
            \draw[fill=teal!60, draw=none] (0,0) rectangle (0.4,0.440); 
            \end{tikzpicture} 
            44.0\% & 
            \begin{tikzpicture}
            \draw[fill=teal!60, draw=none] (0,0) rectangle (0.4,0.440); 
            \end{tikzpicture} 
            44.0\% & 
            \begin{tikzpicture}
            \draw[fill=teal!60, draw=none] (0,0) rectangle (0.4,0.489); 
            \end{tikzpicture} 
            48.9\% & 
            \begin{tikzpicture}
            \draw[fill=teal!60, draw=none] (0,0) rectangle (0.4,0.676); 
            \end{tikzpicture} 
            67.6\% & 
            5560 \\ 

        \bottomrule
    \end{tabular}
    \footnotesize
    \\
    \textit{*} The table displays the proportion of instances in which the \method~14B model outperformed other models in pairwise comparisons, relative to the total number of comparisons.
    \caption{Win rate gap of T2V.}
    \label{tab:t2v_win_rate_gap}
\end{table}



\textbf{\method in public leaderboard.}
\method has demonstrated state-of-the-art performance on the VBench leaderboard \citep{2023VBench}, a widely used evaluation approach for video generation task. 
The VBench employs a multidimensional assessment architecture that decomposes video quality evaluation into 16 distinct, human-aligned dimensions spanning aesthetic quality, motion smoothness, and semantic coherence.

Our analysis focuses on two model variants: the 14B parameter \method 14B and the \method 1.3B variant (shown in Table~\ref{tab:vbench_res}). The \method 14B model achieves a benchmark-leading performance with an aggregate score of 86.22\%, including 86.67\% in visual quality and 84.44\% in semantic consistency, significantly outperforming competitors like OpenAI's Sora and MiniMax's Hailuo. The efficient \method 1.3B variant also remains competitive, scoring 83.96\% and surpassing commercial models such as HunyuanVideo \citep{kong2024hunyuanvideo} and Kling 1.0 \citep{kuaishou2024kling}, as well as the open-sourced CogVideoX1.5-5B \citep{cogvideox}.
\begin{table}[h]
    \centering
    \small
    \renewcommand{\arraystretch}{1.3} 
    \resizebox{\textwidth}{!}{
    \begin{tabular}{@{}l|ccc@{}}
        Model Name & Quality Score & Semantic Score & Total Score \\  
        \shline
        MiniMax-Video-01 \citep{minimax2024hailuo} & 84.85\% & 77.65\% & 83.41\% \\
        Hunyuan (Open-Source Version) \citep{kong2024hunyuanvideo} & 85.09\% & 75.82\% & 83.24\% \\
        Gen-3 (2024-07) \citep{runway2024gen3} & 84.11\% & 75.17\% & 82.32\% \\
        CogVideoX1.5-5B (5s SAT prompt-optimized) \citep{cogvideox} & 82.78\% & 79.76\% & 82.17\% \\
        Kling (2024-07 high-performance mode) \citep{kuaishou2024kling}& 83.39\% & 75.68\% & 81.85\% \\  
        Sora \citep{openaisora2024} & 
        \underline{85.51\%} & 79.35\% & \underline{84.28\%} \\
        \method 1.3B & 84.92\% & \underline{80.10\%} & 83.96\% \\
        \method 14B (2025-02-24) & \textbf{86.67\%} & \textbf{84.44\%} & \textbf{86.22\%} \\
    \end{tabular}
    }
    \caption{Model performance scores on Vbench.}
    \label{tab:vbench_res}

\end{table}

\subsubsection{Ablation Study}
\label{subsubsec:ablation}

We provide ablation studies of key modules within our model design to offer insights into the overall architecture's contributions. Experiments are conducted on 1.3B version for rapid evaluation.

\textbf{Ablation on adaptive normalization layers.} Given the substantial parameter load of adaptive normalization layers~ \citep{perez2018film} (adaLN) in DiT~ \citep{dit}, we explore whether focusing on parameter volume in adaLN or increasing the depth of the network layers is more effective. To adjust the parameter volume, we follow the setup from PixArt~ \citep{chen2023pixartalpha} to study whether to share adaLN or not. In the non-shared configuration, each block’s scale and shift parameters are predicted by a block-specific MLP that takes as input the time embedding. Instead, the shared AdaLN configuration (referred to as AdaLN-single in PixArt) computes a single set of global parameters, which involves predicting scale and shift values, in the first block, which are then shared across all blocks. This sharing significantly reduces the number of parameters.

We evaluate four configurations: (i) Full-shared-adaLN-1.3B: This is our original model, where adaLN is fully shared across all 30 attention blocks.
(ii) Half-shared-adaLN-1.5B: AdaLN is shared in the first 15 attention blocks while not for the remaining blocks. This results in a 1.5B model with an increase of 0.2B parameters.
(iii) Full-shared-adaLN-1.5B (extended): AdaLN is shared across all attention blocks, but the model depth is increased to 35 layers, keeping the 1.5B parameter count.
(iv) Non-shared-AdaLN-1.7B: AdaLN is not shared across any blocks, and the model depth remains at 30 layers, resulting in a 1.7B model.
We maintain other parameters unchanged and train the models from scratch to perform text-to-image task for 200,000 steps (i.e., the first stage of our text-to-video training), with the global batch size of 1536. For comparison, we measure performance using the L2 loss between generated images and real images in the latent space during the training stage, with smaller training loss indicating better convergence.

\begin{figure}[t]  
    \scriptsize
    \centering  
    \includegraphics[width=0.7\textwidth]{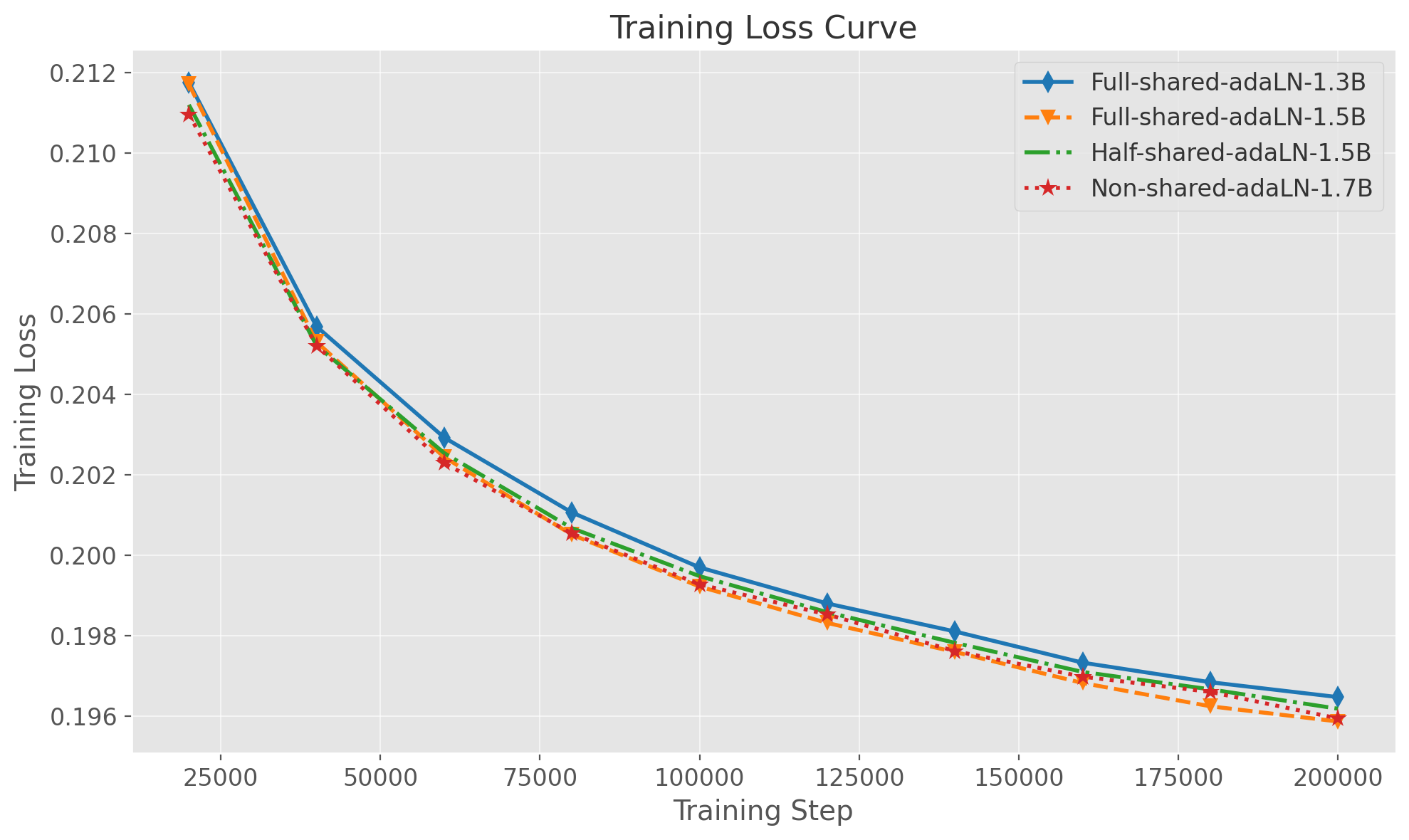}  
    \caption{Training loss curve with different configurations.} 
    \label{fig:ablation_adaln}  
\end{figure}

As presented in Fig.~\ref{fig:ablation_adaln}, the Full-shared-AdaLN-1.3B model, with fewer parameters, results in a slightly higher training loss compared to others. When comparing Half-shared-AdaLN-1.5B and Full-shared-AdaLN-1.5B, which have the same parameter count, Full-shared-AdaLN-1.5B consistently achieves the lowest training loss. This suggests that focusing on model depth rather than the parameters in AdaLN leads to better performance. Additionally, the Non-shared-AdaLN-1.7B model, despite having more parameters, does not outperform the Full-shared-AdaLN-1.5B and further supports this idea.
Therefore, we adopt the fully shared AdaLN design to effectively reduce parameter count while preserving performance.

\begin{figure}[!t]  
    \scriptsize
    \centering  
    \includegraphics[width=0.7\textwidth]{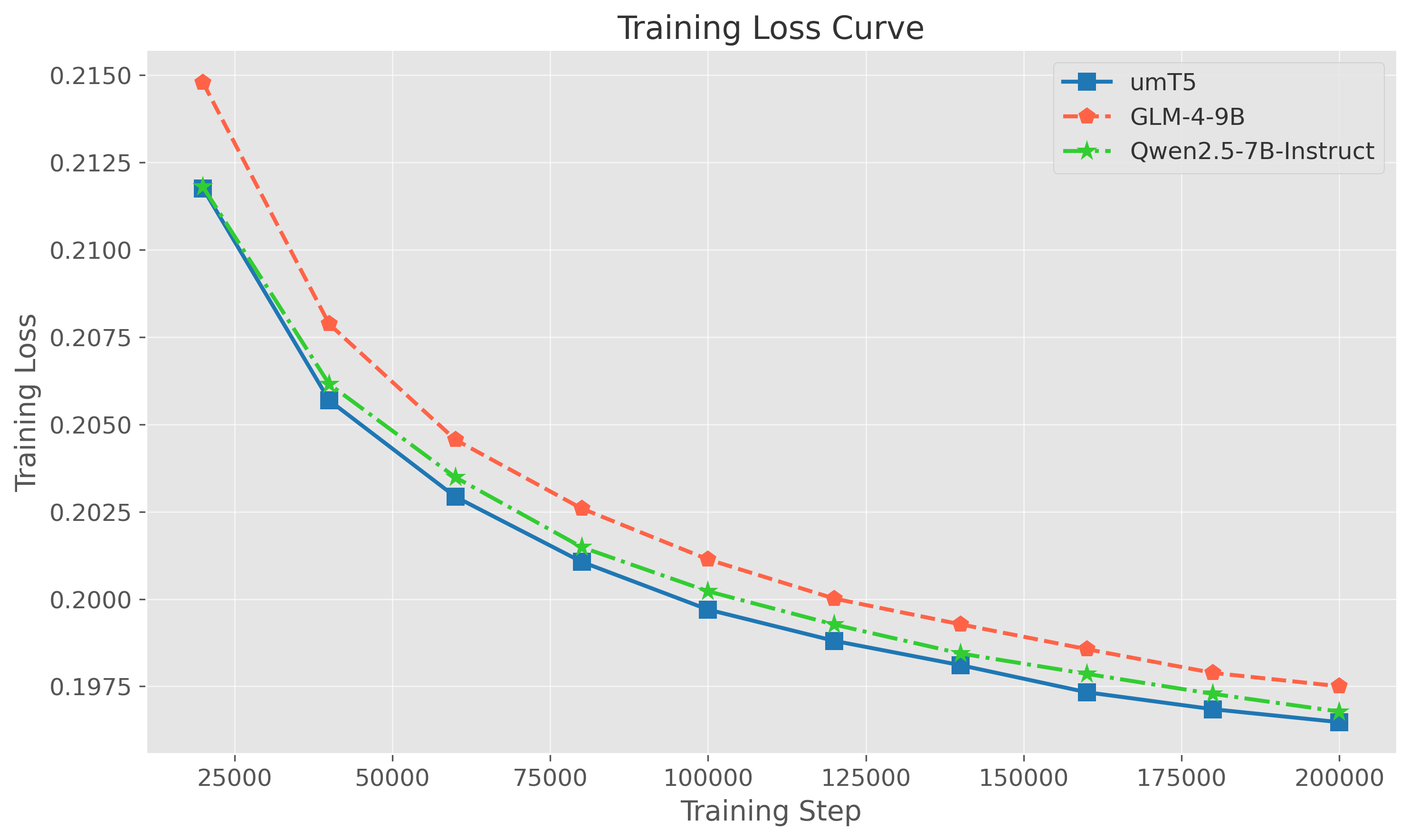}  
    \caption{Training loss curve with different text encoders.} 
    \label{fig:ablation_textenc}  
\end{figure}

\textbf{Ablation on text encoder.} We select three text encoders that can handle bilingual inputs: umT5~ \citep{chung2023unimax} (5.3B), Qwen2.5-7B-Instruct~ \citep{qwen2.5}, and GLM-4-9B~ \citep{glm2024chatglm}. The T5 series has been a popular choice in video generation models~ \citep{genmo2024mochi,cogvideox}, while the latter two encoders excel in language understanding, especially among all LLM models under 10B parameters. For this ablation, we keep other settings unchanged and train a text-to-image task with a global size of 1536.
Text embeddings are derived from the second-to-last layers of Qwen2.5-7B-Instruct and GLM-4-9B.
We measure the training loss as shown in Fig.~\ref{fig:ablation_textenc}.

\begin{wraptable}{r}{0.3\textwidth}
\setlength{\leftmargini}{-50pt}
\setlength{\tabcolsep}{10pt}
\warptablestyle{6pt}{1.4}
\centering
\begin{tabular}{l| c c }
Model & VAE & VAE-D \\
\shline
10k steps & \textbf{42.60} & 44.21 \\
15k steps & \textbf{40.55} & 41.16 \\
\end{tabular}
\caption{FID scores ($\downarrow$) of VAE and VAE-D.}
\label{tab:ablation_vae_fid}
\end{wraptable}
Compared to other powerful LLM-based encoders, umT5 demonstrates superior text embedding performance. Notably, HunyuanVideo~ \citep{kong2024hunyuanvideo} highlights that decoder-only LLMs use causal attention whereas umT5 adopts bidirectional attention, making it suitable for diffusion models.
Following HunyuanVideo's strategy, we incorporate a bidirectional token refiner as an adapter layer. We apply this setup to both Qwen2.5-7B-Instruct and GLM-4-9B, but the training losses and visualizations still show umT5 performing most favorably. Further, we compare umT5 with a pre-trained Multimodal Large
Language Model, i.e., Qwen-VL-7B-Instruct~\citep{qwenvl}.
As shown in Table~\ref{tab:ablation_textenc_fid}, using the second-to-last layer’s features from Qwen-VL results in comparable generation performance to umT5, but with a larger model size.

\begin{table}[h!]
\centering
\renewcommand{\arraystretch}{1.5} 
\setlength{\tabcolsep}{15pt} 
\resizebox{\textwidth}{!}{
\begin{tabular}{l| c c c }
Models & umT5 & Qwen-VL-7B (last layer) & Qwen-VL-7B (second last layer) \\
\shline
FID($\downarrow$) & 43.01 & 43.72 & \textbf{42.91} \\
\end{tabular}
}
\caption{FID scores for different text encoders. The ``last layer'' and ``second last layer'' refer to the text features extracted from the last and second last layers of the encoder, respectively.}
\label{tab:ablation_textenc_fid}
\end{table}

\textbf{Ablation on autoencoder.} In addition to our VAE, we design a variant called VAE-D, where the reconstruction loss is replaced by a diffusion loss. We adopt both the pre-trained VAE and VAE-D on the text-to-image generation task for 150,000 training steps until achieving convergence. We measure the FID scores at both 100,000 and 150,000 steps. As shown in Table~\ref{tab:ablation_vae_fid}, VAE model consistently achieves lower FID compared to VAE-D.

\section{Extended Applications}
\label{sec:applications}

\subsection{Image-to-Video Generation}
\label{sec:i2v}

The image-to-video (I2V) generation task focuses on synthesizing a dynamic video sequence from a static input image guided by a textual prompt. 
This approach significantly enhances the controllability of video generation by anchoring the output to a specific initial frame, thereby garnering substantial interest within the research community.
Existing approaches, such as I2VGen-XL~\citep{zhang2023i2vgen}, SVD~\citep{blattmann2023svd}, and CogVideo~\citep{cogvideox}, extend T2V frameworks to I2V by channel-wise concatenation of the conditional latent representation with the noise latent. 
Building upon this paradigm, our \method-I2V model adopts a similar strategy, capitalizing on the robust prior knowledge embedded in our foundational T2V model to achieve high-quality I2V generation.

\subsubsection{Model Design}

We introduce an additional condition image as the first frame to control video synthesizing. 
Specifically, we concatenate the condition image $I \in {R^{C\times 1\times H\times W}}$ with zero-filled frames along the temporal axis, and these guidance frames $I_{c} \in {R^{C\times T\times H\times W}}$ are compressed by \method-VAE into condition latent $z_{c} \in {R^{c\times t\times h\times w}}$  ($c=16$ denotes the latent channel, $t=1+(T-1)/4$, $h=H/8$, $w=W/8$). 
Additionally, we introduce a binary mask $M \in \{0,1\}^{1\times T\times h\times w}$ where $1$ indicates the preserved frame and $0$ denotes the frames to be generated. 
The spatial size of mask $M$ is consistent with condition latent $z_{c}$, but $M$ shares the same temporal length with the target video, which is then rearranged as shape of ${s\times t\times h\times w}$ where $s$ is temporal stride of \method-VAE. 
The noise latent $z_t$, condition latent $z_c$, and rearranged mask $m $ are concatenated along the channel axis and then passed through the DiT model of \method. 
Since the input of the I2V DiT model has more channels than T2V (\ie $2\times c + s$ v.s. $c$), an additional projection layer is employed, which is initialized with zero values.
Moreover, we utilize the image encoder of CLIP~\citep{radford2021learning} to extract feature representations from the condition image, and the extracted features are projected by a three-layer multi-layer perceptron (MLP), which plays a role as global context. 
The produced global context is then injected into the DiT model via decoupled cross-attention. 

\begin{figure}[t]  
    \centering  
    \includegraphics[width=0.99\textwidth]{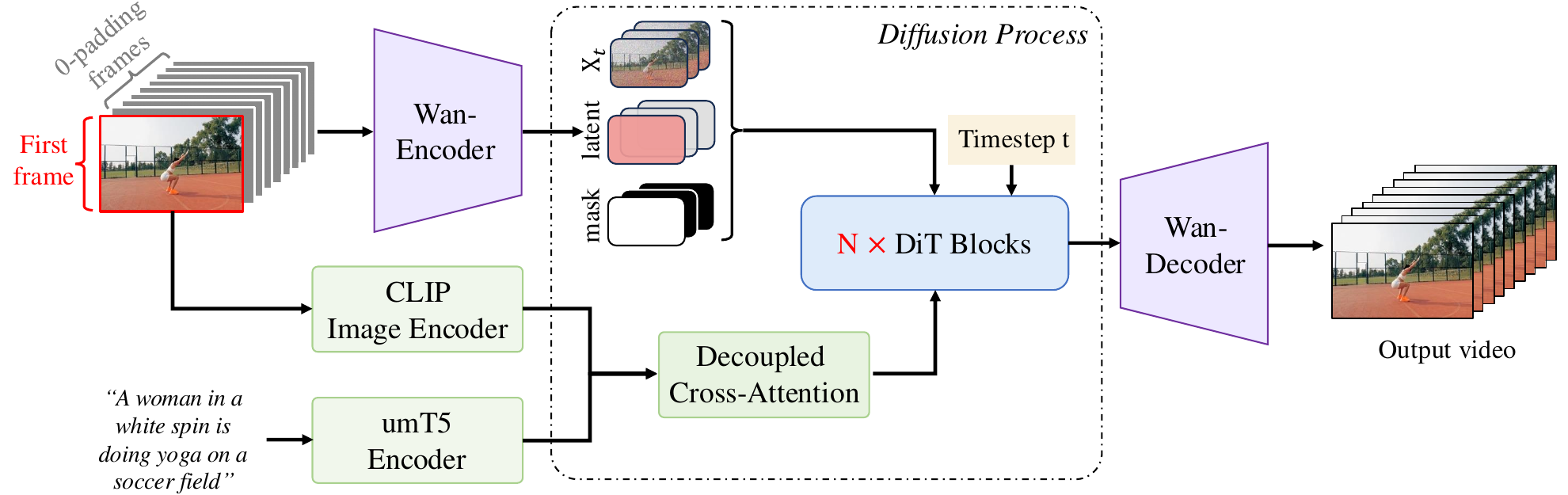}  
    \caption{\method-I2V model framework. The mask mechanism enables the model to selectively focus on input frames such that the proposed framework demonstrates compatibility with video continuation and first-last frame transformation tasks.} 
    \label{fig:v2av2-i2v_pipeline} %
\end{figure}

In practice, the aforementioned approach as illustrated in Fig.~\ref{fig:v2av2-i2v_pipeline} is extensible to other controllable generation tasks, such as first-last frame transformation and video continuation.
The proposed masking mechanism explicitly delineates the input conditions and specifies the frames to be generated.
To optimize the efficacy of the masking strategy, we incorporate multiple tasks, \ie image-to-video generation, video continuation, first-last frame transformation, and random frame interpolation, into our unified framework. 
Specifically, we adopt the following training paradigm for the I2V model. 
During the joint training phase, we conduct unified pre-training of our mask-guided model across these diverse tasks. 
This phase facilitates the model's ability to discern which positions should be preserved and which should be generated, depending on the input content. 
Subsequently, in the fine-tuning stage, we concentrate on refining the model's performance on each individual task.

\subsubsection{Dataset}

During the joint training phase, we utilize the same training dataset employed for T2V pertaining.
This phase equips the model with the foundational capability to generate motion conditioned on reference images. 
In the fine-tuning phase, we curate task-specific datasets tailored to the unique characteristics of each individual task.

\textbf{Image-to-video dataset}. In the early experiments, we found that when the first frame in the training data significantly differed from the video content, the model struggled to learn stable image-driven video generation. 
From this perspective, a smaller difference between the first frame and the video content in the training data is beneficial for model training. 
We calculate the difference between the first frame and the video content based on SigLIP~\citep{zhai2023sigmoid} features. Specifically, we compute the cosine similarity between the features of the first frame and the mean of the features of the remaining frames, and then we retain only the videos with a similarity greater than a predefined threshold.

\textbf{Video continuation dataset.} 
Similarly, our findings also indicate that training on temporally consistent videos significantly improves the performance of video continuation tasks. 
To quantify the consistency between the initial and final segments of a video (\ie the first 1.5 seconds and the last 3.5 seconds), we compute the cosine similarity of their respective SigLIP features, followed by using the computed similarity scores to filter and select the required training dataset.

\textbf{First-last frame transformation dataset.} Unlike the I2V task, the community places greater emphasis on the smooth transition between the initial and final frames in the mage transition task. To this end, we increase the proportion of data samples with significant transition between the first and last frames in the training dataset.

\begin{figure}[htp]  
    \centering  
    \includegraphics[width=1.0\textwidth]{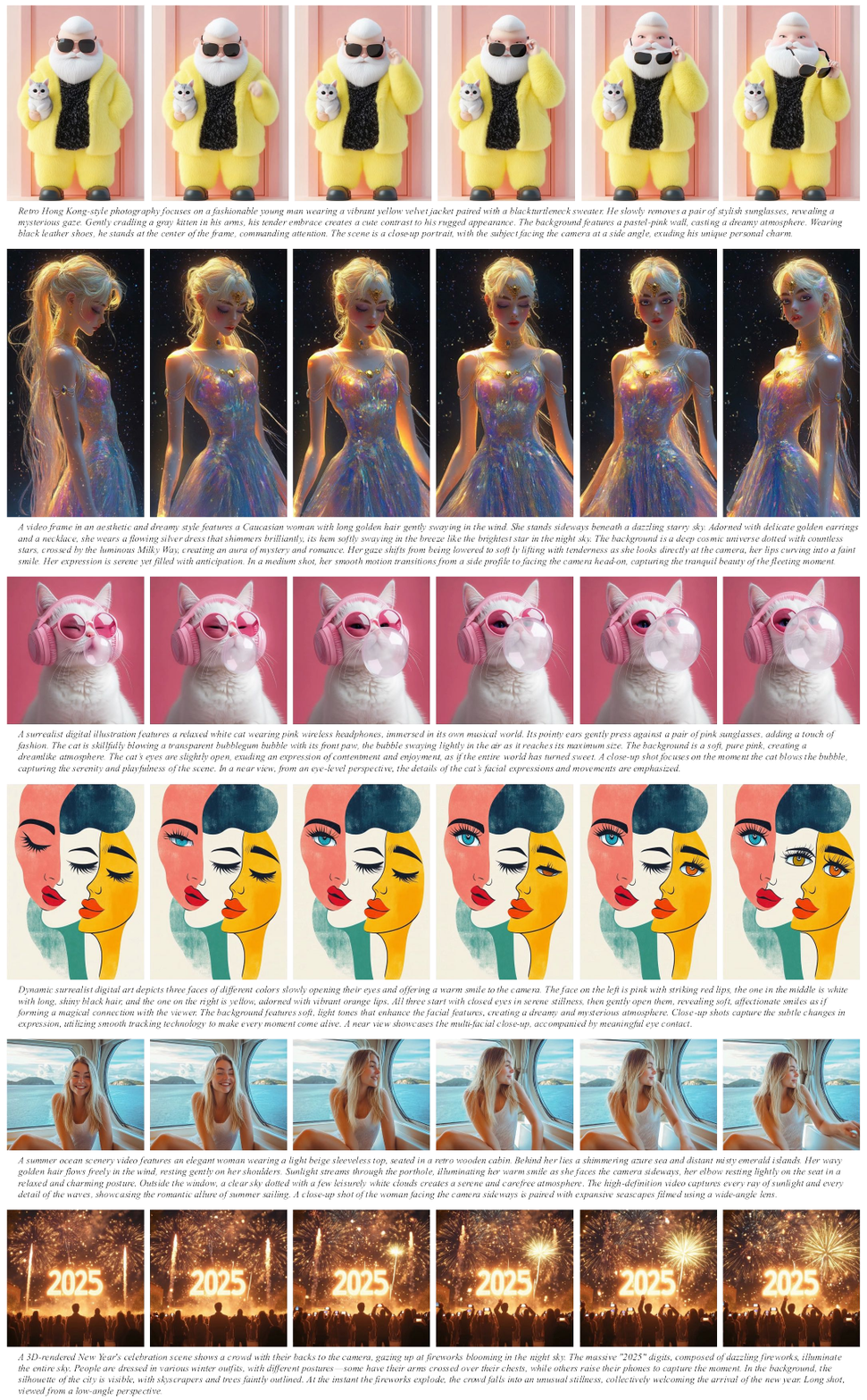}  
    \caption{Videos generated by \method-I2V model. The results demonstrate that our I2V model effectively captures and replicates the dynamics of real-world scenes.} 
    \label{fig:i2v_results_v1} %
\end{figure}

\begin{figure}[htp]  
    \centering  
    \includegraphics[width=1.0\textwidth]{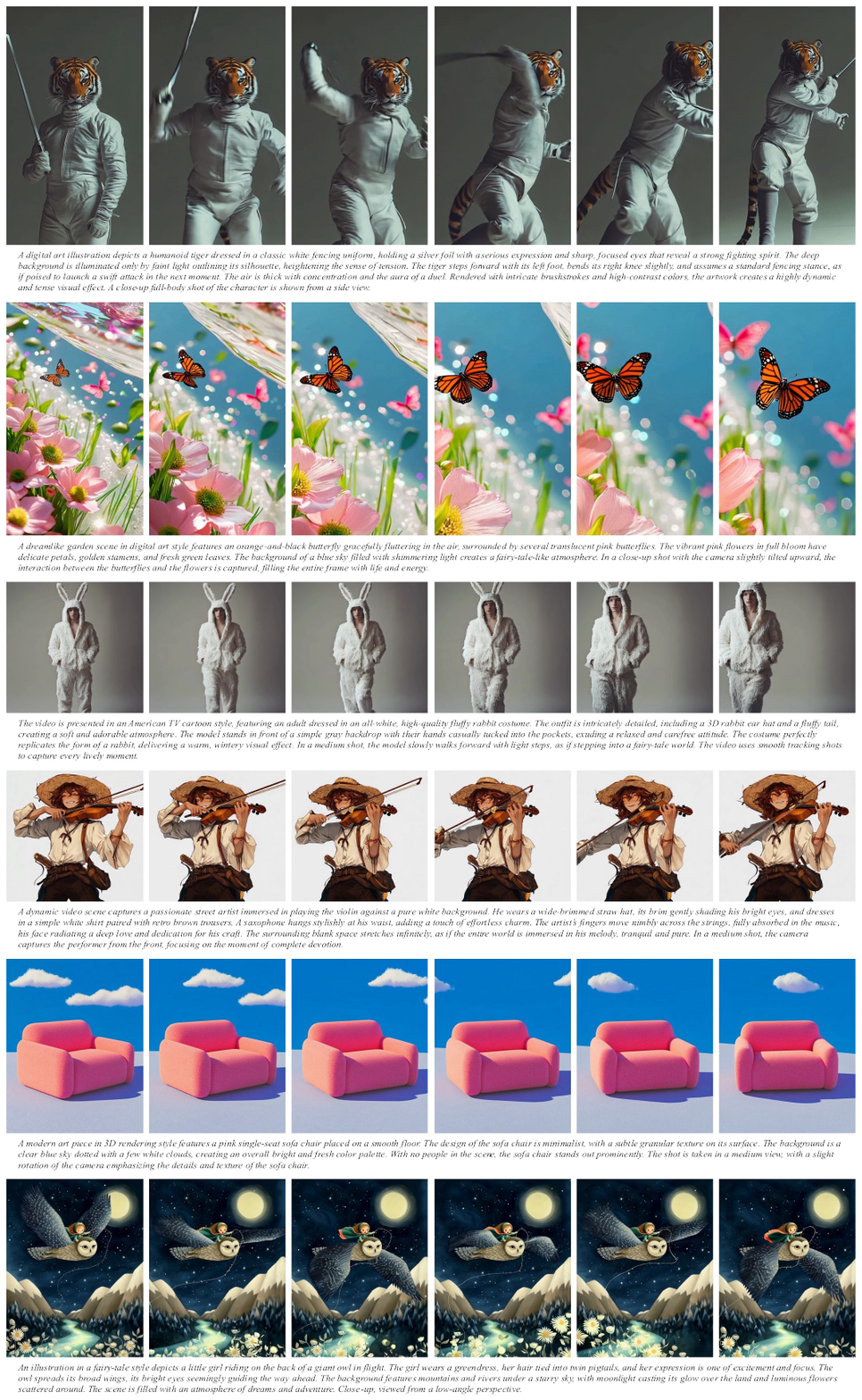}  
    \caption{Videos generated by \method-I2V model. The results indicate that our I2V model can effectively animate diverse types of images into highly realistic and dynamic videos.} 
    \label{fig:i2v_results_v2} 
\end{figure}

\subsubsection{Evaluation}

The training of the I2V model, along with first-last frame transformation and video continuation, is structured into two stages. In the initial pre-training stage, we employ the same dataset as that used in the T2V task. 
Given the slightly larger number of frames involved in this stage, the image encoder branch is excluded, which is motivated by the need to maintain a strong alignment between textual input and visual output, ensuring that the model retains its sensitivity to nuanced textual semantics.
Upon the stabilization of the pre-training stage, the model has already manifested a variety of capabilities, including I2V generation, video continuation, and so on. This demonstrates the efficacy of the pre-training process in establishing a robust foundation for both spatial and temporal understanding.

Our empirical observations reveal that relying solely on frame-level information is inadequate for achieving optimal performance in specific tasks, such as image-to-video generation, video continuation, and first-last frame transformation.

This insight indicates that while the pre-trained model offers a foundational framework for frame-based video generation, it lacks the requisite contextual and semantic depth to fully capture the temporal consistency of dynamic video sequences, particularly when the number of conditioning frames is highly limited.

To address this limitation, during the SFT stage, we incorporated an image encoder to extract image features through decoupled cross-attention~\citep{ye2023ip} like~\citep{i2vadapter}, providing global contextual information to enhance the model's capacities.
We conducted experiments based on two pre-trained checkpoints at 480p and 720p resolutions, covering the training of image-to-video generation, video continuation, and first-last frame transformation tasks.

Similarly to human evaluation for our T2V model, we benchmark our I2V model against state-of-the-art techniques to provide a comprehensive comparison focusing on key aspects such as visual quality, motion quality, and matching. The results shown as Table~\ref{tab:i2v_win_rate_gap} demonstrate that our model performs favorably across all  dimensions evaluated. Fig.~\ref{fig:i2v_results_v1} and Fig.~\ref{fig:i2v_results_v2} present additional visualizations of our I2V model, offering a clear illustration of its superior performance in animating images into high-quality videos.

\begin{table}[t]
    \centering
    \begin{tabular}{l| c c c c c}
        & CN-TopA & CN-TopB & CN-TopC & CN-TopD & \textbf{All Rounds} \\ 
        \shline
        Visual Quality & \begin{tikzpicture}
            \draw[fill=teal!60, draw=none] (0,0) rectangle (0.4,0.292); 
            \end{tikzpicture} 
            29.2\% & 
            \begin{tikzpicture}
            \draw[fill=teal!60, draw=none] (0,0) rectangle (0.4,0.608); 
            \end{tikzpicture} 
            60.8\% & 
            \begin{tikzpicture}
            \draw[fill=teal!60, draw=none] (0,0) rectangle (0.4,0.246); 
            \end{tikzpicture} 
            24.6\% & 
            \begin{tikzpicture}
            \draw[fill=teal!60, draw=none] (0,0) rectangle (0.4,0.556); 
            \end{tikzpicture} 
            55.6\% & 
            896 \\

        Motion Quality & \begin{tikzpicture}
            \draw[fill=teal!60, draw=none] (0,0) rectangle (0.4,0.217); 
            \end{tikzpicture} 
            21.7\% & 
            \begin{tikzpicture}
            \draw[fill=teal!60, draw=none] (0,0) rectangle (0.4,0.217); 
            \end{tikzpicture} 
            \space 21.7\% & 
            \begin{tikzpicture}
            \draw[fill=teal!60, draw=none] (0,0) rectangle (0.4,0.325); 
            \end{tikzpicture} 
            32.5\% & 
            \begin{tikzpicture}
            \draw[fill=teal!60, draw=none] (0,0) rectangle (0.4,0.670); 
            \end{tikzpicture} 
            67.0\% & 
            890 \\

        Matching & \begin{tikzpicture}
            \draw[fill=red!60, draw=none] (0,0) rectangle (0.4,0.042); 
            \end{tikzpicture} 
            -4.2\% & 
            \begin{tikzpicture}
            \draw[fill=teal!60, draw=none] (0,0) rectangle (0.4,0.350); 
            \end{tikzpicture}
            35.0\% & 
            \begin{tikzpicture}
            \draw[fill=teal!60, draw=none] (0,0) rectangle (0.4,0.517); 
            \end{tikzpicture} 
            51.7\% & 
            \begin{tikzpicture}
            \draw[fill=teal!60, draw=none] (0,0) rectangle (0.4,0.722); 
            \end{tikzpicture} 
            72.2\% & 
            896 \\

        Overall Ranking & \begin{tikzpicture}
            \draw[fill=teal!60, draw=none] (0,0) rectangle (0.4,0.108);
            \end{tikzpicture} 
            10.8\% & 
            \begin{tikzpicture}
            \draw[fill=teal!60, draw=none] (0,0) rectangle (0.4,0.475); 
            \end{tikzpicture} 
            47.5\% & 
            \begin{tikzpicture}
            \draw[fill=teal!60, draw=none] (0,0) rectangle (0.4,0.508);
            \end{tikzpicture} 
            50.8\% & 
            \begin{tikzpicture}
            \draw[fill=teal!60, draw=none] (0,0) rectangle (0.4,0.816);
            \end{tikzpicture} 
            81.6\% & 
            890 \\ 

    \end{tabular}
    \caption{Win rate gap of I2V models. The values in the table represent the proportion of instances in which the \method-I2V model was preferred in pairwise comparisons against other models, relative to the total number of comparisons conducted.}
    \label{tab:i2v_win_rate_gap}
\end{table}
\subsection{Unified Video Editing}
\label{sec:video_editing}

The foundational pre-trained models for text-to-image and text-to-video generation have facilitated the expansion of various downstream tasks and applications, including repainting~\citep{sdinp, propainter}, editing~\citep{sdedit, ip2p, magicbrush, wang2023videocomposer, dreamVideo2}, controllable generation~\citep{controlnet, scedit, wang2024unianimate}, customized generation~\citep{anydoor, wei2025DreamRelation},
frame reference generation~\citep{cogvideox, i2vadapter}, efficient generation~\citep{wang2023videolcm, yuan2024instructvideo, liu2024timestep}, and ID-referenced video synthesis~\citep{largen, consisid, phantom}. To enhance task flexibility and minimize the overhead associated with deploying multiple models, researchers have increasingly focused on developing unified model architectures~\citep{ominictr, unireal}, such as ACE~\citep{ace, acepp} and OmniGen~\citep{omnigen}. These unified architectures aim to integrate diverse tasks within a single image model, thereby facilitating the creation of various application workflows while maintaining ease of use. In the domain of video, the collaborative transformations in both temporal and spatial dimensions suggest that leveraging a unified model can unlock limitless possibilities for video creation. 

In our previous work, VACE~\citep{vace}, we introduced a unified framework for controllable video generation and editing, addressing tasks such as reference-to-video generation, video-to-video editing, and masked video-to-video editing. Building upon the pre-trained model of \method, we integrate multiple modalities, including images and videos for editing, references, and masks, into the Video Condition Unit (VCU) to support diverse input formats. VACE employs a concept decoupling strategy that enables the model to identify which aspects should be retained and which should be modified, thereby distinctly separating visual modality information in editing and reference tasks. We offer two training modes. The first mode involves training with the VCU as input, fully fine-tuning the entire \method model. The second mode utilizes a pluggable Context Adapter structure in a Res-Tuning~\citep{restuning} manner, allowing support for controllable and editing tasks without modifying the base model's weights but with an increased overall model scale. Leveraging the robust video generation capabilities of \method, this innovative framework demonstrates significant competitiveness in both quantitative and qualitative evaluations. It facilitates the compositional expansion of fundamental tasks, enabling scenarios such as long video re-rendering. Consequently, the framework offers a versatile and efficient solution for video synthesis, opening new avenues for user-driven video content creation and editing.

\begin{figure}[t]
    \scriptsize
    \centering
    \includegraphics[width=0.95\linewidth]{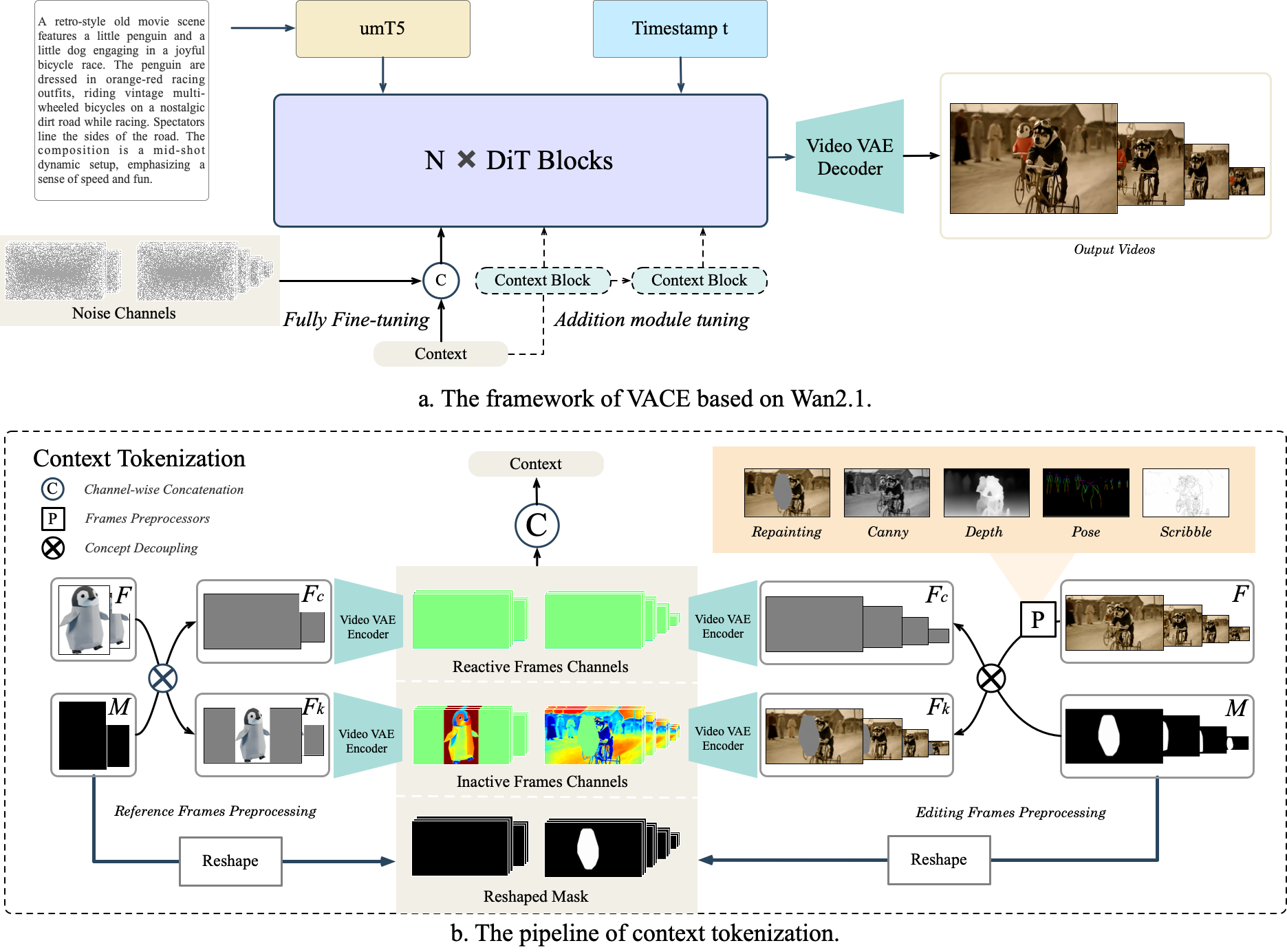}
    \caption{Unified controllable generation and editing framework. Frames and masks are tokenized through Concept Decoupling, Context Latent Encode, and Context Embedder. We propose two training strategies for \method: Fully Fine-tuning and Context Adapter Tuning.}
    \label{fig:framework_editing}
\end{figure}

\subsubsection{Model Design}

The Video Condition Unit (VCU), proposed in VACE, serves as an input paradigm that unifies diverse input conditions into textual inputs, frame sequences, and mask sequences. It can be formally represented as follows:
\begin{equation}
   V=[T;F;M],
\end{equation}
where $T$ is a text prompt, while $F$ and $M$ are sequences of context video frames $\{u_1,u_2,...,u_n\}$ and masks $\{m_1,m_2,...,m_n\}$ respectively. Here, $u$ is in RGB space, normalized to \([-1, 1]\) and $m$ is binary, in which ``1"s and ``0"s symbolize where to edit or not. $F$ and $M$ are aligned both in spatial size $h \times w$ and temporal size $n$. 
Under this paradigm, we employ the previously described \method-VAE to tokenize the input video frames and mask information into context tokens. 
As illustrated in  Fig.~\ref{fig:framework_editing} (a), these context tokens are then combined with noisy video tokens to fine-tune the \method model. Additionally, a Context Adapter Tuning strategy is proposed, which allows context tokens to pass through Context Blocks and be reintegrated into the original DiT blocks. This approach facilitates the seamless integration of contextual information, enhancing the model's ability to handle diverse editing and generation tasks without altering the base model's weights.

\begin{figure}[t]
    \scriptsize
    \centering
    \includegraphics[width=0.9\linewidth]{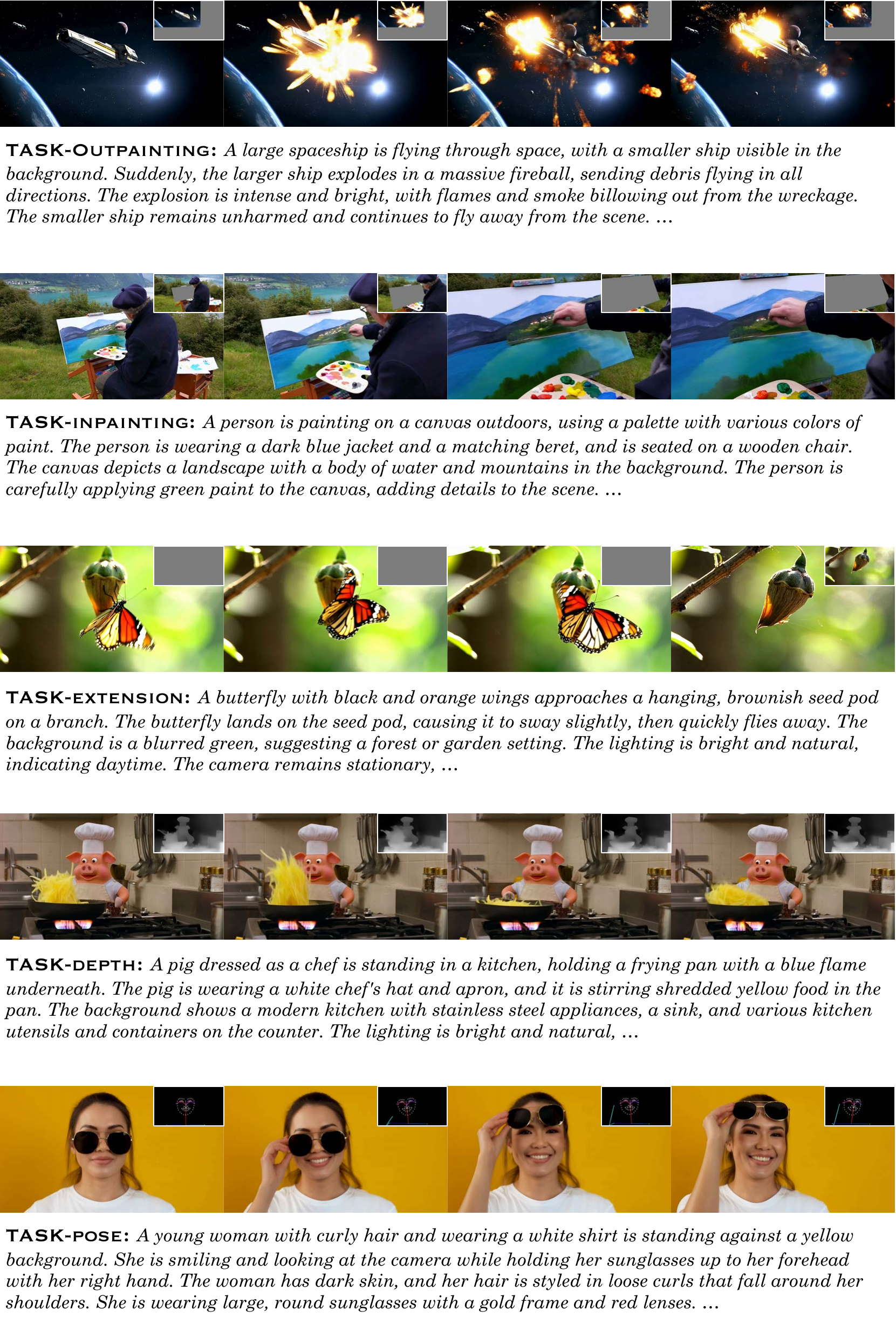}
    \caption{The visualization results of the proposed VACE.}
    \label{fig:video_editing_showcase1}
    \vspace{-2mm}
\end{figure}

\begin{figure*}[h]
    \scriptsize
    \centering
    \includegraphics[width=0.93\linewidth]{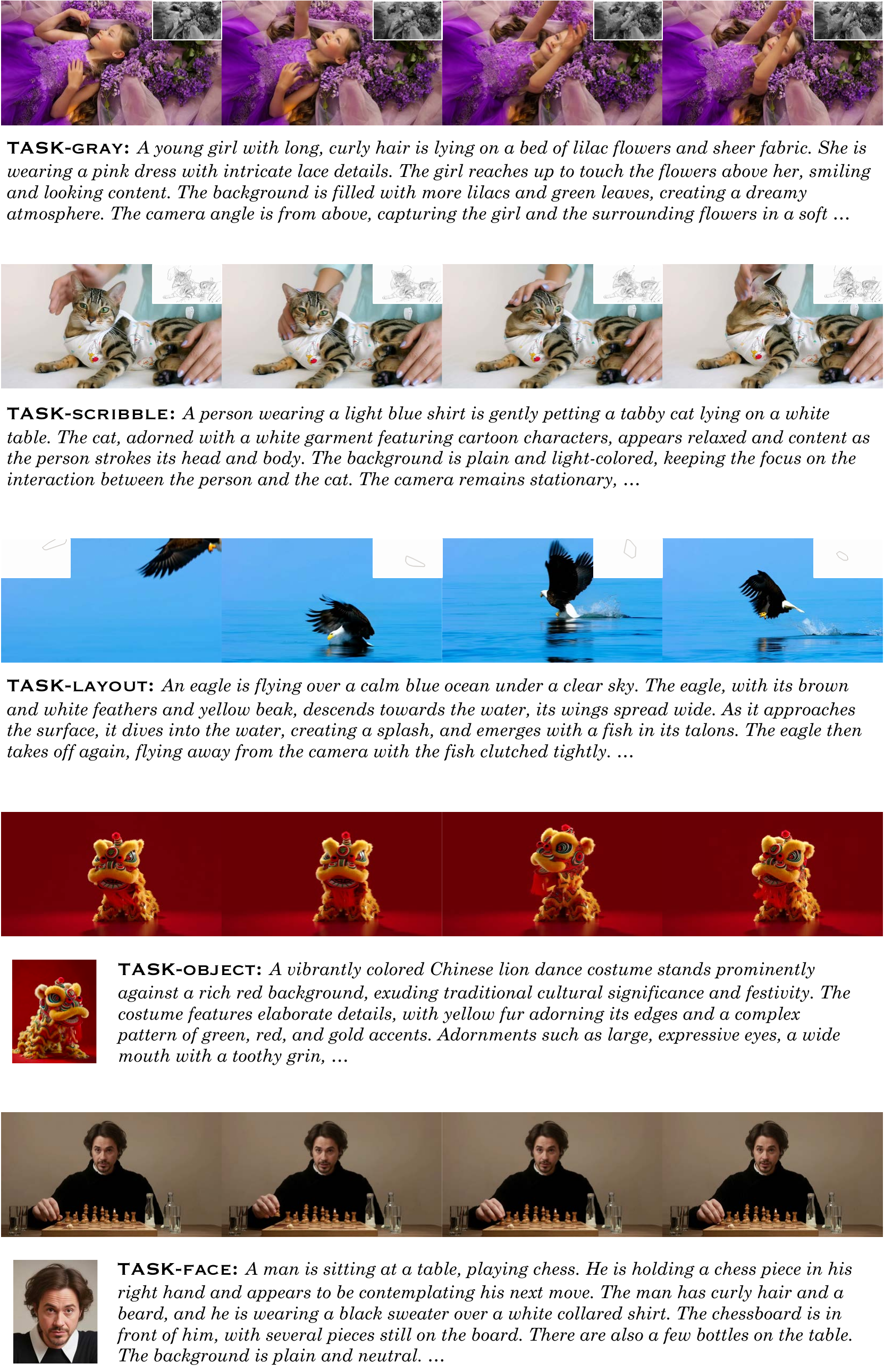}
    \caption{The visualization results of the proposed VACE.}
    \label{fig:video_editing_showcase2}
\end{figure*}

Before tokenization, pixel-level context frames and masks are preprocessed using a concept decoupling strategy and then encoded into the latent space. As shown in Fig.~\ref{fig:framework_editing} (b), the concept decoupling strategy explicitly separates the data of different modalities and distributions based on masks into two frame sequences of identical shape: $F_c = F \times M$ and $F_k = F \times (1-M)$, where $F_c$ refers to reactive frames containing all pixels to be modified, while $F_k$ denotes inactive frames that retain all pixels to be preserved, named inactive frames. This mechanism ensures clear task definitions and guarantees the model's convergence across different tasks. $F_c$, $F_k$ are processed by \method-VAE and mapped into the same latent space of $X$, maintaining their spatio-temporal consistency. To avoid any mishmash of images and videos, reference images are separately encoded by the \method-VAE encoder and concatenated back along the temporal dimension, while the corresponding parts need to be removed during decoding. $M$ is directly reshaped and interpolated. After that, $F_c$, $F_k$, and $M$ are mapped into latent spaces and spatio-temporal aligned with $X$ in the shape of $n' \times h' \times w'$.

\begin{figure*}[!t]
    \scriptsize
    \centering
   \includegraphics[width=0.9\linewidth]{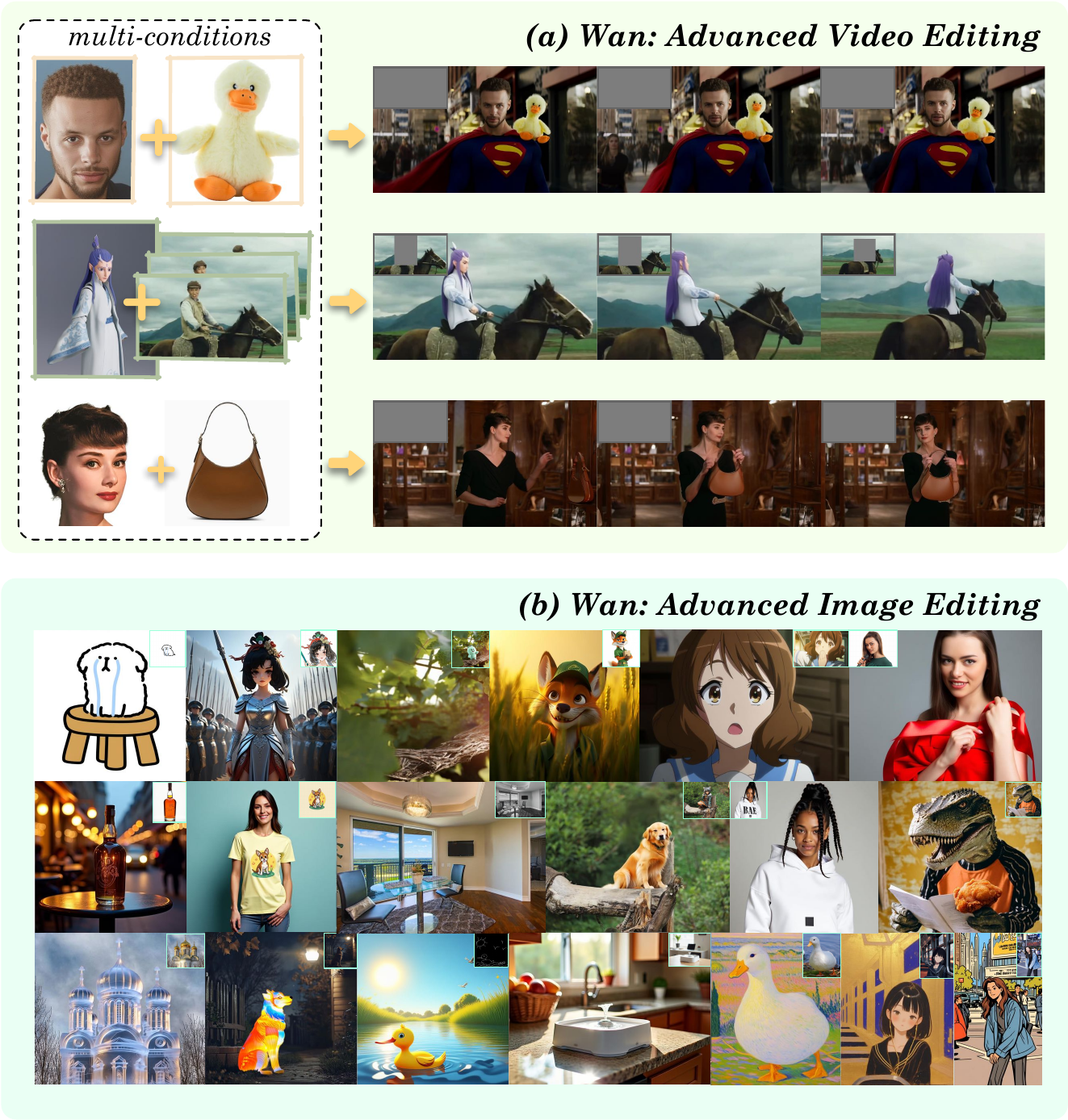}
    \caption{The visualization results of advanced video/image editing.}
    \label{fig:video_editing_showcase3}
\end{figure*}

\subsubsection{Datasets and Implementation}

\textbf{Data construction}. To develop an all-in-one model, the diversity and complexity of data construction must increase. For controllable video generation and editing tasks, input modalities are expanded to include target videos, source videos, local masks, references, and more. Efficient and rapid data acquisition for various tasks necessitates maintaining video quality while performing instance-level analysis and understanding. We begin by performing shot slicing on the video data and initially filtering based on resolution, aesthetic score, and motion amplitude. The first frame is then labeled using RAM~\citep{ram} and processed with Grounding DINO~\citep{groundingdino} for detection, enabling secondary filtering of videos with target areas that are either too small or too large. Additionally, we utilize the propagation operation of SAM2~\citep{sam2} for video segmentation to obtain instance-level information throughout the video. Based on the segmentation results, we further filter instances temporally by calculating the effective frame ratio according to a mask area threshold. The construction for different tasks must be tailored to each task's specific characteristics. For detailed information, refer to the work~\citep{vace}.

\textbf{Implementation details}. We train the controllable and editing model, pre-trained on \method-T2V-14B, to support resolutions up to 720p through a multi-stage training process. Initially, we focus on foundational tasks such as inpainting and extension to complement the pre-trained text-to-video models. 
This phase incorporates masks and develops contextual generation capabilities in both spatial and temporal dimensions. Subsequently, we expand the task by transitioning from single-input to multiple-input reference frames and from individual to composite tasks. Finally, we enhance the model's quality by fine-tuning it with higher-quality data and longer sequences.

\subsubsection{Evaluation}

In Fig.~\ref{fig:video_editing_showcase1} and Fig.~\ref{fig:video_editing_showcase2}, we present the results of the \method single model across various tasks. 
It is evident that the model achieves a high level of performance in video quality and temporal consistency. Additionally, we demonstrate new applications resulting from the combination of different generative capabilities, as illustrated in Fig.~\ref{fig:video_editing_showcase3} (a). Our model performs effectively on these tasks, showcasing significant potential for capability expansion. 
The unified controllable generation and editing framework proposed by VACE is also applicable to image generation and editing, as shown in Fig.~\ref{fig:video_editing_showcase3} (b).

\subsection{Text-to-Image Generation}
\label{sec:t2i}

\begin{figure}[ht!]  
    \scriptsize
    \centering  
    \includegraphics[width=0.92\textwidth, height=0.93\textheight]{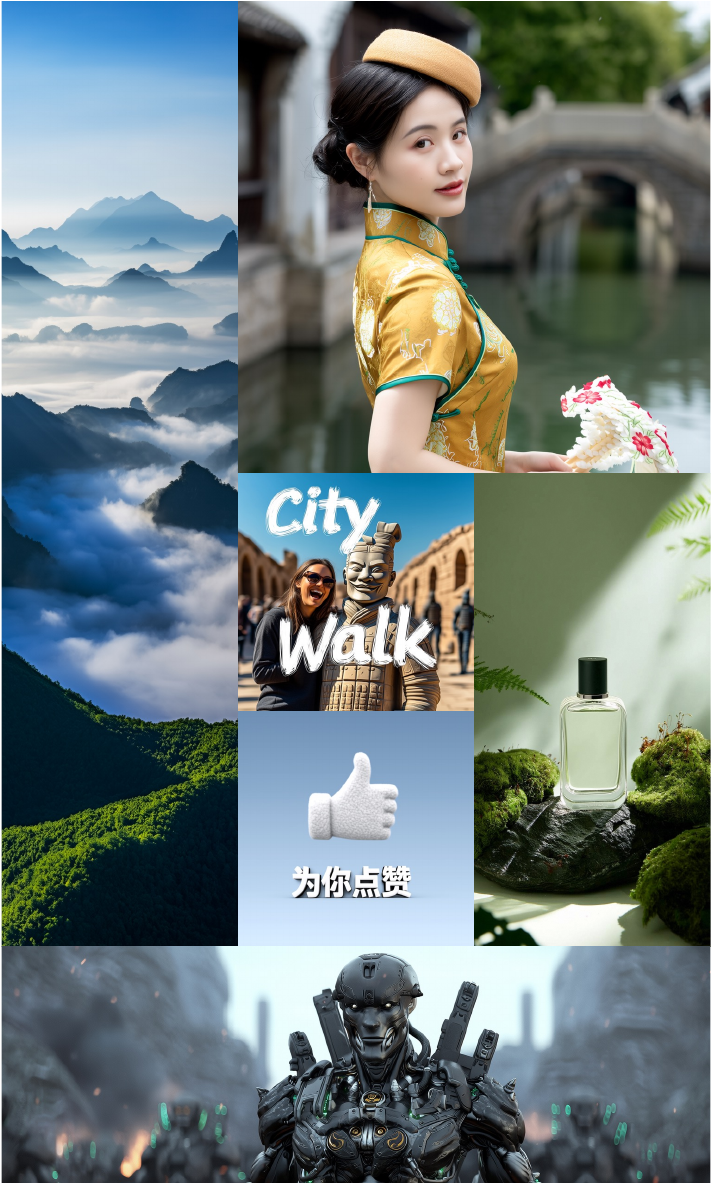}  
    \caption{Image samples generated from \method model.} 
    \label{fig:image_show}  
\end{figure}

\method is jointly trained on both image and video datasets, equipping it with not only advanced video generation capabilities but also exceptional image synthesis performance.
This dual-training strategy facilitates cross-modal knowledge transfer, creating a highly versatile framework that achieves powerful results in both domains.
In practice, our model has been trained on an image dataset nearly ten times larger than its video counterpart, allowing for synergistic benefits between image and video generation tasks.
This extensive training regimen has led to outstanding performance in various aspects of image synthesis.
As demonstrated in Fig.~\ref{fig:image_show}, \method generates high-fidelity images across diverse categories, encompassing artistic text-based visuals, photorealistic portraits, imaginative creative designs, and professional-grade product photography.

\subsection{Video Personalization}
\label{sec:video_personalization}

Video personalization aims to generate videos that maintain consistent identity with the user-provided reference~\citep{dreamvideo, li2024personalvideo, consisid}.
In this section, we integrate advanced personalization techniques into the video generation pipeline, achieving state-of-the-art performance.
The subsequent discussion provides a comprehensive elaboration of our approach, encompassing the model architecture, personalization data, and experimental validation.

\subsubsection{Model Design}

The core techniques behind video personalization tackle two primary challenges: (1) the acquisition of high-fidelity personalized identities and (2) the seamless integration of these identity features into the video generation pipeline.

We start from our \method-T2V foundation model by first obtaining personalized identity information. 
Existing video personalization approaches usually rely on ID extractors (\textit{e.g.,} ArcFace~\citep{deng2019arcface}) or versatile visual extractors (\textit{e.g.,} CLIP~\citep{radford2021learning}) to obtain identity information.
While they make promising progress, their performance is bottlenecked by the limitations of the feature extractors. 
For instance, ID extractors mainly focus on features for face recognition but may not capture other critical visual cues such as scars or stickers. 
Moreover, they are susceptible to low-resolution faces, illumination variations, and partial occlusions (such as glasses, masks, or hair). 
On the other hand, versatile visual extractors are not specifically tailored to the face domain and tend to focus on coarse-grained semantic information. 
Therefore, we choose not to rely on any feature extractor to avoid information loss, and our generation process is directly conditioned on the input face image in the latent space of \method-VAE.

\begin{figure}[t]
    \scriptsize
    \centering
    \includegraphics[width=1.0\linewidth]{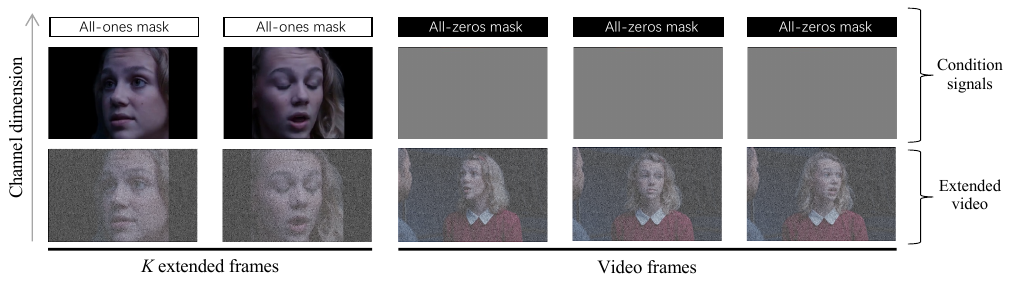}
    \caption{The core design of the video personalization approach in the latent space of \method-VAE.}
    \label{fig:framework_personalization}
\end{figure}

There are many ways to inject the personalized identity information into the video generation process. In our experiments, we found that cross-attention operation is suited to interact with dense representations obtained from identity extractors, while self-attention is more appropriate for modeling data in the same latent space. Fig.~\ref{fig:framework_personalization} illustrates the design of our video personalization approach, which follows a self-attention paradigm.
Specifically, in the latent space, we first extend $K$ additional frames prepended to the given video using the segmented face images. 
These face images are extracted from the paired video using face detection and segmentation, and the facial landmarks are further aligned to a black canvas that has the same size as the video frame. 
Then, along the channel dimension of this extended video, we concatenate the face images with all-ones masks for the first $K$ frames, and blank images with all-zeros masks for the remaining frames. 
These concatenated signals serve as the conditions. 
Finally, the diffusion process is performed on this temporally extended video, conditioned on the channel-wise condition signals in an inpainting fashion. 
Note that no other modifications are applied to our \method-T2V model. 

During training, we randomly drop a portion of face images in the $K$ extended frames, so as to support $0$ to $K$ reference face(s) video generation. 
In the inference stage, starting from the pure noise of the extended video, we concatenate the user-provided face images (no more than $K$) to the extended frames in the channel dimension and set the corresponding masks as all-ones. 
The goal is to reconstruct the face images in the first $K$ frames, and synthesize a new personalized video in the subsequent frames that preserves the provided identity.

\subsubsection{Dataset}
We curate a collection of personalization data from our T2V foundation model's training dataset through careful filtering and automatic data synthesizing. 
We first filter out about $\mathcal{O}(100)$M videos using our internal human classifiers and perform face detection at 1 FPS on these videos. 
A video shall be discarded if more than one face is detected in any frame or if over 10\% of the frames have no face detected. 
Then we calculate ArcFace similarities between consecutive frames and discard videos with low similarities. 
Afterwards, we perform face segmentation to remove the distracting backgrounds, and perform face landmark detection so as to facilitate canvas alignment during our training. 
Note that we do not filter out faces with the small areas, because we find that such videos usually contain full-body human figures. 
Finally, we construct about $\mathcal{O}(10)$M personalized videos, where each video is accompanied by 5 segmented faces in average.

We also resort to automatic data synthesizing to improve facial diversity. 
Particularly, we randomly select $\mathcal{O}(1)$M personalized videos from above, and adopt Instant-ID~\citep{wang2024instantid} to synthesize diverse faces for each video. 
Notably, we build a text template with over 100 prompts, including anime, line art, cinematic, Minecraft, and so on. Every time we take a random prompt from this template, as well as a random human pose estimated from the rest of the videos, to serve as inputs to Instant-ID. We further measure the ArcFace similarity on the generated faces and filter out those with lower similarities. 
In total, we gather about $\mathcal{O}(1)$M videos with synthesized faces, which greatly improve the diversity of styles, poses, illuminations, and occlusions on our personalization dataset.

\begin{table}[!h]
    \centering
    \begin{tabular}{l| c c c c c}
         & \method & CN-TopA & CN-TopB & CN-TopC \\ 
        \shline
        \rule{0pt}{15pt} 
        Arcface Similarity & 
            0.5526 & 
            0.5655 & 
            0.5197 &
            0.4998\\

    \end{tabular}
    \caption{Comparison of other leading video personalization methods.}
    \label{tab:video_personalization_compare}
\end{table}

\subsubsection{Evaluation}
\begin{figure}[htbp]
    \scriptsize
    \centering
    \includegraphics[width=\linewidth]{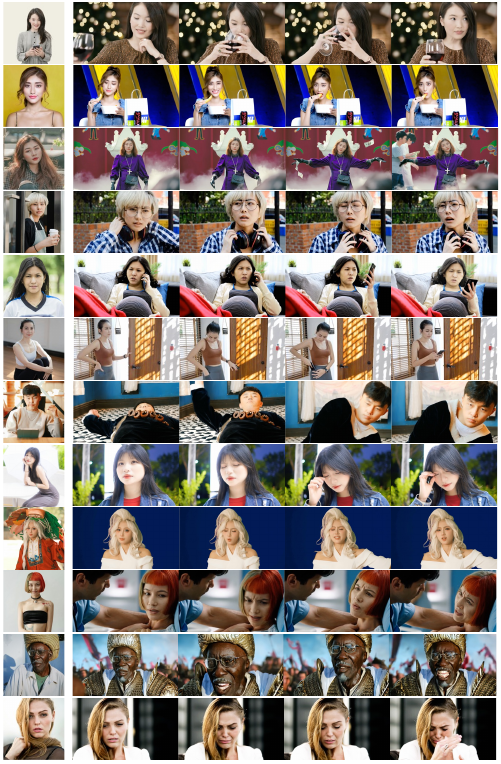}
    \caption{Visualization of our video personalization approach.}
    \label{fig:ID_cases}
\end{figure}

We present our video personalization results in Fig.~\ref{fig:ID_cases}, where the left part represents the input identities and the right part represents the synthesized personalized videos. 
All the testing images are randomly picked from Pexels\footnote{www.pexels.com}. 
We also evaluate our model on an unseen evaluation set and measure the ArcFace similarity between input faces and output personalized videos by sampling faces at 1 FPS from videos. 
The final similarity scores are averaged from these sampled faces. 
The corresponding results are illustrated in Table~\ref{tab:video_personalization_compare}, where our approach demonstrates competitive video personalization performance compared to other leading commercial and closed-source Chinese competitors.

\begin{figure}[ht!]  
    \scriptsize
    \centering  
    \includegraphics[width=0.9\textwidth]{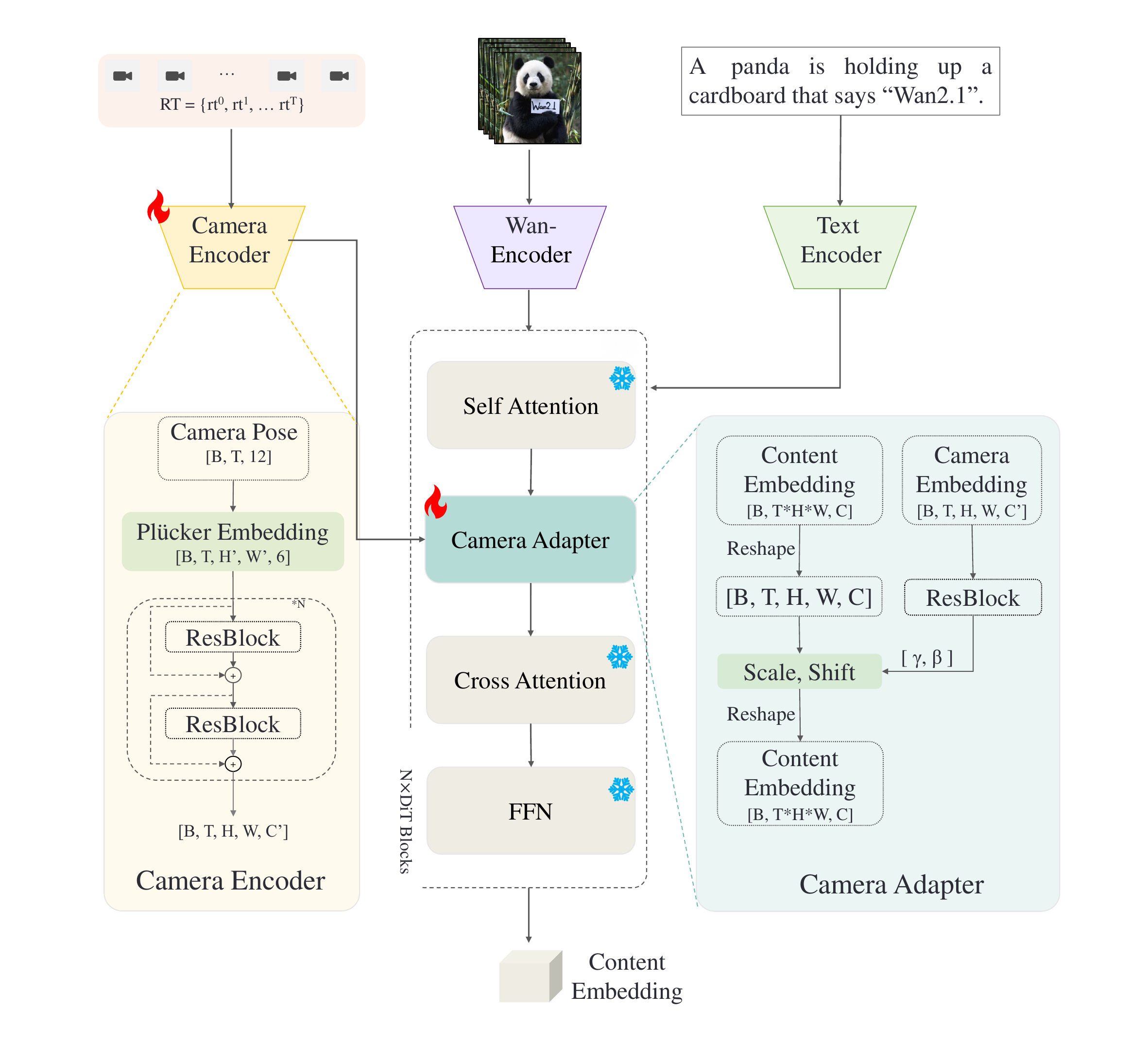}\vspace{-4mm}
    \caption{Framework of video generation guided by camera motion.} 
    \label{fig:framework_camera_motion}  
\end{figure}

\subsection{Camera Motion Controllability}
\label{sec:camera_motion}

The camera motion control module is designed to accurately match the motion and viewpoint of the video by leveraging camera trajectory.
Specifically, we utilize the extrinsic parameters $[R,t] \in \mathbb{R}^{3 \times 4}$ and intrinsic parameters $K_{f} \in \mathbb{R}^{3 \times 3}$ for each frame.
Our approach consists of two main components to effectively inject camera motion conditions: camera pose encoder and camera pose adapter, as illustrated in Fig.~\ref{fig:framework_camera_motion}.

\begin{figure}[t]  
    \scriptsize
    \centering  
\includegraphics[width=0.95\textwidth]{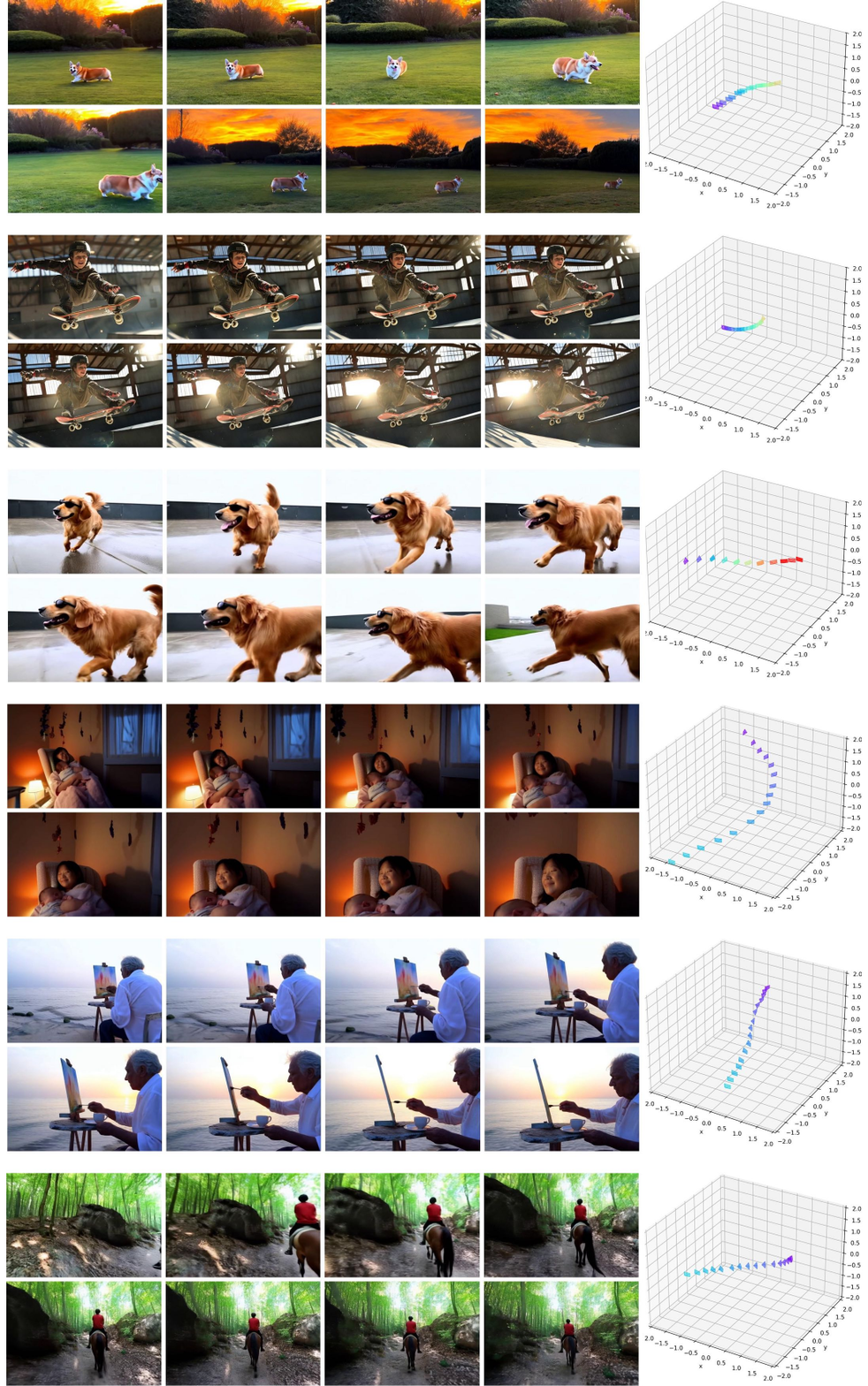}  
    \caption{Generated video results guided by camera motion.} 
    \label{fig:results_camera_motion}  
\end{figure}

\textbf{Camera pose encoder.} First, for each pixel in each video frame, we use Plücker coordinates to transform the given extrinsic and intrinsic parameters into a sequence of fine-grained positions $P\in \mathbb{R}^{6\times F\times H \times W}$.
To adjust the compression resolution of video latent features, we apply the PixelUnshuffle operation in Pytorch~\citep{paszke2019pytorch} to reduce the spatial resolution of $P$ while increasing the number of channels. Finally, we encode the output using a series of convolutional modules aligned with the number of DiT blocks to extract multi-level camera motion features.

\textbf{Camera pose adapter.} To integrate camera motion features into video latent features,  we adopt an adaptive normalization layer.
Specifically, we transform the input camera motion feature sequence into scaling factors $\gamma_{i}$ and shifting parameters $\beta_{i}$ using two zero-initialized convolution layers. Then, $\gamma_{i}$ and $\beta_{i}$ are incorporated into each DiT block through a simple linear projection defined by the formula $f_{i} =(\gamma_{i}+1)*f_{i-1}+ \beta_{i}$, where $f_{i}$ represents the video latent features at layer $i$.

Regarding the training data, we utilize an advanced camera pose estimation algorithm named VGG-SfM ~\citep{wang2024vggsfm} to extract camera trajectories from our training videos that exhibit significant camera motion.
This process yields approximately $\mathcal{O}(1)$ thousand video segments.
We train our module within our text-to-video generation framework using the Adam optimizer.
Fig.~\ref{fig:results_camera_motion} presents various example results of videos generated under camera motion guidance.

\subsection{Real-time Video Generation}
\label{sec:real_time_gen}

Current video generative methods often require significant computational resources and considerable time to produce even short video clips. Systems relying on high-end hardware remain slow, often taking several minutes to generate just a few seconds of video. While the visual quality of these generated videos has seen substantial improvements in recent years, the extended processing times present a major bottleneck in practical applications and iterative design workflows. In fields where rapid prototyping and real-time adjustments are crucial—such as interactive entertainment, virtual reality, and video production—these long generation times limit creative flexibility and slow down the process of refining visual content.

\begin{figure*}[ht]
    \centering
    \includegraphics[width=1\linewidth]{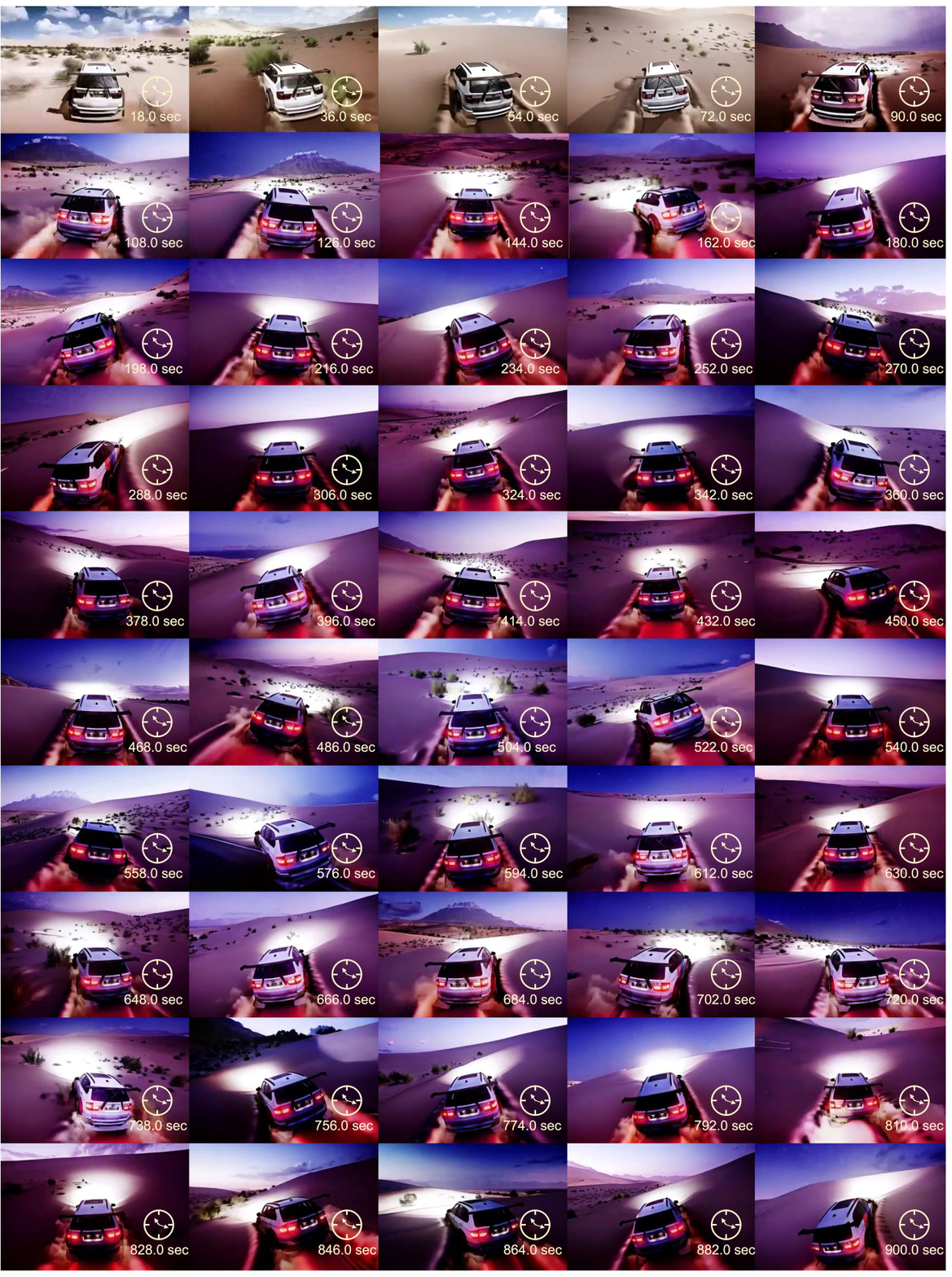}
    \caption{Generated 15 minutes video with 8 A100 GPUs in real-time 8 FPS.}
    \label{fig:streamer1}
\end{figure*}

\begin{figure*}[ht]
    \centering
    \includegraphics[width=1.\linewidth]{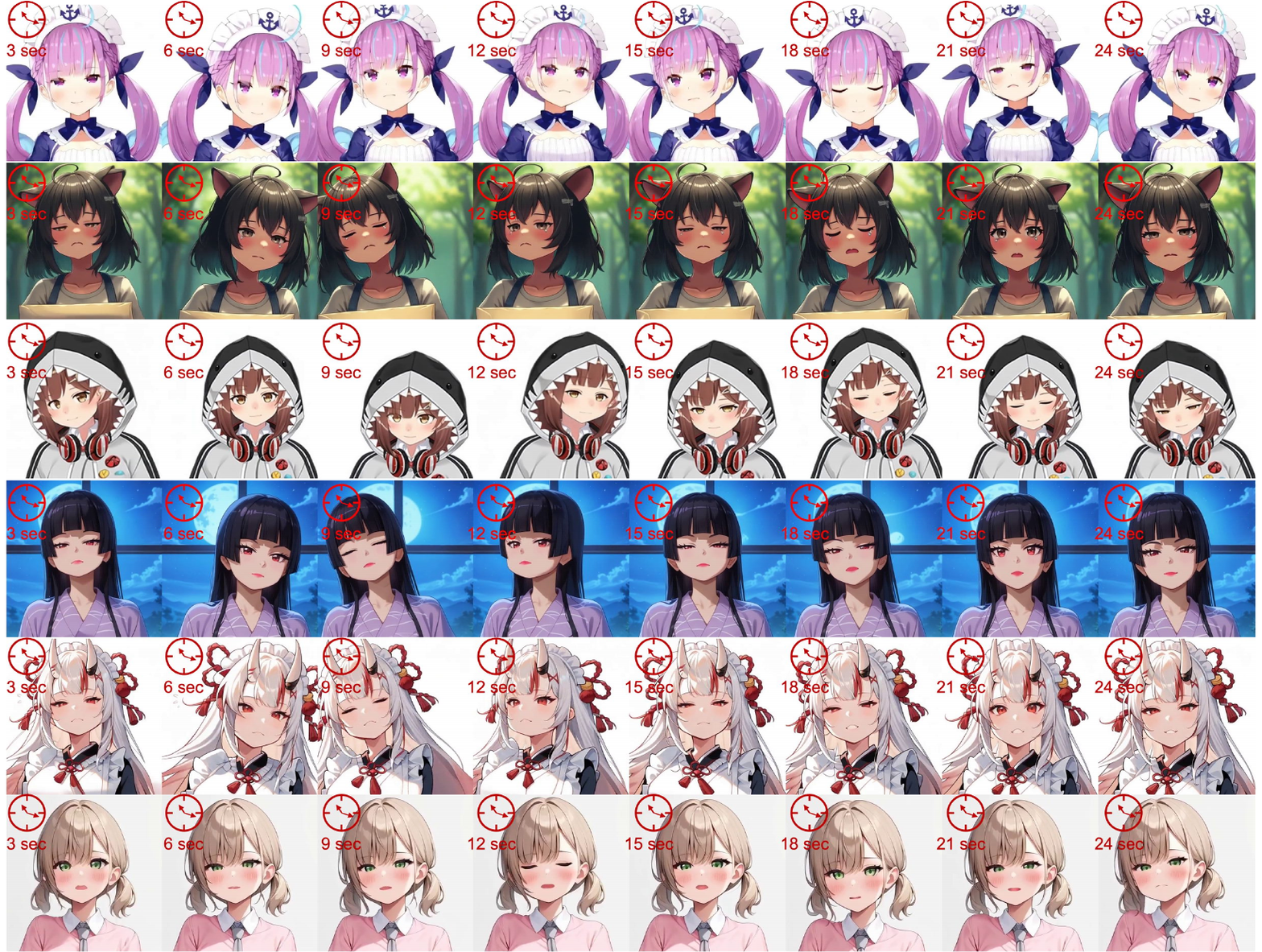}
    \caption{Generated videos with a single RTX 4090 GPU, int8 and TensorRT quantization in real-time 20 FPS.}
    \label{fig:streamer2}
\end{figure*}

Moreover, the slow speed of current methods limits their use in dynamic environments where real-time feedback is essential. In contexts like live-streaming or gaming, there’s an increasing demand for systems that can instantly render synthetic scenes and characters in response to user actions or changing inputs. Any delay in generating high-quality video clips can result in a suboptimal user experience, reducing immersion and interactivity. By focusing on real-time generation, \method seeks to enhance both the performance and applicability of video generation techniques, broadening their potential in fast-paced, interactive environments. Fig.~\ref{fig:streamer1} and \ref{fig:streamer2} provide such cases of real-time long video stream generation. We then introduce how to achieve this using our \method models.

\subsubsection{Method}
To construct the real-time generation pipeline, we build upon a previously pre-trained version of the \method model. This design choice offers two primary benefits. First, starting from a well-trained model significantly accelerates convergence and improves training stability, as the model already possesses a strong initialization. Second, the pre-trained model has already captured valuable knowledge about motion patterns and continuous temporal dynamics, which can be directly inherited by the new pipeline to enhance the smoothness and coherence of generated content. Therefore, we adopt the pre-trained \method model as the foundation for our real-time pipeline~\footnote{Part of this section is introduced in a previously released tech report \emph{The Matrix}~\citep{feng2024matrixinfinitehorizonworldgeneration}.}.

The pre-trained \method model is designed to generate fixed-length video clips, typically 5 to 10 seconds long. To transform this into a real-time streaming model capable of generating continuous video streams with no predefined length (potentially infinite), two key modifications are required:
\begin{itemize}
    \item \textbf{Streaming pipeline adaptation.} We replace the static generation process, which produces a fixed-length sequence of video tokens in a single pass, with a streaming mechanism that incrementally generates video tokens. In this streaming setup, video tokens are processed through a denoising queue — each time the oldest token is generated and dequeued, a new token is added to the bottom of the queue, enabling continuous generation without length constraints.
    \item \textbf{Real-time acceleration.} In addition to enabling streaming, we optimize the generation speed to ensure the system meets real-time performance requirements, \ie the pipeline must generate new frames fast enough to keep up with real-time playback.
\end{itemize}
In the following sections, we will describe each of these modifications in detail, including their design rationale, technical implementation, and empirical performance.

\subsubsection{Streaming Video Generation}
Traditional Diffusion Transformer (DiT)~\citep{dit} models are constrained in their ability to generate videos longer than a few seconds, even when employing significant spatial and temporal compression through Variational Autoencoders (VAEs)~\citep{kingma2013auto}. This limitation stems primarily from the computational and memory-intensive nature of attention mechanisms over extended temporal sequences. To overcome this challenge, we introduce Streamer, a novel approach that leverages a sliding temporal window to manage temporal dependencies efficiently, enabling the generation of long or even infinite-length videos in a streaming fashion.

\paragraph{Shift window denoise process models.} The key innovation of Streamer lies in its assumption that temporal dependencies are confined within a limited time window. By focusing attention computations within this window, Streamer significantly reduces computational overhead while maintaining video continuity. Specifically, Streamer processes video tokens in a queue, denoising them over multiple noise levels within the window. After a fixed number of denoising steps, the leftmost token (with the lowest noise level) is dequeued and cached, while a new token with Gaussian noise is appended to the rightmost position. This sliding mechanism ensures that the window maintains a consistent length while facilitating the generation of continuous video frames.

\paragraph{Training and inference.}
Streamer is fine-tuned from a pre-trained DiT model. During training, a sequence of $2w$ video tokens is sampled, where $w$ is the size of the sliding window (typically set to $w=T$, the number of diffusion solver steps). The first $w$ tokens are used solely for `warming up' the model and do not contribute to the loss computation. The loss is calculated only on the last $w$ tokens, ensuring that the model learns to generate coherent video frames within the window. At inference time, the same warmup strategy is employed: the first $w$ tokens are discarded, and the generated video begins from the $(w+1)$-th token.

\paragraph{Ensuring continuity during training and inference.}
To maintain smooth transitions between consecutive windows, cached tokens are re-introduced into the token queue at a noise level of 0. This allows previously generated tokens to participate in the denoising process of subsequent windows, ensuring temporal consistency and continuity in the generated video.

Streamer represents a significant advancement in video generation by adapting DiT models for streaming applications. Its sliding temporal window approach addresses the limitations of traditional DiT models, enabling the efficient generation of infinite-length videos while maintaining high quality and temporal coherence. This method opens new possibilities for real-time video generation and long-form content creation. We summarize its advantages as follows:
\begin{itemize}
    \item Infinite-Length Video Generation: By leveraging a sliding window mechanism, Streamer can generate videos of arbitrary length without incurring prohibitive computational costs;
\item Efficient Attention Computation: Limiting attention computations to a confined temporal window reduces memory and processing demands;
\item Seamless Continuity: The caching and re-introduction of tokens ensure smooth transitions between windows, preserving video quality over extended durations.
\end{itemize}

\subsubsection{Consistency Model Distillation}
After successfully extending the DiT model to Streamer for infinite-length video generation, the next critical step is to address the challenge of achieving real-time rendering of the simulated world. While Streamer enables the generation of long or even infinite videos by efficiently managing temporal dependencies within a sliding window, the computational demands of the diffusion process still pose a bottleneck for real-time applications. To overcome this, we propose a novel integration of Streamer with Consistency Models, a cutting-edge approach for accelerating diffusion-based generative processes. Specifically, we leverage the Latent Consistency Model (LCM~\citep{luo2023latent}) and its video version (VideoLCM~\citep{wang2023videolcm}), which distill the original diffusion process and  class-free guidance into a highly efficient four-step consistency model. This distillation process significantly reduces the number of sampling steps required for high-quality generation while maintaining the temporal coherence and streaming capabilities inherent to Streamer.

The integration of LCM with Streamer is designed to preserve the denoising window mechanism, ensuring that the sliding temporal window continues to manage dependencies effectively. During training, we optimize this combined framework to balance efficiency and quality. The result is a 10 - 20 $\times$ acceleration in inference speed, enabling the model to achieve a rendering rate of 8 - 16 FPS. This dramatic improvement in performance makes the system viable for real-time applications, such as interactive simulations, live video synthesis, and dynamic content generation in virtual environments. By combining the strengths of Streamer—its ability to generate infinite-length videos with temporal coherence—and LCM—its efficiency in accelerating the diffusion process—we bridge the gap between high-quality video generation and real-time responsiveness. This advancement not only enhances the practicality of diffusion models for streaming applications but also opens up new possibilities for immersive, interactive, and real-time media creation.

\paragraph{Quantization for running in customer-level devices.} While the generating speed has been accelerated to real-time levels, deploying the model on consumer-level devices remains challenging due to the high computational and memory requirements, even for high-end GPUs like the NVIDIA 4090. To address this, we introduce quantization techniques to optimize the model for efficient deployment. Specifically, we employ two distinct quantization strategies: int8 quantization~\citep{torchao} for attention layers and linear heads, and TensorRT quantization~\citep{tensorrt} for the overall model. The int8 quantization approach significantly reduces memory consumption by converting weights and activations to 8-bit integers, while maintaining the overall generation quality. However, this method provides minimal acceleration in generating speed, as it primarily focuses on memory efficiency rather than computational optimization. On the other hand, TensorRT quantization offers a more comprehensive solution, enabling substantial acceleration in generation speed. This technique allows the model to achieve real-time performance of 8 FPS even on a single 4090 GPU, making it feasible for consumer-level devices. However, TensorRT quantization introduces a modest increase in network error, which can be observed even when its built-in error-checking mechanisms are enabled. This error manifests as slight deviations in output quality, such as minor artifacts or inconsistencies in the generated video. In some cases, it may also lead to measurable instability, such as flickering or temporal incoherence. Despite these trade-offs, the combination of int8 and TensorRT quantization provides a balanced approach to optimizing the model for real-time, consumer-level deployment, ensuring both efficiency and practicality while maintaining acceptable generation quality. By carefully tuning the quantization parameters and leveraging TensorRT’s error-checking features, we mitigate these issues to the greatest extent possible, enabling robust and efficient performance on consumer-grade hardware.

\subsection{Audio Generation}
\label{sec:audio_gen}

The primary objective of audio generation in this work is to produce synchronized soundtracks for video clips, formulated as a video-to-audio (V2A) generation framework. 
The generated soundtracks consist of ambient sound and background music, explicitly excluding speech or vocal elements. 
In contrast to text-to-audio models ~\citep{audioldm, audioldm2}, the ambient sound generated by video-to-audio models must be temporally aligned with the visual content of the video, while the accompanying music should accurately reflect the emotional tone and contextual setting of the video, as would naturally be expected. Furthermore, to enhance user control over sound design, our framework integrates both video and textual prompts, enabling users to specify on-screen or off-screen sounds and define the style and presence of background music.

\subsubsection{Model Design}
\begin{figure}[t]  
    \centering  
    \includegraphics[width=0.99\textwidth]{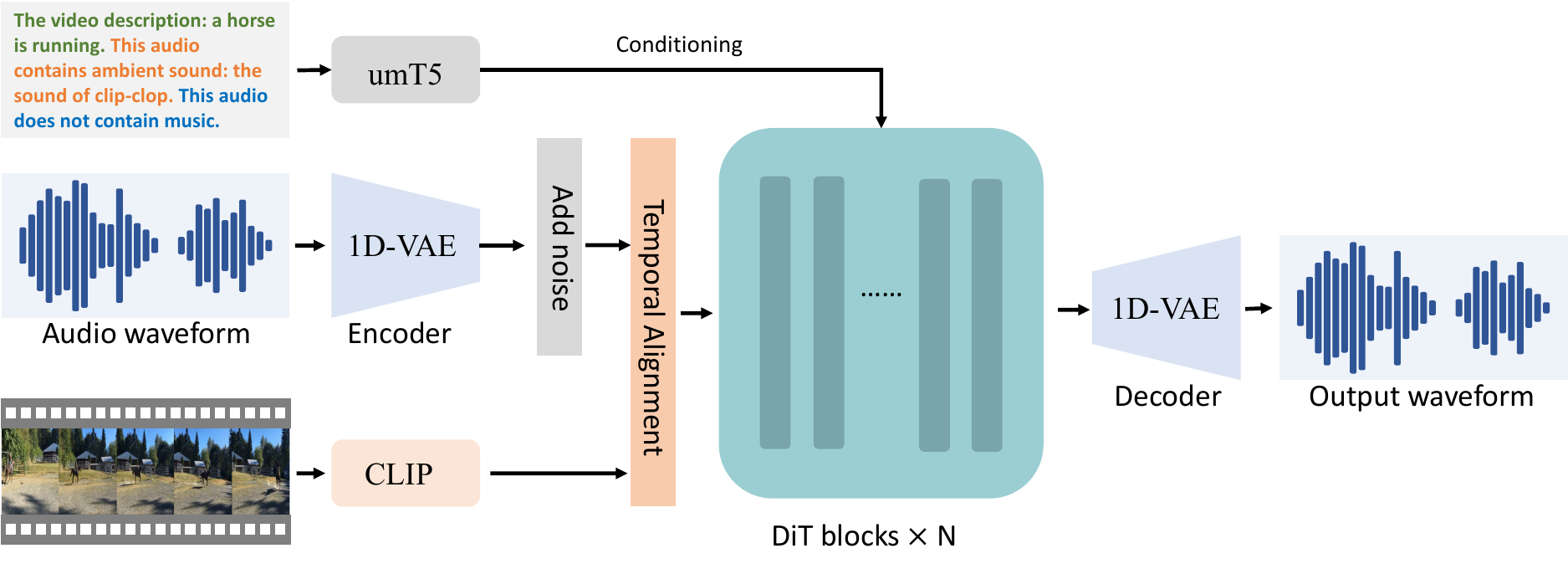}  
    \caption{Framework of video-to-audio generation. Our model processes both a video chunk and its associated textual description as dual inputs to synthesize high-quality, semantically coherent audio.} 
    \label{fig:v2av2-pipeline} %
\end{figure}

Consistent with our video generation framework, our video-to-audio model also utilizes a diffusion transformer (DiT) ~\citep{dit} based on the flow-matching~\citep{flow-matching} to model the denoising diffusion process in the audio domain. The detailed pipeline of our V2A model is illustrated in Fig.~\ref{fig:v2av2-pipeline}. 

\begin{figure}[htbp]  
    \centering  
    \includegraphics[width=0.99\textwidth]{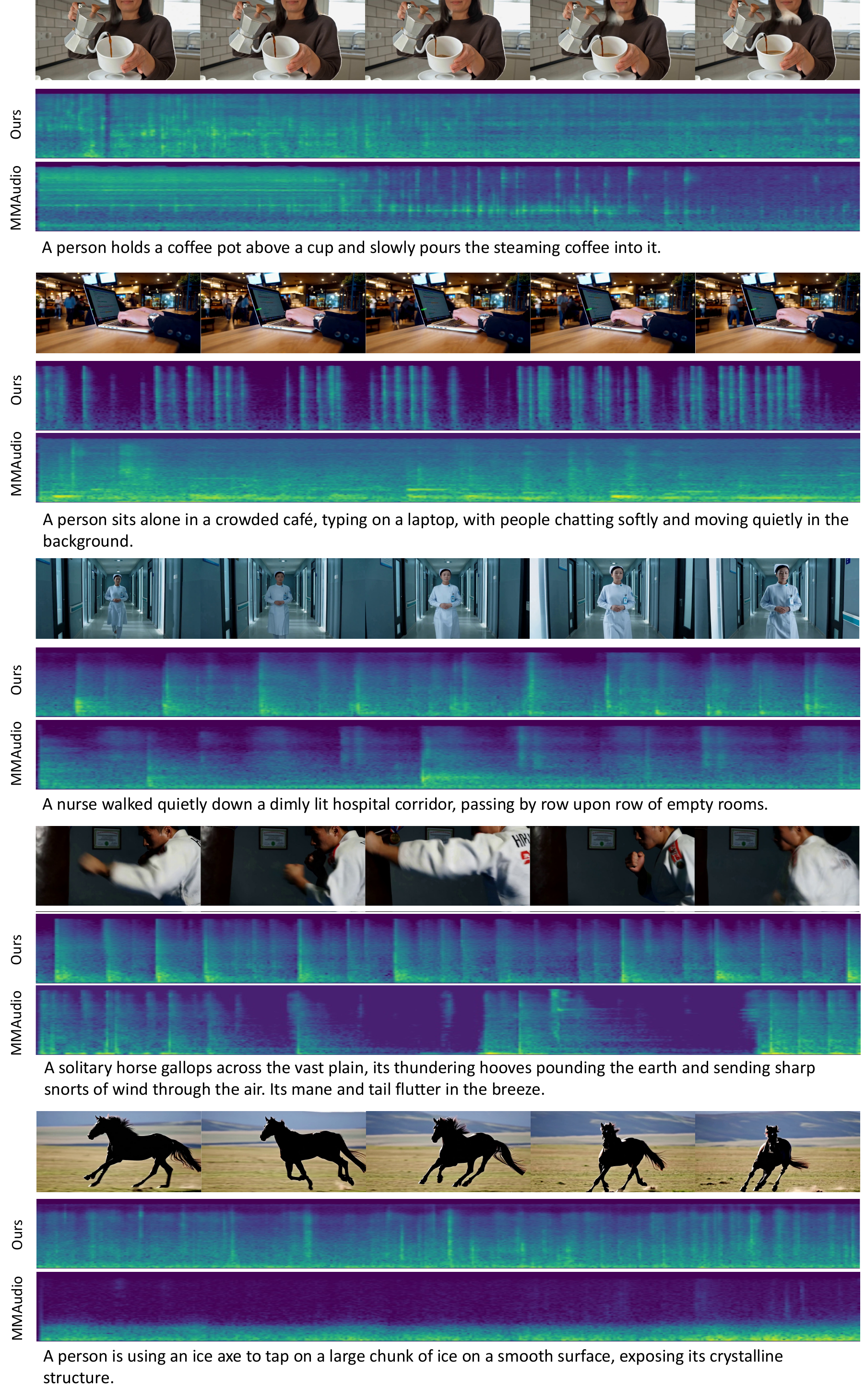}  
    \caption{Audio generation samples of MMAudio and ours.} 
    \label{fig:v2av2-goodcases-pickup} %
\end{figure}

\textbf{Audio autoencoder}. 
A widely adopted audio compression approach involves transforming raw waveforms into mel-spectrograms, followed by encoding them into $H^{a}\times W^{a}\times C^{a}$ latent features using an image-based variational autoencoder, as demonstrated in ~\citep{audioldm, audioldm2}. However, in the context of a diffusion transformer backbone, these latent features must be partitioned into patches and reshaped into a tensor of size $(H^a\cdot W^a)\times C^a$. This patching-and-reshaping process can disrupt the temporal alignment with the corresponding video content, posing a significant challenge for synchronization. Considering this limitation, we train a compression model that operates directly on raw waveforms. Our 1D-VAE generates latent features of size $T^{a}\times C^{a}$, where $T^a$ represents the sequence length along the time axis, preserving explicit temporal information essential for accurate synchronization.

\textbf{Video and text encoder}. To achieve frame-accurate synchronization between synthesized audio and visual sequences, we establish temporal coherence through multimodal feature fusion. Our architecture first extracts frame-wise visual embeddings using a CLIP~\cite{radford2021learning} model, then performs temporal rate adaptation through feature replication to match the audio feature's sampling rate, following~\citep{polyak2024moviegencastmedia}. The synchronized visual and acoustic features undergo dimension transformation via linear projection layers to align with DiT's latent space, followed by element-wise summation for multimodal fusion. For linguistic understanding across languages, we leverage the pre-trained umT5 model~\citep{chung2023unimax} as our text encoder, capitalizing on its cross-lingual representation capabilities through frozen parameters.

\textbf{Data.} The training data for our V2A model is derived from the video generation dataset through a rigorous filtering process. We systematically remove videos lacking soundtracks or containing speech/vocal music, resulting in a refined subset of $\mathcal{O}(1)$ thousand hours. To enhance multimodal representation, we employ a comprehensive captioning strategy: dense video descriptions are complemented with audio-specific captions generated by Qwen2-audio~\citep{qwen2-audio}. These audio captions are categorized into ambient sounds and musical compositions, with the latter characterized by style, rhythm, melody, and instrumentation. The final structured caption integrates three components: (1) dense video description, (2) ambient sound characterization, and (3) background music analysis, providing a unified multimodal representation for training.

\textbf{Implementation details.} Our V2A architecture generates high-fidelity stereo audio with a maximum duration of 12 seconds at a 44.1 kHz sampling rate. The model processes multimodal inputs through dedicated encoders: umT5-XXL produces 4096-dimensional text embeddings, while the CLIP extracts 1024-dimensional visual embeddings per frame. The audio and video features are projected into a unified 1536-dimensional latent space within the DiT backbone. For temporal alignment, we downsample input videos and yield precisely 48 frames for 12-second clips, ensuring frame-accurate synchronization between visual and audio modalities. The system processes latent sequences of length 256 in one batch. To enhance the model's capability of generating audio solely from visual cues, we implement a random masking strategy during training where ambient sound and music captions are selectively omitted with a predefined probability. This conditioning mechanism forces the model to establish robust cross-modal associations between visual content and corresponding audio patterns, while maintaining the flexibility to utilize textual descriptions when available.

\subsubsection{Evaluation}
We provide qualitative evaluations of audio generation in Fig.~\ref{fig:v2av2-goodcases-pickup}, where we compare our method with the recently released open-sourced approach MMAudio~\citep{mmaudio}. This comparative analysis is particularly meaningful as MMAudio represents the most similar and competitive baseline to our method, demonstrating impressive performance in audio generation tasks. The videos are generated by our text-to-video model with a duration of 5 seconds.

Our V2A model demonstrates superior performance in several key aspects of audio generation, as evidenced by the comparative results. Specifically, the model exhibits enhanced long-term consistency, particularly noticeable in the `pouring coffee' scenario (the first case in Fig.~\ref{fig:v2av2-goodcases-pickup}). In the second case of `typing', our approach generates significantly cleaner audio output compared to the noisier results produced by MMAudio. Furthermore, our method excels in synthesizing rhythmic sound patterns, as demonstrated in the `walking', `boxing', and `clip-clop' examples (the last three cases in Fig.~\ref{fig:v2av2-goodcases-pickup}), where it achieves more natural and coherent audio generation.

\textbf{Limitations.} Our method exhibits limitations in generating human vocal sounds, including but not limited to laughter, crying, and speech. This limitation is primarily attributed to our data preparation process, where speech-related vocal data is deliberately excluded from the training dataset. The MMAudio model's ability to produce random speech-like sounds stems from its retention of speech-related data in its training corpus. In our future work, we plan to incorporate speech generation capabilities to address this limitation.

\section{Limitation and Conclusion}
\label{sec:conclusion}

\textbf{Limitation}. This study presents \method, a foundational video generation model that has achieved notable advancements across multiple benchmarks.
Specifically, significant improvements have been observed in motion amplitude (\emph{e.g}., sports, dance) and instruction-following capabilities. 
Nevertheless, several limitations persist, which should be addressed to realize more potential of \method.
First, preserving fine-grained details in scenarios involving large motion remains a challenge for our method, as well as a broader issue within the field of video generation. 
Substantial efforts are required to improve the fidelity of videos with large-scale motion dynamics.
Second, the computational cost associated with large-scale models remains prohibitive. 
Inference on a 14B model currently requires approximately 30 minutes on a single head-end GPU without additional optimization. 
To democratize video generation as a universally accessible AI tool, further research into efficiency and scalability is essential.
Finally, there is still a lack of domain-specific expertise. As a foundational video model, we are committed to enhancing \method's general capabilities, yet performance in specific localized scenarios, such as education and medicine, may be insufficient. 
We aim to address this limitation by open-sourcing our latest models and fostering community-driven development across diverse specialized domains.

\textbf{Conclusion}. In this report, we publicly release \method, our latest video model, which establishes a new benchmark for video generation.
We provide a comprehensive overview of the architectural design of our \method-VAE and DiT models, including detailed insights into their training approach, data curation processes, evaluation methods, and empirical results.
Furthermore, we meticulously analyze our data preprocessing pipeline and the strategic adaptations implemented to optimize model training. 
This holistic approach is designed to drive advancements in the domain of video generation.
We also extensively investigate downstream applications, such as image-to-video generation, video editing, and personalized video generation, to demonstrate the versatility and practical utility of \method across diverse scenarios.
In addition to open-sourcing the 14B model, we explore the feasibility of leveraging smaller-scale models for efficient video generation. 
Notably, our 1.3B model not only achieves competitive performance compared to larger counterparts but also enables seamless inference on consumer-grade GPUs, significantly enhancing accessibility and practicality for content creators.
Looking ahead, we plan to focus on scaling both data and model architectures to tackle the most pressing challenges in video generation. 
Our ongoing efforts aim to provide the research community with more robust and versatile tools for video creation, fostering innovation and broader adoption in this rapidly evolving field.

\newpage
\section{Contributors}

\large{Authors are listed \textbf{alphabetically by the first name}.} 

\definecolor{damaired}{RGB}{100, 0, 0}

\large{
\color{damaired}%
\begin{multicols}{3}
\raggedcolumns
Ang Wang\\
Baole Ai\\
Bin Wen\\
Chaojie Mao\\
Chen-Wei Xie\\
Di Chen\\
Feiwu Yu\\
Haiming Zhao\\
Jianxiao Yang\\
Jianyuan Zeng\\
Jiayu Wang\\
Jingfeng Zhang\\
Jingren Zhou\\
Jinkai Wang\\
Jixuan Chen\\
Kai Zhu\\
Kang Zhao\\
Keyu Yan\\
Lianghua Huang\\
Mengyang Feng\\
Ningyi Zhang\\
Pandeng Li\\
Pingyu Wu\\
Ruihang Chu\\
Ruili Feng\\
Shiwei Zhang\\
Siyang Sun\\
Tao Fang\\
Tianxing Wang\\
Tianyi Gui\\
Tingyu Weng\\
Tong Shen\\
Wei Lin\\
Wei Wang\textasciitilde1\\
Wei Wang\textasciitilde2\\
Wenmeng Zhou\\
Wente Wang\\
Wenting Shen\\
Wenyuan Yu\\
Xianzhong Shi\\
Xiaoming Huang\\
\\
Xin Xu\\
Yan Kou\\
Yangyu Lv\\
Yifei Li\\
Yijing Liu\\
Yiming Wang\\
Yingya Zhang\\
Yitong Huang\\
Yong Li\\
You Wu\\
Yu Liu\\
Yulin Pan\\
Yun Zheng\\
Yuntao Hong\\
Yupeng Shi\\
Yutong Feng\\
Zeyinzi Jiang\\
Zhen Han\\
Zhi-Fan Wu\\
Ziyu Liu
\end{multicols}}

\clearpage

\bibliography{biblio}
\bibliographystyle{iclr2023_conference}
\clearpage

\end{document}